\documentclass[11pt]{article}

\divide \topmargin by 2 \advance \topmargin -1in
\leftmargin -\textwidth
\divide \leftmargin by 2 \advance \leftmargin -1in
\oddsidemargin \leftmargin \evensidemargin \leftmargin
\usepackage[english]{babel}
\usepackage{amsfonts}
\usepackage{amssymb}
\usepackage{mathrsfs}
\usepackage{amsthm}
\usepackage{amsmath}
\usepackage{listings}
\usepackage{relsize}
\usepackage{graphicx}
\usepackage{float}
\usepackage{bbm}
\usepackage{bm}
\usepackage{natbib}
\usepackage{tikz}
\usetikzlibrary{arrows.meta}
\usepackage{pgfplots}
\usepackage{mathtools}
\usepackage{setspace}
\spacing{1.5}

\usepackage{verbatim}

\usepackage{multirow}
\usepackage[linesnumbered,lined,boxed,commentsnumbered]{algorithm2e}
\usepackage{pgfplots}
\usepackage[pdfborder={0 0 0}]{hyperref}
\usepackage[utf8]{inputenc}

\usepackage{fmtcount}

\usepackage{enumitem}

\usepgfplotslibrary{fillbetween}
\usetikzlibrary{patterns}
\usetikzlibrary{arrows}
\usetikzlibrary{datavisualization.formats.functions}
\usetikzlibrary{decorations.text}

\pgfplotsset{
    box plot/.style={
        /pgfplots/.cd,
        black,
        only marks,
        mark=-,
        mark size=0.2em,
        /pgfplots/error bars/.cd,
        error bar style={dashed},
        y dir=plus,
        y explicit,
    },
    box plot top whisker/.style={
        /pgfplots/error bars/draw error bar/.code 2 args={%
            \pgfkeysgetvalue{/pgfplots/error bars/error mark}%
            {\pgfplotserrorbarsmark}%
            \pgfkeysgetvalue{/pgfplots/error bars/error mark options}%
            {\pgfplotserrorbarsmarkopts}%
            \path ##1 -- ##2;
        },
        /pgfplots/table/.cd,
        y index=2,
        y error expr={\thisrowno{1}-\thisrowno{2}},
        /pgfplots/box plot
    },
    box plot bottom whisker/.style={
        /pgfplots/error bars/draw error bar/.code 2 args={%
            \pgfkeysgetvalue{/pgfplots/error bars/error mark}%
            {\pgfplotserrorbarsmark}%
            \pgfkeysgetvalue{/pgfplots/error bars/error mark options}%
            {\pgfplotserrorbarsmarkopts}%
            \path ##1 -- ##2;
        },
        /pgfplots/table/.cd,
        y index=3,
        y error expr={\thisrowno{1}-\thisrowno{3}},
        /pgfplots/box plot
    },
    box plot median/.style={
        /pgfplots/box plot
    }
}

\newcommand{\R}{\ensuremath{\mathbb{R}}}
\newcommand{\Z}{\ensuremath{\mathbb{Z}}}
\renewcommand{\P}{\ensuremath{\mathbb{P}}}

\newcommand{\E}{\ensuremath{\mathbb{E}}}

\newcommand{\F}{\ensuremath{\mathcal{F}}}

\newcommand{\N}{\ensuremath{\mathbb{N}}}

\renewcommand{\O}{\ensuremath{\mathcal{O}}}

\newcommand{\floor}[1]{\ensuremath{\left\lfloor #1 \right\rfloor}}
\newcommand{\ceil}[1]{\ensuremath{\left\lceil #1 \right\rceil}}

\renewcommand{\d}{\ensuremath{{\rm d}}}
\newcommand{\dx}{\ensuremath{{\rm d}x}}

\newcommand{\ds}{\ensuremath{\displaystyle}}
\newcommand{\com}[1]{{\textcolor{blue}{#1}}}

\newcommand{\CL}{\kappa}
\renewcommand{\footnote}[1]{\textcolor{blue}{#1}}
\newcommand{\blue}{\color{blue}}

\newcommand{\KL}{\textrm{KL}}

\newcommand{\vmax}{\overline{v}}
\newcommand{\cV}{\mathcal{V}}
\newcommand{\HRule}{\rule{\linewidth}{0.3mm}}

\newcommand{\cS}{\mathcal{S}}
\newcommand{\ra}{ \rightarrow }		
\newcommand{\thetamax}{\theta_{\text{max}}}
\newcommand{\thetamin}{\theta_{\text{min}}}
\newcommand{\cZ}{\mathcal{Z}}

\newcommand{\Cmin}{\underline{C}}
\newcommand{\Cmax}{\overline{C}}

\DeclareMathOperator*{\argmax}{arg\,max}

\usepackage{natbib}
 \bibpunct[, ]{(}{)}{,}{a}{}{,}%

\newtheorem{theorem}{Theorem}
\newtheorem{lemma}{Lemma}

\newtheorem{proposition}{Proposition}

\theoremstyle{definition}
\newtheorem{remark}{Remark}
\newtheorem{assumption}{Assumption}

\allowdisplaybreaks

\title{Continuous Assortment Optimization with Logit Choice Probabilities under Incomplete Information}
\author{Yannik Peeters, Arnoud V. den Boer, Michel Mandjes\footnote{The authors are with Amsterdam Business School, University of Amsterdam, the Netherlands. A.V. den Boer and M. Mandjes are in addition affiliated with Korteweg-de Vries Institute for Mathematics, University of Amsterdam, the Netherlands.}}
\date{March 14, 2021}

\begin{document}
\selectlanguage{english}
\maketitle
\thispagestyle{empty}

\textit{Abstract.} We consider assortment optimization over a continuous spectrum of products represented by the unit interval, where the seller's problem consists of determining the optimal subset of products to offer to potential customers. To describe the relation between assortment and customer choice, we propose a probabilistic choice model that forms the continuous counterpart of the widely studied discrete multinomial logit model. We consider the seller's problem under incomplete information, propose a stochastic-approximation type of policy, and show that its regret -- its performance loss compared to the optimal policy -- is only logarithmic in the time horizon. We complement this result by showing a matching lower bound on the regret of any policy, implying that our policy is asymptotically optimal. We then show that adding a capacity constraint significantly changes the structure of the problem: we construct a policy and show that its regret after $T$ time periods is bounded above by a constant times $T^{2/3}$ (up to a logarithmic term); in addition, we show that the regret of any policy is bounded from below by a positive constant times $T^{2/3}$, so that also in the capacitated case we obtain asymptotic optimality. Numerical illustrations show that our policies outperform or are on par with alternatives.

\newpage

\section{Introduction}
\label{sec:introduction} 

\subsection{Background and motivation}
In the management science and operations research literature, assortments are traditionally thought of as being of a discrete nature. However, in several applications, attributes of products or services are adjusted in a {\it continuous} manner, leading to a spectrum of similar but distinct commodities, each with a possibly different selling price. In these situations, customers can be offered highly personalized, custom-made products -- a phenomenon that the marketing literature refers to as {\it mass customization} \citep[see, e.g.,][]{Pine1993,Fogliatto2012}. Examples of attributes that can be customized in such a continuous manner include the duration of renting a commodity, the duration or amount of a mortgage, the amount of cellular data usage, or the amount of (voluntary) deductible excess in insurances. A seller of such products or services faces, in particular in the product design phase, the concrete problem of having to decide which specific subset of the spectrum to offer to potential customers, so as to maximize expected profit.

The seller's problem can be translated into a mathematical optimization problem over an uncountable space of subsets of an interval. This type of problem can only be solved efficiently when some structure is imposed on how the consumers' purchase behavior and the seller's revenue depend on the assortment that is offered. In the extensive literature on assortment optimization with a finite number of products, arguably the most-studied choice model is the so-called multinomial logit (MNL) model \citep[see, e.g.,][and the references therein]{BenAkiva1985,Mahajan2001}.
In this model, a nonnegative preference value is associated to each product (and also to the option of not purchasing a product), and the probability that a customer selects a particular product from an assortment of products is proportional to this preference value. To align our work with this rich strand of literature, we propose a choice model that is the continuous counterpart of the discrete MNL model, with the preference values replaced by a preference function. 

Importantly, we study the seller's continuous assortment optimization problem in an \emph{incomplete information} setting, meaning that the preference function is a priori unknown to the seller. To arrive at profitable assortment decisions, the seller thus has to learn the unknown preference function from accumulating sales data. This requires designing a policy that judiciously balances the two (sometimes conflicting) goals of \emph{learning} and \emph{earning}: on the one hand, the seller needs to offer assortments that support high-quality estimates of the unknown preference function; on the other hand, assortments need to be offered that yield a high profit given an available estimate of the preference function. This is an example of the well-known \emph{exploration-exploitation} trade-off in multi-armed bandit (MAB) problems: a paradigm for sequential decision problems under uncertainty. Indeed, the problem studied in this paper can be seen as a continuous, combinatorial MAB problem, where the objective is to dynamically learn which subset of the continuum maximizes the seller's expected revenue function. Designing and analyzing optimal decision policies for this novel and relevant question is the topic of this paper.

\subsection{Contributions}
\label{sec:contributions}
The contributions in this paper are as follows.
\begin{quotation}
$\circ$~First, we propose a probabilistic choice model for the setting where customers select from assortments that are subsets of the unit interval. The choice model is the continuous counterpart of the widely studied multinomial logit (MNL) model, in the sense that the continuous model arises as a limit of discrete MNL models where the number of products grow large, and, conversely, that discretizing the product space in the continuous model gives rise to a discrete MNL model.

\noindent $\circ$ Next, assuming that products are labeled in increasing order of marginal profit, we show that the optimal assortment is an interval of the form $[y,1]$, for some $y \in [0,1]$, and that the corresponding optimal expected profit is the unique solution to a fixed point equation. Leveraging this property, 
we construct a stochastic-approximation type policy, and show that its regret (the cumulative expected revenue loss compared to the optimal policy) after $T$ time periods is $\mathcal{O}(\log T)$. In addition, relying on the Van Trees inequality (which can be seen as a Bayesian version of the well-known Cram\'er-Rao lower bound), we show that the worst case regret for any policy grows as $\Omega(\log T)$, implying that our policy is asymptotically optimal. 

\noindent $\circ$ Inspired by analogous problems in the discrete setting, we then consider assortment optimization with a capacity constraint. We first show that the optimal assortment is not necessarily an interval anymore, but can have a much more complex structure. As a consequence it becomes necessary -- in contrast to the uncapacitated case -- to explore the whole product space in order to learn the optimal assortment. We propose a policy and show that, up to a logarithmic term, its regret after $T$ time periods is bounded from above by a constant times $T^{2/3}$. We then construct an instance in which the regret of any policy grows as $\Omega(T^{2/3})$, indicating that the capacitated setting indeed exhibits intrinsically different behavior than the uncapacitated case in which logarithmic regret is attainable.

\noindent $\circ$ In a numerical study we compare our algorithms against alternatives from the literature that are designed for discrete assortment optimization, and show that our algorithms outperform or are on par with these alternatives. 
Additional numerical experiments included in the Appendix show that our continuous assortment model has good predictive properties compared to its discrete counterpart, even if the true data-generating model is discrete.  


\end{quotation}

\subsection{Organization of the paper}
\label{sec:organization}

After providing an overview of relevant literature in Section \ref{sec:litreview}, we introduce our model for continuous assortment optimization in Section \ref{sec:model}. In Section \ref{sec:uncapacitated} we study assortment optimization without capacity constraints: we propose a stochastic-approximation type policy, provide an upper bound on its regret, and prove a matching lower bound on the regret of any policy. 
The capacitated problem is discussed in Section \ref{sec:capacitated}: we propose a policy, prove an upper bound on its regret, and prove a matching lower bound (up to a logarithmic term) on the regret of any policy. Section 6 contains our numerical study. 
Mathematical proofs, a discussion of the relation between the continuous and discrete logit choice model, a bisection algorithm to compute the optimal continuous assortment, and  additional numerical experiments are collected in the Appendix. 

\section{Literature}
\label{sec:litreview}
To put our work into the right perspective, we proceed by providing an account of the most relevant branches of the existing literature. 

The idea of considering a continuous spectrum of products is a well-established concept in several branches of the literature. Within the economics literature, for example, this idea is studied in the context of vertical product differentiation and customer self-selection. The seminal work by \cite{Mussa1978} assumes a linear utility-based model in which a seller offers a continuous spectrum of quality levels and tries to optimally match customers of different types to prices and quality levels. Their model was generalized by \cite{Moorthy1984} to include preferences that are nonlinear in the customer's type. 
More recently, \cite{Pan2012} considered vertical product differentiation in the context of assortment optimization, focusing on determining the optimal positioning of products to offer and corresponding selling prices. \cite{Keskin2019} consider a continuum of quality levels in a customer self-selection framework, and analyze dynamic learning of uncertain production costs. \cite{denBoer2020b} study the problem of optimally pricing and positioning a finite number of horizontally differentiated products represented by points on the unit interval, and design asymptotically optimal learning policies. Assortment optimization with product sets with a continuous structure have also been studied by \cite{Gaur2006} and \cite{Fisher2014}, who both view products as entities in an attribute space and focus explicitly on modeling substitution for finding the optimal assortment. Another example is \cite{Dewan2003}, which studies optimal product customization using the continuous, locational Salop model to determine an optimal (sub)spectrum of products to offer. With the exception of \cite{denBoer2020b} and \cite{Keskin2019}, the literature mentioned above assumes that the model primitives are known to the seller. 

The continuous choice model studied in the present paper aligns well with the widely studied discrete multinomial logit (MNL) choice model. 
Recently, several authors have studied assortment optimization under this choice model while assuming incomplete information: that is, the model parameters are unknown in advance and have to be learned from data. \cite{Rusmevichientong2010a} focus on assortment optimization with a capacity constraint, and provide a bi-section algorithm to compute the optimal assortment under full information. Under incomplete information, they show under mild conditions that the expected loss (regret) of an explore-then-exploit type of algorithm after $T$ time periods is bounded by a (instance-dependent) constant times $N^2 \log T$, 
where $N$ denotes the number of products. \cite{Saure2013} consider a similar framework with a more general utility based choice model, 
and implement procedures to quickly detect sub-optimal products. \cite{Agrawal2019} study an Upper Confidence Bound (UCB) algorithm for capacitated assortment optimization under the MNL model, and provide both a $\mathcal{O}(\sqrt{NT\log T})$ upper bound on the worst-case regret of their policy as well as an $\Omega(\sqrt{NT/K})$ lower bound for the regret of any policy, where $N$ is the total number of products and $K$ is the maximum number of products in the assortment. In addition, \cite{Agrawal2017} present a Thompson Sampling (TS) algorithm in the same setting, and provide an $\mathcal{O}(\sqrt{NT}\log TK)$ upper bound on the worst-case regret of the policy. 

The lower bound of \cite{Agrawal2019} is improved by \cite{Chen2017} to $\Omega(\sqrt{NT})$, under the assumption that $K\leqslant N/4$. Without capacity constraint, \cite{Chen2018} provide an $\mathcal{O}(\sqrt{T})$ upper bound for the regret of their policy and an $\Omega(\sqrt{T})$ lower bound for the regret of any policy, under the assumption that only the first two products have positive marginal profit. A combination of a spatially structured product set and learning is studied by \cite{Ou2018}. They present a learning algorithm for the assortment planning problem under the MNL model when the utility is a linear function of product attributes, as in the numerical study done by \cite{Rusmevichientong2010a}, and derive regret bounds.

The problem of learning the optimal assortment from accumulating data relates our work to \emph{multi-armed bandit (MAB) problems}: a framework to study sequential learning-and-optimization problems. A central theme in these problems is to determine the optimal balance between exploration (`learning') and exploitation (`earning'). 
Classically, the number of arms is assumed to be finite \citep[see, e.g.,][]{Robbins1952,Lai1985, Agrawal1995a,Auer2002}. More recently, MAB problems have been studied where the action set is a continuum \citep[see, e.g.,][]{Agrawal1995a,Agarwal2011,Kleinberg2005,Auer2007,Kleinberg2008,Bubeck2009,Cope2009,Bubeck2011a,Bubeck2011b,Flaxman2005,Shamir2013}, or where the action set consists of a (typically large) number of combinatorial structures \citep[see, e.g.,][]{CesaBianchi2012,Chen2013,Combes2015}. Our work is related to both these strands of literature: we study a MAB problem where the action sets consists of \emph{subsets} of the unit interval, comprising a combinatorial MAB problem with uncountable action set. To the best of our knowledge, such a continuous, combinatorial MAB problem has not been considered before in the literature. 

\section{Model}
\label{sec:model}

We consider a seller of a commodity or service with an attribute that can be infinitesimally adjusted to any value in the interval $[0,1]$. Each value in $[0,1]$ is referred to as a product, and the seller has to decide which assortment of products, i.e.,\ which subset of $[0,1]$, to offer to each potential customer. Upon being offered an assortment, a customer either purchases a product from the assortment, or decides not to purchase -- such a \textit{no-purchase} is denoted by $\emptyset$. 
The \textit{total collection of products} $\mathcal{X}$ is the union of the unit interval and the no-purchase option:
\[\mathcal{X} := [0,1]\cup\{\emptyset\}.\]
The goal of the seller is to identify an assortment that maximizes her expected revenue; as we shall see, this is not necessarily the entire interval $[0,1]$. We consider both capacitated and uncapacitated settings: in the former, the size of the assortment is bounded by a known constant $c<1$, whereas in the latter case, this maximum size is $c=1$. The set of feasible assortments is thus given by all (measurable) sets $S \subset [0,1]$ with volume at most $c$: 
\[\cS := \{S\in\mathcal{B}[0,1]:{\rm vol}(S)\leqslant c\},\]
where $\mathcal{B}[0,1]$ is the Borel sigma-algebra on $[0,1]$  and where 
\[{\rm vol}(S) := \int_{x\in S} {\rm d}x.\]
For each product $x\in [0,1]$, the marginal revenue that the retailer obtains if $x$ is purchased is denoted by $w(x)$; no revenue is obtained from a no-purchase. We assume that $w$ is a  continuously differentiable  function $[0,1] \rightarrow [0,1]$ with positive derivative bounded away from zero. It is worth observing that, in case $x$ is a measure of quality, it is natural to assume that $w$ is increasing. 

For all $S\in \cS $, we let $X^S$ denote the random choice of an arbitrary customer who is offered assortment $S$. We assume the following structure on the distribution of $X^S$: 
\begin{align} \label{eq:prob-X-in-A}
  \P(X^S\in A) = \frac{\int_{x\in A}v(x){\rm d}x}{1+\int_{x\in S}v(x){\rm d}x}, 
\end{align}
for all (Borel measurable) $A\subseteq S$, and 
\begin{align*}
\P(X^S = \emptyset) = \frac{1}{1+\int_{x\in S}v(x){\rm d}x},
\end{align*}
where $v:[0,1]\to\mathbb{R}_+$ is an integrable function. The function $v$ is referred to as the \emph{preference function}, and is \emph{unknown} to the seller. The expected revenue earned by the seller after offering assortment $S \in \cS$ to a customer is denoted by
\[r(S, v) := \frac{\int_{x\in S}v(x)w(x){\rm d}x}{1+\int_{x\in S}v(x){\rm d}x}.\]
The aim of the seller is determining an assortment $S \in \cS$ that maximizes $r(S, v)$. This is not directly possible, however, since the preference function is unknown.
We therefore consider a sequential version of the problem that enables the seller to learn the optimal assortment from accumulating sales data. The seller offers assortments during $T \in \N$ consecutive time periods, indexed by $t=1, \ldots, T$. Each time period $t$ corresponds to a visit of a single customer. The assortment offered at time $t$ is denoted by $S_t$, while $X_t \in \mathcal{X}$ denotes the (no-)purchase of the customer at time $t$. Conditionally on $S_t = S$, the purchase $X_t$ is distributed as $X^{S}$, for all $S \in \cS$ and all $t = 1, \ldots, T$. 

The seller's decisions which assortments to offer are described by her \textit{policy}: a sequence of mappings from available sales data (consisting of previously offered assortments and corresponding (no-)purchases) to a new assortment. Formally, a policy ${\pi} = (\pi_1,\ldots,\pi_T)$ is a vector of mappings $\pi_t: (\cS \times \mathcal{X})^{t-1}\to \cS$, such that 
\begin{align} \label{eq:use-a-policy}
 S_t = \pi_t(S_1, X_1, \ldots, S_{t-1}, X_{t-1}) \:\:\:
 \text{ for all } t=1, \ldots, T; 
\end{align}
here, we write $S_1 = \pi_1(\emptyset)$ for the initial assortment. Thus, a policy describes for each possible data-set of assortments and purchases how the seller selects the next assortment. The performance of a policy is measured by its \textit{regret}: the cumulative expected loss caused by using sub-optimal assortments. Formally, the regret of a policy $\pi$ is defined as
\begin{align} \label{eq:regret-definition}
\Delta_{\pi}(T, v) := \sum_{t=1}^T \mathbb{E}_\pi \left[ \max_{S \in \cS} r(S, v) - r(S_t, v) \right],   
\end{align}
where $S_1,\ldots,S_T$ satisfy \eqref{eq:use-a-policy}, and where the subscript in the expectation operator indicates the dependence on the policy $\pi$. In the next sections we show that the maximum in \eqref{eq:regret-definition} is attained. 
We also consider the \textit{worst-case} regret over a class $\cV$ of preference functions:
\begin{align*}
\Delta_{\pi}(T) := \sup_{v \in \cV} \Delta_{\pi}(T, v).
\end{align*}
The class of preference functions $\cV$ under consideration consists of all functions $v$ defined on the unit interval that satisfy the following assumptions.
\begin{assumption}\label{A1}
(i) For all $v \in \cV$ and $y \in [0,1]$,
\[ \underline{v} \leqslant v(y) \leqslant \vmax, \]
for some $\vmax > \underline{v} > 0$ with 
$\vmax > {w(0)}/{\int_0^1 (w(x)-w(0)) \dx}$.

\noindent
(ii) All $v \in \cV$ are differentiable on $(0,1)$ with uniformly bounded derivative, i.e., 
\[\sup_{y \in (0,1), v \in \cV} |v'(y)| < \infty.\]
\end{assumption}
These assumptions are arguably mild, and allow us to obtain instance-independent regret upper bounds. If  one is only interested in an instance-dependent bound of the form $\Delta_{\pi}(T,v) \leqslant C \log T$, where $C$ may depend on $v$, then Assumption 1(ii) can be weakened; see Remark~\ref{rem:relax-assumptions-c-is-1} for details. 
The assumption $\vmax > {w(0)}/{\int_0^1 (w(x)-w(0)) \dx}$ is used in Section \ref{sec:uncapacitated} to exclude trivialities; without this assumption, the unit interval $[0,1]$ is an optimal assortment for all $v \in \cV$ (in case $c=1$), and there is nothing to learn.



\begin{remark} \label{rem:need-for-a-model}
It is worth emphasizing that without assuming a particular structure of the choice probabilities $\P(X^S\in A)$, learning the optimal assortment from data is hopeless since the action space is uncountable. Our proposed model is motivated by its similarity to the well-known and frequently used \emph{discrete} multinomial logit (MNL) choice model. In this model, the probability that a customer's choice lies in $A \subseteq S$ when being offered assortment $S$ is equal to $\sum_{x \in A} v(x) / (1 + \sum_{x \in S} v(x))$, for a function $v$ defined on the product space and taking values in $(0, \infty)$. We essentially assume the same probabilistic structure, but with sums replaced by integrals.
\end{remark}
\begin{remark}\label{rem:utilitybase}
The discrete multinomial logit choice model can be derived from an assumed underlying random utility model in which a customer assigns utility $u(x) = \log(v(x)) + \epsilon(x)$ to each product $x$ and utility $\epsilon(0)$ to the no-purchase option; here $\{ \epsilon(x)\}$ and $\epsilon(0)$ are i.i.d.\ standard Gumbel distributed random variables. If the customer selects the product (or no-purchase option) that maximizes her utility, then the probability that her choice lies in $A \subseteq S$ when being offered assortment $S$ has a closed form and is equal to the above mentioned expression $\sum_{x \in A} v(x) / (1 + \sum_{x \in S} v(x))$ \citep[see][for a derivation]{Train2009}. Whether a similar relation between choice probabilities and an underlying choice model exists when the product space is the continuum is not known. With uncountably many products, the arguments from the discrete case do not carry over, as one, e.g.,\ would need to take a maximum over uncountably many random variables. Investigating the relation between choice probabilities and random utility models in case of a continuum of products is an interesting problem in its own right, but is outside the scope of the current paper. 
That said, our continuous model is closely connected to the discrete variant: it arises as a limit of discrete MNL models with the number of products $N$ going to infinity, and, conversely, discretizing the continuum product space generates choice probabilities that are described by a discrete MNL model (see Appendix C for details). Furthermore, the policy that we propose in Section \ref{sec:capacitated} to learn the optimal assortment with capacity constraint is effectively based on the fact that the continuous model can be approximated up to arbitrary precision by a discrete model.
\end{remark}

\section{Uncapacitated continuous assortment optimization}
\label{sec:uncapacitated}

In this section we investigate the uncapacitated case $c=1$, in which the assortment can in principle cover the full interval $[0,1]$.
Our main finding is that the optimal asymptotic growth rate of regret is logarithmic in the time horizon. In what follows, we first show how to compute an optimal assortment.  Next, we construct a policy and show that its  regret is bounded from above by $\Cmax\log T$ for some positive $\Cmax$ independent of $T$. Then we show that for \emph{any} policy $\pi$ the regret majorizes $\Cmin\log T$ for some $\Cmin>0$ independent of $T$. This implies that our constructed policy achieves the smallest possible growth rate of regret, and is therefore  asymptotically optimal.

The intuitive ideas underlying the mathematical statements in this section are given in the main text; the full proofs are contained in Appendix A.

\subsection{Full information optimal solution} \label{subsec:full-info-c-is-1}
It is known that the optimal assortment under the discrete MNL model without capacity constraints is of the form `offer the $k$ most expensive products' for some integer $k$ \citep[cf.][ Proposition 6]{Talluri2004}. This result carries over to our model of continuous assortment optimization. Since we assume that products are labeled in such a way that $w$ is increasing, the optimal assortment is of the form $[y,1]$, for some $y \in [0,1]$. The argument to show this is as follows \cite[cf.][Section 2.1]{Rusmevichientong2010a}:
\begin{align}
\max\{ r(S, v) : S \in \cS\} &= 
\max\left\{\varrho\in[0,1]: \exists S\in \cS : r(S,v)\geqslant \varrho\right\} \notag \\
&= \max\left\{\varrho\in[0,1]: \exists S\in \cS : \int_{S} v(x)\big(w(x)-\varrho\big){\rm d}x\geqslant \varrho\right\} \notag \\
&= \max\left\{\varrho\in[0,1]: \max_{S \in \cS} \int_{S} v(x)\big(w(x)-\varrho\big){\rm d}x\geqslant \varrho\right\}. 
\label{eq:rewrittenSb} 
\end{align}
The inner maximization problem in \eqref{eq:rewrittenSb} is maximized by $\{ x \in [0,1] : w(x) \geqslant \varrho\}$. Let $w^{-1}(\cdot)$ denote the generalized inverse of $w(\cdot)$, i.e.,
\[w^{-1}(\varrho) := \min\{x\in[0,1]:w(x)\geqslant \varrho\}, \qquad \varrho \in [0,1]. \]
Since $w$ is strictly increasing and continuous, 
the set
$\{ x \in [0,1] : w(x) \geqslant \varrho\}$ is equal to the interval $[w^{-1}(\varrho), 1]$, and it follows that
\begin{align*}
\max\{ r(S, v) : S \in \cS\} =
\max\{ r( [w^{-1}(\varrho),1],v) : \varrho \in [0,1] \}.
\end{align*}
The fact that the optimal assortment is an interval of the form $[y,1]$ has evident attractive computational implications, most notably that it reduces the original optimization problem over all subsets of the unit interval to an optimization problem in one variable $y \in [0,1]$. 

\subsection{A policy for incomplete information}
We proceed by defining a data-driven policy that iteratively approximates the optimal assortment. The policy is parameterized by $\alpha > 0$ and $\beta \geqslant 0.$

\vspace{\baselineskip}
\noindent\HRule
\begin{center}
{\tt Stochastic Approximation Policy SAP}$(\alpha, \beta)$
\end{center}
{\tt 1. Initialization.} {\tt Let $\alpha > 0$, $\beta \geqslant 0$ and $\varrho_1 \in[0,1]$. For all $t \in \N$ let $a_t := \alpha / (t + \beta)$. 
} {\tt Put $t:=1$. Go to 2.} \\
{\tt 2. Assortment selection.} {\tt Let} 
\begin{align*}
 S_t := [w^{-1}(\varrho_t), 1], \:\:\:\: R_t := w(X_t) {\boldsymbol 1}\{X_t \in S_t\},
\end{align*}
{\tt and}
\[\varrho_{t+1} = \varrho_t + a_t\big(R_t - \varrho_t\big).\]
{\tt Put $t:=t+1$. If $t\leqslant T$, then go to 2, else to 3.}\\
{\tt 3. Terminate.}

\noindent\HRule

The policy SAP$(\alpha, \beta)$ is a classic stochastic approximation policy \citep{Robbins1951,Kushner1997} that aims at finding the value of $\varrho \in [0,1]$ such that $r( [w^{-1}(\varrho),1], v)$ equals $\varrho$. This condition uniquely defines the optimal $\varrho$ that corresponds to the optimal assortment $[w^{-1}(\varrho), 1]$. Since only noisy observations $R_t$ of the revenue function $r([w^{-1}(\varrho),1],v)$ are available, the policy keeps changing $\varrho_t$ based on observations of $R_t - \varrho_t$. The step sizes $a_t$ decay roughly as $1/t$; this rate ensures that, on the one hand, $\varrho_t$ does not converge `too slowly' to the optimal value, while on the other hand, $\varrho_t$ does not keep jumping `over' the optimal $\varrho$ which could potentially lead to a slow convergence rate. 

\subsection{Regret upper bound}
\label{sec:upperbounduncapacitated}

We proceed by showing that the worst-case regret of SAP($\alpha,\beta)$ grows at most logarithmically in $T$. 

\begin{theorem}\label{th1}
Let $\pi$ correspond to {\rm SAP(}$\alpha,\beta{\rm )}$ with $\alpha \geqslant \overline{v} + 1$ and $\beta\geqslant \max\{0,\alpha-1\}$. 
Then there is a $\Cmax > 0$ such that, for all $T \geqslant 2$, 
\[ \Delta_{\pi}(T) \leqslant \Cmax \log T. \]
\end{theorem}

Write $g(y) := r([y,1],v)$ and $h(\varrho) := g(w^{-1}(\varrho))$, for $y,\varrho \in [0,1]$. The key idea underlying the algorithm and the regret upper bound is the observation that the optimal expected revenue
\[\varrho^*:= \max\{r(S,v):S\in\cS\},\]
solves the fixed-point equation
\[h(\varrho) = \varrho.\]
Because the noisy observation $R_t$ has conditional expected value $h(\varrho_t)$, we can apply a Robbins-Monro scheme to find $\varrho^*$ and the corresponding optimal assortment, without, e.g., having to estimate the gradient of the revenue function. This explains why we achieve a small regret rate of $\mathcal{O}(\log T)$ instead of, e.g., $\mathcal{O}(\sqrt{T})$ which is commonly seen in continuous multi-armed bandit problems.


\begin{remark}
The logarithmic growth rate of the regret in Theorem \ref{th1} holds for all choices of $\alpha \geqslant \overline{v} + 1$ and $\beta \geqslant \max\{0,\alpha-1\}$. As the constant in front of the $\log T$ term may depend on these parameters, the finite-time performance of the policy may be fine-tuned by carefully selecting these $\alpha$ and $\beta$, for example based on initial simulations. 
\end{remark}

\begin{remark}\label{rem:relax-assumptions-c-is-1}
Theorem \ref{th1} presents a \emph{worst-case} bound: the constant $\Cmax$ is independent of $v \in \cV$. To obtain this result we need to impose assumptions on uniform bounds on the derivative of $v \in \cV$. If we are only interested in an \emph{instance-dependent} upper bound $\Delta_{\pi}(T, v) \leqslant C_v \log T$, for all $v \in \cV$ and some $v$-dependent constant $C_v > 0$, then Assumption \ref{A1}(ii) can be relaxed to $v$ being continuously differentiable: this ensures inequality \eqref{eq:k3} in the proof of Lemma \ref{lem:properties-of-g}. 
\end{remark}

\subsection{Regret lower bound}
\label{sec:uncapacitatedlowerbound}

Now that we have proven an upper bound on the regret of the policy SAP($\alpha,\beta)$, we proceed by showing that this bound is, up to a multiplicative constant, asymptotically tight as $T$ grows large. This implies that our policy is asymptotically optimal. 
\begin{theorem} \label{thm:lower-bound-c-is-1}
There is a $\Cmin > 0$ such that, for all policies $\pi$ and all $T \geqslant 2$, 
\[ \Delta_{\pi}(T) \geqslant \Cmin \log T. \]
\end{theorem}

To prove Theorem \ref{thm:lower-bound-c-is-1} we first define a collection of preference functions $v_{\theta}$, indexed by a parameter $\theta$ that takes values in a closed interval $\Theta$. Next, we show that the instantaneous regret incurred by offering assortment $S$ instead of the optimal assortment $[y(\theta), 1]$ corresponding to $\theta$, is bounded from below by a constant times the squared difference between the volumes of $[y(\theta),1]$ and $S$, for any $S \in \cS$ and $\theta \in \Theta$. This result is obtained by exploiting local quadratic behavior of the instantaneous regret for assortments close to the optimal one. Furthermore, this relation implies that it suffices to prove a lower bound on the mean squared error of any estimate of the \emph{volume} of the optimal assortment: a reduction from subsets of $[0,1]$ to one-dimensional variables in $[0,1]$. To mitigate difficulties with the atom of the purchase distributions $X^S$ on $\emptyset$, we define new, absolutely continuous random variables $Z_1,Z_2, \ldots$ and show that it suffices to prove a regret lower bound based on observations $Z_1, Z_2, \ldots$ instead of the purchases $X_1, X_2, \ldots$. Next, we bound the Fisher information corresponding to $Z_1,\ldots, Z_t$ from above by a positive constant times $t$, and define a probability measure $\lambda$ on the support of $\theta$. By the Van Trees inequality \citep{Gill1995}, we then conclude that the expected instantaneous regret in period $t+1$, where the expectation is with respect to $\lambda$, is bounded from below by a constant times $1/t$, for all $t$. By summing over all $t=1, \ldots, T$, the logarithmic lower bound follows. 


\section{Capacitated continuous assortment optimization}
\label{sec:capacitated}

In this section we consider the setting in which the capacity $c$ is strictly less than $1$. We first characterize the optimal assortment under full information, and show that the optimal solutions in the capacitated case exhibit richer behavior than the intervals $[y ,1]$ observed in the uncapacitated case. Next, we show that this structural difference translates into a different complexity of the dynamic learning problem, finding that the optimal growth rate of regret behaves as $T^{2/3}$ instead of $\log T$ as established in the previous section. 

The intuitive ideas underlying the mathematical statements in this section are given in the main text; the full proofs are contained in Appendix B.

\subsection{Full information optimal solution}
\label{sec:capacitatedfullinfo}
As shown in Section \ref{subsec:full-info-c-is-1}, the assortment optimization problem under full information can be written as 
\begin{align}
\max\{ r(S, v) : S \in \cS\} 
&= \max\left\{\varrho\in[0,1]: \max_{S \in \cS} \mathcal{I}(S,\varrho)\geqslant \varrho\right\}, 
\label{eq:rewrittenSbb} 
\end{align}
where 
\begin{equation*}
  \mathcal{I}(S,\varrho) := \int_S v(x)\big(w(x)-\varrho\big){\rm d}x,
\end{equation*}
for $S\in \cS$ and $\varrho\in[0,1]$, and where $\cS$ denotes the collection of all measurable subsets of the unit interval with volume at most $c$. Without a capacity constraint, $\mathcal{I}(S,\varrho)$ is maximized by the upper level set 
\[W_{\varrho} := \{ x \in [0,1] : w(x) \geqslant \varrho\},\]
for all $\varrho \in [0,1]$, since $v(x) (w(x) - \varrho)$ is nonnegative if and only if $x \in W_{\varrho}$. 
With capacity constraint, however, the optimization becomes slightly more subtle, because the set $W_{\varrho}$
 may have volume larger than $c$. 
We discuss how to solve the inner maximization problem in \eqref{eq:rewrittenSbb}, i.e.,\ how to construct an $S_{\varrho}$, for each $\varrho \in [0,1]$, such that
\begin{equation}
  \mathcal{I}(S_\varrho,\varrho) = \max\{\mathcal{I}(S,\varrho):S\in\cS\}.
  \label{eq:innermaximization}
\end{equation}
Next, we utilize this result to obtain an optimal solution for \eqref{eq:rewrittenSbb}. 
To this end, let
\begin{align} \label{eq:def-of-h}
h(x,\varrho):=v(x)\big(w(x)-\varrho\big), \quad x \in [0,1], \varrho \in [0,1],    
\end{align}
be the function that $\mathcal{I}$ integrates, let
\begin{align*}
L_{\varrho}(\ell):= \{x\in[0,1]: h(x,\varrho) \geqslant \ell\}, \qquad& \varrho\in[0,1],\:\ell\in[0,\infty),
\end{align*}
be the upper level sets of $h(\,\cdot\,, \varrho)$, and let 
\begin{align*}
m_\varrho(\ell):=\text{vol}(L_{\varrho}(\ell)), \qquad& \varrho\in[0,1],\:\ell\in[0,\infty),
\end{align*}
denote their volume. We first give an explicit characterization of the optimal solution(s) of \eqref{eq:innermaximization}. 

\begin{lemma} \label{lem:perrho}
Let $\varrho\in[0,1]$. 
\begin{itemize}
\item[\rm ($i$)] If ${\rm vol}(W_\varrho)\leqslant c$, then the maximum of $\mathcal{I}(S,\varrho)$ over sets in $\cS$ is attained by $S= W_\varrho.$ 
\item[\rm ($ii$)] If ${\rm vol}(W_\varrho)> c$, then the maximum
\[\ell_\varrho:= \max\{\ell\geqslant 0: m_\varrho(\ell)\geqslant c\}\]
exists, and the maximum of $\mathcal{I}(S,\varrho)$ over sets in $\cS$ is attained by $S=L_\varrho^+
\cup L_\varrho^\circlearrowleft$,
where
\[L_\varrho^+:=\{x\in[0,1]: h(x,\varrho) >\ell_\varrho\},\]
\[L_\varrho^=:=\{x\in[0,1]: h(x,\varrho) =\ell_\varrho\},\]
and $L_\varrho^\circlearrowleft$ is a subset of $L_\varrho^=$ such that ${\rm vol}(S) = {\rm vol}(L_\varrho^+)+ {\rm vol}(L_\varrho^\circlearrowleft)=c.$
\end{itemize}
\end{lemma}
As is intuitive, the upper level set $W_{\varrho}$ maximizes $\mathcal{I}(S,\varrho)$ with respect to $S$ if this does not result in a violation of the capacity constraint (case ($i$)). On the other hand, if the volume of $W_{\varrho}$ exceeds the maximum capacity (case ($ii$)), then we construct an optimal assortment as follows. First, we `fill' the assortment by the upper level set $\{ x \in [0,1] : h(x, \varrho) > \ell\}$, where $\ell$ is as large as possible given the capacity constraint; this largest value of $\ell$ is denoted by $\ell_{\varrho}$ in Lemma \ref{lem:perrho}. If the resulting assortment has size $c$ then we are done; if not, then the function $h(x,\varrho)$ has `flat' regions; that is, the level set $\{x \in [0,1] : h(x, \varrho) = \ell_{\varrho}\}$ has positive measure, and adding this set to the assortment would result in a violation of the capacity constraint. In that case, the optimal assortment $S$ constructed in Lemma \ref{lem:perrho} consists of 
$\{x \in [0,1] : h(x, \varrho) > \ell_{\varrho}\}$ and a subset of $\{x \in [0,1] : h(x, \varrho) = \ell_{\varrho}\}$, such that the volume of the union of the two parts is exactly equal to $c$. 

Based on the explicit solution of the inner maximization problem \eqref{eq:innermaximization} given in Lemma \ref{lem:perrho}, we now characterize an optimal solution to \eqref{eq:rewrittenSbb}.
\begin{proposition}\label{prop:fullinformationcapacitated}
For each $\varrho\in[0,1]$ let $S_\varrho\in\cS$ satisfy \eqref{eq:innermaximization}. Then there is a unique solution $\varrho^* \in [0,1]$ to the fixed-point equation 
\[ \mathcal{I}(S_\varrho,\varrho) = \varrho, \quad \varrho \in [0,1], \]
and $S_{\varrho^*}$ is an optimal assortment:
\[r(S_{\varrho^*},v) = \max\{r(S,v):S\in\cS\}.\]
\end{proposition}
We prove the proposition by showing that $
\mathcal{I}(S_\varrho,\varrho)$ is continuous and non-increasing as function of $\varrho$, with $\mathcal{I}(S_0,0) \geqslant 0$ and $\mathcal{I}(S_1,1) = 0$. By the equality \eqref{eq:rewrittenSbb} and the observation
\[\mathcal{I}(S_\varrho,\varrho) = \varrho \quad\Longleftrightarrow\quad r(S_\varrho,v) = \varrho,\]
we conclude that if $\varrho^*$ solves the fixed-point equation, then $S_{\varrho^*}$ is an optimal assortment.

\begin{remark}
The optimal assortment can be efficiently computed up to any desired accuracy via a bisection method. In Appendix D we present an implementation of such a bisection algorithm. 
\end{remark}

\begin{remark}\label{rem:interval} In contrast to the setting discussed in Section \ref{sec:uncapacitated}, the optimal assortment in the presence of a capacity constraint does not have to be a connected interval. Consider, for example, the bi-modal preference function plotted in the left-hand panel of Figure \ref{fig:simulationv}, and let $c = 0.5$ and $w(x) = x$ for all $x \in [0,1]$. The optimal assortment $S^*$ in this instance consists of the union of two disjoint intervals: 
\[S^* = [0.33, 0.48]\cup
 [0.63, 0.98], \]
 with corresponding optimal expected profit
 $r(S^*,v) = 0.19$. 
 In contrast, the largest expected profit that can be obtained from a single closed interval in this instance is equal to 0.13 (attained at the interval $[0.5, 1]$); a reduction in profit of more than thirty percent. This shows that restricting to single intervals can leave a significant amount of profit on the table. 
\end{remark}

\begin{figure}[H]
\begin{minipage}{\textwidth}
\centering
\begin{tikzpicture}
\begin{axis}[
  title={},
  width = 0.4*\linewidth,
  xlabel={$x\vphantom{\varrho}$},
  ylabel={$v(x)$},
  xmin=0, xmax=1,
  ymin=0, ymax=1.6,
  legend pos=north west,
]

\addplot[
  thin,
  color=black,
  ]
  table {Simulation_v.dat};

\end{axis}
\end{tikzpicture}
\begin{tikzpicture}
\begin{axis}[
  title={},
  width = 0.4*\linewidth,
  xlabel={$\varrho$},
  ylabel={$\mathcal{I}(S_\varrho,\varrho)$},
  xmin=0, xmax=1,
  ymin=0, ymax=1,
  legend pos=north west,
]

\addplot[
  thin,
  color=black,
  ]
  table {IrhoSrho.dat};

\addplot[
  thin,
  color=gray,
  ]
  {x};

\end{axis}
\end{tikzpicture}
\end{minipage}
\caption{The left-hand panel shows the bi-modal preference function $v(x) = \tfrac{1}{10} + \tfrac{1}{5} (2+x)(1-x) + \tfrac{2}{7}\phi(x;0.33,0.1) + \tfrac{1}{5}\phi(x;0.8,0.1)$, $x \in [0,1]$, where $\phi(\,\cdot\,;\mu,\sigma)$ is the  normal probability density function with parameters $\mu$ and $\sigma$. The right-hand panel shows the corresponding function $\varrho \mapsto \mathcal{I}(S_\varrho,\varrho)$. The optimal $\varrho^* = 0.19$ is the unique $\varrho$ such that $\mathcal{I}(S_\varrho,\varrho)$ is equal to $\varrho$.}
\label{fig:simulationv}
\end{figure}

The continuous model offers insight in the role of the capacity constraint in its discrete counterpart. To illustrate this, consider the instance of the discrete MNL assortment optimization problem discussed by \cite{Rusmevichientong2010a} with $N=4$ products, and preference values $v_i$ and marginal revenues $w_i$ given by 
\[{\boldsymbol v} = (0.2,0.6,0.3,5.2)\qquad \text{and}\qquad{\boldsymbol w} = (9.5,9.0,7.0,4.5).\]
\cite{Rusmevichientong2010a} 
shows that the optimal assortment, as function of the maximum assortment size $C$, is given by 
\[\begin{array}{lcccc}
\hline
C &1 &2 &3 &4 \displaystyle\vphantom{\bigcup}\\
\hline
\text{Optimal assortment} &\{4\} &\{2,4\} &\{1,2,3\} &\{1,2,3,4\} \displaystyle\vphantom{\bigcup}\\
\hline
\end{array}\]
By defining
\[ v(x) = N \sum_{i=1}^N v_i {\bf 1}\left\{ \frac{i-1}{N} \leqslant x < \frac{i}{N} \right\},  \]
and 
\[ w(x) = \sum_{i=1}^N w_i {\bf 1}\left\{ \frac{i-1}{N} \leqslant x < \frac{i}{N} \right\}, \]
for all $x \in [0,1]$, we translate the problem into our continuous assortment optimization setting. 
For each fixed $\varrho$, the function $x \mapsto h(x, \varrho)$ defined in \eqref{eq:def-of-h} is a piece-wise constant function that attains the values $N v_i(w_i-\varrho)$, for $i=1,\ldots,N$. The ordering of the quantities $\{ N v_i (w_i - \varrho) : i=1,\ldots, N\}$ does not change when $\varrho$ is slightly changed, except possibly if $\varrho$ is of the form \[ 
\varrho_{i,j} := \frac{v_iw_i - v_jw_j}{v_i-v_j},\qquad \text{ for some } 1 \leq i < j \leq N.\]
If we consider the optimal revenue $\varrho^*(c)$ as function of the capacity constraint $c$, then it follows that 
the fraction of a product that is included in the optimal assortment might be discontinuous at points $c$ such that $\varrho^*(c) = \varrho_{i,j}$, for some $i,j$. In our example, this happens at $c \approx 0.32$, $c \approx 0.61$, and $c \approx 0.66$. Figure \ref{fig:revisitpaat} illustrates this behavior. 
The fraction of a particular product that is included in the optimal assortment is not monotone in $c$, and can in fact make jumps. 

 
\newpage\null
\begin{figure}[H]
\begin{center}
\begin{minipage}{\textwidth}
\centering
\begin{tikzpicture}
\begin{axis}[
    title={},
    width = 0.45*\linewidth,
    xlabel={$c$},
    ylabel={Fraction of offered product},
    xmin=0, xmax=1,
    ymin=0, ymax=1,
    legend pos=north west,
]

\addplot[
      name path = l1,
    thin,
    color=black,
    forget plot,
    ]
    table {resall1.dat};

\addplot[
      name path = l2,
    thin,
    color=black,
    forget plot,
    ]
    table {resall2.dat};

\addplot[
      name path = l3,
    thin,
    color=black,
    forget plot,
    ]
    table {resall3.dat};

\addplot[
      name path = l4,
    thin,
    color=black,
      forget plot,
    ]
    table {resall4.dat};
    
\addplot[
    name path = zero,
    draw = none,
    forget plot,
    ]
    {0};

\addplot[pattern=dots, pattern color=black!100]fill between[of=l1 and zero, soft clip={domain=0:1}];

\addplot[pattern=grid, pattern color=black!100]fill between[of=l2 and l1, soft clip={domain=0:1}];

\addplot[pattern=north west lines, pattern color=black!100]fill between[of=l3 and l2, soft clip={domain=0:1}];

\addplot[pattern=crosshatch, pattern color=black!100, area legend]fill between[of=l4 and l3, soft clip={domain=0:1}];

\addlegendentry{Product 1};
\addlegendentry{Product 2};
\addlegendentry{Product 3};
\addlegendentry{Product 4};

\end{axis}
\end{tikzpicture}~
\begin{tikzpicture}
\begin{axis}[
    title={},
    width = 0.45*\linewidth,
    xlabel={$c$},
    ylabel={$\varrho^*(c)$},
    xmin=0, xmax=1,
    ymin=3.5,
    legend pos=north west,
]

\addplot[
    thin,
    color=black,
    ]
    table {rhostarofc.dat};

\addplot[
	smooth,
  dashed,
  samples=10,
  domain=0:1,
  ]
  {2};

\addplot +[mark=none,black,dashed] coordinates {(0.0384,0) (0.0384,2)};

\addplot[
	smooth,
  dashed,
  samples=10,
  domain=0:1,
  ]
  {4.3};

\addplot +[mark=none,dashed,black] coordinates {(0.61,0) (0.61,4.3)};

\addplot[
	smooth,
  dashed,
  samples=10,
  domain=0:1,
  ]
  {3.913043478260870};

\addplot +[mark=none,dashed,black] coordinates {(0.32,0) (0.32,3.913043478260870)};

\addplot[
	smooth,
  dashed,
  samples=10,
  domain=0:1,
  ]
  {4.346938775510204};

\addplot +[mark=none,dashed,black] coordinates {(0.66,0) (0.66,4.346938775510204)};

\node at (axis cs: 0.93,3.963043478260870) {\small $\varrho_{2,4}$};
\node at (axis cs: 0.93,4.246938775510204) {\small $\varrho_{3,4}$};
\node at (axis cs: 0.93,4.4) {\small $\varrho_{1,4}$};

\end{axis}
\end{tikzpicture}
\end{minipage}
\caption{The left-hand panel shows the optimal amount of each  products, as function of $c$. The right-hand panel shows the corresponding optimal expected profit $\varrho^*(c)$.}
\label{fig:revisitpaat}
\end{center}
\end{figure}

\subsection{A policy for incomplete information}
\label{sec:capacitatedpolicy}

We proceed by presenting a policy for the continuous assortment optimization problem with capacity constraint and incomplete information. We first discuss the underlying intuition and the method of establishing upper confidence bounds, after which we formally present our policy Discretized Upper Confidence Bounds (DUCB). In what follows,  we use $[n]$ as a compact notation for the set $\{1,\ldots,n\}$ (where $n\in\N$).

The proposed policy is parameterized by an integer $N\in\N$. The policy DUCB($N$) discretizes the set of products $[0,1]$ into $N$ bins of equal size, after which the policy exploits the similarity with the discrete multinomial logit (MNL) model. This is done by applying the UCB policy from \cite[][(Algorithm 1)]{Agrawal2019} to the bin structure. We regard a continuous purchase in the $i$-th bin as a purchase of product $i$ in the discrete MNL model. The policy establishes upper confidence bounds on the preference parameters corresponding to the discrete MNL model. More specifically, define the bins as
\begin{equation}
B_i:=\left[\frac{i-1}{N},\frac{i}{N}\right)
\label{eq:binsuptoN}
\end{equation}
for $i=1,\ldots,N-1$ and
\begin{equation}
B_N:=\left[\frac{N-1}{N},1\right],
\label{eq:binN}
\end{equation}
and define the parameters
\[v_i := \int_{B_i}v(x)\,\d x\qquad\text{and}\qquad w_i := N\int_{B_i}w(x)\,\d x,\qquad i\in [N].\]
Note that by our choice of $v_i$ and $w_i$ for $i\in[N]$, the expected profit of an assortment consisting of a collection of bins is the same for the continuous and discrete MNL model. 

At each time $t$ we observe a purchase $X_t \in S_t\cup\{\emptyset\}$, and translate this $X_t$ to a discrete purchase $Y_t$ by
\[Y_t := \sum_{i=1}^N i \mathbf{1}_{B_i}(X_t).\]
Observe that $X_t\in B_{Y_t}$ if $X_t\in S_t$ and $Y_t=0$ if $X_t=\emptyset$.
The policy at time $t$ computes upper confidence parameters $v^{\rm UCB}_{1,t},\ldots,v^{\rm UCB}_{N,t}$ of the parameters $v_1,\ldots,v_N$ using observed discrete purchases $Y_1,\ldots,Y_t$. In the next step, at time $t+1$, the chosen assortment is the collection of bins $S_{t+1}=\bigcup_{i\in D_{t+1}}B_i$ where $D_{t+1}$ is a subset of $[N]$ of size at most $\floor{cN}$, which maximizes
\[D\mapsto\frac{\sum_{i\in D}v^{\rm UCB}_{i,t}w_i}{1+\sum_{i\in D}v^{\rm UCB}_{i,t}}.\]
If such an optimal assortment is not unique, ties are broken by applying an arbitrary fixed ordering of assortments.

The DUCB($N$) policy starts by setting $v^{\rm UCB}_{i,0}=1$ for all $i\in[N]$. To compute the upper confidence parameters $v_{1,t}^{\rm UCB},\ldots,v_{N,t}^{\rm UCB}$ for $t=1,\ldots,T$, the observed discrete purchases $Y_1,\ldots,Y_t$ are used as follows. The time horizon is partitioned into epochs, where each epoch corresponds to a sequence of consecutive actual purchases. An epoch ends when a no-purchase is observed, i.e., $X_t=\emptyset$ or, equivalently, $Y_t=0$. Specifically, let $t_0:=0$ and recursively define
\[t_\ell := \min\{t\in \{t_{\ell-1}+1,\ldots,T\}:Y_t=0\},\qquad \ell\in\N_{\geqslant 1},\]
and $t_\ell:=T$ if $\{t\in \{t_{\ell-1}+1, \ldots, T\}:Y_t=0\} = \emptyset$. Let $L$ denote the first index such that $t_L=T$, that is, 
\[L := \min\{\ell\in\N_{\geqslant 1}:t_\ell = T\}.\]
Then the $\ell$-th epoch $\mathcal{E}_\ell$ is defined as
\[\mathcal{E}_\ell:=\{t_{\ell-1}+1,\ldots,t_\ell\}, \qquad \ell\in[L].\]
Within each epoch $\mathcal{E}_\ell$ the upper confidence parameters remain unchanged, that is, $v_{i,t}^{\rm UCB} = v_{i,s}^{\rm UCB}$ for all $i\in[N]$ when $s,t\in\{t_{\ell-1},\ldots,t_{\ell}-1\}$. As a result, and by the fixed tie-breaking rule, $D_t$ remains the same within each epoch. Define $D^\ell:= D_{t_{\ell-1}+1}$. At the end of an epoch, the upper confidence parameters are updated. Then the upper confidence bounds $v_{1,t}^{\rm UCB},\ldots,v_{N,t}^{\rm UCB}$ become
\begin{equation}
v_{i,t}^{\rm UCB} :=\left\{\begin{array}{ll}
    \ds\bar{v}_{i,\ell} + \sqrt{\bar{v}_{i,\ell}\frac{48\,\log(\sqrt{N}\ell+1)}{|\mathcal{T}_i(\ell)|}} + \frac{48\,\log(\sqrt{N}\ell+1)}{|\mathcal{T}_i(\ell)|} & \text{ if }t=t_\ell \text{ for some }\ell\in[L] \text{ and }i\in D^\ell, \\
    v_{i,t-1}^{\rm UCB} & \text{ otherwise.} 
\end{array}\right.
\label{eq:updatevUCB}
\end{equation}
Here $\mathcal{T}_i(\ell)$ is the set of epochs up to $\ell$ in which product $i$ is offered, that is,
\[\mathcal{T}_i(\ell):=\{\tau\in[\ell]:i\in D^\tau\} ,\qquad i\in[N],\]
and $\bar{v}_{i,\ell}$ is the average of the number of times product $i$ is purchased in epoch $\tau$ for epochs $\tau\in\mathcal{T}_i(\ell)$, that is,
\[\bar{v}_{i,\ell} :=\frac{1}{|\mathcal{T}_i(\ell)|}\sum_{\tau\in\mathcal{T}_i(\ell)} \sum_{t\in \mathcal{E}_\tau}\mathbf{1}\{Y_t = i\}.\]
For all $i\in D^\ell$, $\bar{v}_{i,\ell}$ is an unbiased estimator of the discrete preference parameters $v_i$ \citep[see Corollary A.1 by][]{Agrawal2019}. Note that in \eqref{eq:updatevUCB} there exists an $\ell\in[L]$ such that $t=t_\ell$ if and only if $Y_t=0$.

After the verbal description of our DUCB($N$) policy, we now present the formal algorithm.

\vspace{\baselineskip}
\noindent\HRule
\begin{center}
{\tt Discretized Upper Confidence Bounds DUCB($N$)}
\end{center}
{\tt 1. Initialization.} {\tt Let $N\in\N$ and put $K:=\floor{cN}$. Let $B_i$ for $i\in[N]$ be as in \eqref{eq:binsuptoN} and \eqref{eq:binN}. Let $w_i:=N\int_{B_i}w(x)\d x$ and $v_{i,0}^{\rm UCB}:=1$ for $i\in[N]$ and $t:=1$. Go to 2.}\\
{\tt 2. Assortment selection.} {\tt Let}
\begin{equation}
D_t \in \argmax_{D\subseteq[N]:|D|\leqslant K} \frac{\sum_{i\in D}v_{i,t-1}^{\rm UCB}w_i}{1+\sum_{i\in D}v_{i,t-1}^{\rm UCB}},
\label{eq:maxdiscreteassortment}
\end{equation}
{\tt and}
\[S_t := \bigcup_{i\in D_t} B_i.\]
{\tt Determine $v_{1,t}^{\rm UCB},\ldots,v_{N,t}^{\rm UCB}$ as in \eqref{eq:updatevUCB}, and let $t:=t+1$. If $t\leqslant T$, then go to 2, else to 3.}\\
{\tt 3. Terminate.}

\noindent\HRule

If the discrete assortment $D_t$ as in \eqref{eq:maxdiscreteassortment} is not unique, ties are dealt with by applying an arbitrary fixed ordering of assortments.

\subsection{Regret upper bound}
\label{sec:capacitatedupperbound}

We proceed by showing that the worst-case regret of DUCB$(N)$, with appropriately chosen $N$, grows at most as $T^{2/3}$ up to a logarithmic term.

\begin{theorem}\label{th:upperboundc<1}
Let $T \geqslant 2$, $\gamma = \max\{\vmax,1/c+1\}$, $N = \floor{\gamma T^{1/3}}$, and let  $\pi$ correspond to {\rm DUCB($N$)}. There is a $\Cmax > 0$, independent of $T$, such 
\[ \Delta_{\pi}(T) \leqslant \Cmax \,T^{2/3}(\log T)^{1/2}.\]
\end{theorem}

To prove the theorem, we first establish a relation between the regret in our model and that of the discrete regret in the context of \cite{Agrawal2019}. There is an obvious misalignment between those two notions: one deals with functions and the other with discrete parameters. However, we are able to bound the regret of DUCB($N$) from above by the regret of UCB plus a discretization error of order $T/N$. Since the regret of UCB is of order $\sqrt{NT}$ (up to a logarithmic term), the optimal value of $N$ is proportional to $T^{1/3}$ which results in a $T^{2/3}$ upper bound for the regret of DUCB($N$) (also up to a logarithmic term).

Then it is observed that the discretization error consists of three sources. The first source is due to the fact that the discrete model approximates the actual preference function and marginal profit function by a piecewise constant function.  
The second source is caused by the fact that the true optimal assortment is not 
necessarily exactly equal to a collection of bins. 
The third source is the effect of the misalignment between the regret within our model with that of the regret of UCB as analyzed by \cite{Agrawal2019}. When considering the regret of DUCB($N$), we need to take this translation error into account. 

To facilitate the analysis of the performance of DUCB($N$), we define 
\begin{align}\label{eq:checkv}
\check{v}(x) &:= N \sum_{i=1}^N \mathbf{1}_{B_i}(x)\int_{B_i}v(y)\,\d y,\qquad x\in [0,1],\\
\label{eq:checkw}
\check{w}(x) &:= N \sum_{i=1}^N \mathbf{1}_{B_i}(x)\int_{B_i}w(y)\,\d y,\qquad x\in [0,1].
\end{align}
In addition, we introduce an adjustment of the currently used notation of the expected profit of an assortment $S\in\cS$. We will explicitly denote that this expected profit depends on marginal profit function $w(x)$ as well as preference function $v(x)$:
\[r(S,v,w) := \frac{\int_S v(x)w(x)\,\d x}{1+\int_S v(x)\,\d x}.\]
The effect of the first component of the discretization error is captured by Proposition \ref{prop:bridge} below.

\begin{proposition}
Let $\check{v}$ and $\check{w}$ be as in \eqref{eq:checkv} and \eqref{eq:checkw}, respectively. Let $S^{*}$ and $\check{S}$ in $\cS$ be optimal assortments corresponding to $v$ and $w$, and $\check{v}$ and $\check{w}$, respectively, that is,
\begin{equation}
r(S^{*},v,w) = \max_{S\in\cS}r(S,v,w) \quad\text{and} \quad r(\check{S},\check{v},\check{w}) = \max_{S\in\cS} r(S,\check{v},\check{w}).
\label{eq:SstarandcheckS} 
\end{equation}
Then the difference between the expected revenue of $S^{*}$ under $v$ and $w$ and the expected revenue of $\check{S}$ under $\check{v}$ and $\check{w}$ is bounded from above by
\begin{equation}
r(S^{*},v,w)-r(\check{S},\check{v},\check{w}) \leqslant |\!|v-\check{v}|\!|_1 + \vmax\,|\!|w-\check{w}|\!|_1,
\label{eq:L1diff}
\end{equation}
where $|\!|\cdot|\!|_1:=\int_0^1|\cdot|\,{\rm d}x$.
\label{prop:bridge}
\end{proposition}
Note that the optimal assortment $\check{S}$ in the result stated above is the optimal assortment within $\cS$. The UCB algorithm only considers discrete assortments, which translates to a collection of bins within our model. The effect of this is stated in Lemma \ref{lem:optimizingoversubset} below.

\begin{lemma} \label{lem:optimizingoversubset}
Let $\check{v}$ and $\check{w}$ be as in \eqref{eq:checkv} and \eqref{eq:checkw}, respectively. Let $\mathcal{A}_K$ be the set of all collections of at most $K=\floor{cN}$ bins $B_i$, that is,
\begin{equation} \label{eq:setofbincollections}
\mathcal{A}_K:= \left\{\bigcup_{i\in D} B_i:D\subset[N]\:\text{ and }\,\: |D|\leqslant K\right\}.
\end{equation}
In addition, let $\check{S}$ in $\cS$ and $S^d$ in $\mathcal{A}_K$ be optimal assortments corresponding to $\check{v}$ and $\check{w}$, that is,
\begin{equation}
r(\check{S},\check{v},\check{w}) = \max_{S\in\cS} r(S,\check{v},\check{v})\quad\text{and}\quad r(S^d,\check{v},\check{w}) = \max_{S\in\mathcal{A}_K} r(S,\check{v},\check{v}).
\label{eq:checkSandSd}
\end{equation}
Then the difference between the expected revenue under $\check{v}$ and $\check{w}$ of $\check{S}$ and $S^d$ is bounded from above by
\[r(\check{S},\check{v},\check{w})-r(S^d,\check{v},\check{w}) \leqslant \frac{\vmax}{N}.\]
\end{lemma}

Recall that the first two components address the effect of the discretization error regarding the specifics of the optimal assortment. The third and last component concerns the translation error regarding the offered assortments $S_1,\ldots,S_T$. Since all these assortments lie in $\mathcal{A}_K$, as in \eqref{eq:setofbincollections}, we present the result below for a general set in $\mathcal{A}_K$.

\begin{lemma}
Let $\check{v}$ and $\check{w}$ be as in \eqref{eq:checkv} and \eqref{eq:checkw}, respectively. Let $\mathcal{A}_K$ be as in \eqref{eq:setofbincollections} and let $S\in\mathcal{A}_K$. Then the difference between the expected profit of $S$ under $\check{v}$ and $\check{w}$, and $v$ and $w$ is bounded from above by
\[r(S,\check{v},\check{w}) - r(S,v,w)\leqslant |\!|v-\check{v}|\!|_1 + \vmax\,|\!|w-\check{w}|\!|_1,\]
where $|\!|\cdot|\!|_1:=\int_0^1|\cdot|\,{\rm d}x$.
\label{lem:mistranslatingSt}
\end{lemma}
The three components of the discretization error are combined as follows. Let $S^*$, $\check{S}$ and $S^d$ be as in \eqref{eq:SstarandcheckS} and \eqref{eq:checkSandSd} and let $S_1,\ldots,S_T$ be the offered assortments. The instantaneous regret at time $t\in[T]$ can be split into four parts as
\begin{align}
r(S^*,v,w) - r(S_t,v,w) =\:& r(S^*,v,w) - r(\check{S},\check{v},\check{w}) \:+ \label{eq:split1}\\
&r(\check{S},\check{v},\check{w}) - r(S^d,\check{v},\check{w}) \:+\label{eq:split2}\\
& r(S^d,\check{v},\check{w}) - r(S_t,\check{v},\check{w}) \:+\label{eq:split3}\\
& r(S_t,\check{v},\check{w}) - r(S_t,v,w);\label{eq:split4}
\end{align}
the idea is to apply the triangle inequality.
For the right-hand side of \eqref{eq:split1}, \eqref{eq:split2}, and \eqref{eq:split4}, we apply Proposition \ref{prop:bridge}, Lemma \ref{lem:optimizingoversubset} and Lemma \ref{lem:mistranslatingSt}, respectively. Note that the term in \eqref{eq:split3} corresponds to the instantaneous regret of UCB. The remainder of the proof of Theorem \ref{th:upperboundc<1} consists of showing that both the $L_1$-distances $|\!|v-\check{v}|\!|_1$ and $|\!|w-\check{w}|\!|_1$ are of the order $1/N$ and applying Theorem 1 from \cite{Agrawal2019}.

\begin{remark}
The analysis of the upper bound on the regret of DUCB extends to higher  dimensional  continuous assortment problems.  In particular, if the dimension is $d\geqslant 2$, then one can discretize the set of products $[0,1]^d$ into $N^d$ bins. Under a smoothness assumption of the preference function and the marginal profit function, the order of the $L_1$-distance between the actual functions and the discretized functions remains $\O(1/N)$ as the difference can be bounded from above by a sum of $N^d$ terms that each are of order $N^{-(d+1)}$, similar as in \eqref{eq:L1discretizationerror}. As a result, the cumulative discretization error is of order $T/N$ and the total regret in higher dimensions is of the order (up to a logarithmic factor)
\[\frac{T}{N} + \sqrt{N^dT}.\]
Hence, the optimal value of $N$ is proportional to $T^{\frac{1}{d+2}}$ which results in a $T^{\frac{d+1}{d+2}}$ regret. This corresponds to the regret rate for continuum-armed bandit in higher dimensions \citep[see, e.g.,][]{Kleinberg2008,Bubeck2011a,Bubeck2011b}.
\end{remark}

\subsection{Regret lower bound}
\label{sec:capacitatedlowerbound}

In this section we construct an instance for the assortment optimization problem with capacity constraint, and we show that the regret of \emph{any} policy after $T$ time periods is at least a constant times $T^{2/3}$. This shows that the structural differences between optimal assortments with or without a capacity constraint under full information (Section \ref{subsec:full-info-c-is-1} and  \ref{sec:capacitatedfullinfo}) translate into a different complexity of the corresponding data-driven optimization problem, characterized by the growth rate of regret. 

We consider the following instance. Let $\underline{v} \in (0, 0.79)$, $\vmax \geqslant 4$, let $c \in (0, \tfrac{1}{4}]$, $s = 0.8 c$, $\delta=\tfrac{1}{2}$, 
and consider the marginal profit function
%
\begin{equation*}
  w(x) = (1-s)\frac{1-\delta}{1-\delta x} + s, \quad x \in [0,1].
\end{equation*}
To obtain a lower bound on regret, we construct `difficult instances' of preference functions that are hard to distinguish statistically, but that correspond to different optimal assortments. To this end, we first define a `baseline' preference function $v_0$ by
\begin{equation*}
  v_0(x) := \frac{s}{c(w(x) - s)} = \frac{s(1-\delta x)}{c(1-s)(1-\delta)}, \quad x \in [0,1].
\end{equation*}
This preference function has the property that $\varrho^*_0 := \max_{S\in\cS}r(S,v_0)$ is equal to $s$ (see Appendix B, Lemma \ref{lem:anicecoincidence}), and that $v_0(x) (w(x) - \varrho^*_0)$ does not depend on $x$. As a result, \emph{any} assortment of volume $c$ is optimal for this preference function.

The next step is to perturb the baseline preference function with small, positive `bumps' at different locations, such that the corresponding optimal assortment will be a collection of intervals centered around these bumps. The perturbed preference functions are, in a sense, close to each other (measured, e.g., by the $L_1$ norm), but correspond to different and possibly even disjoint optimal assortments. In particular, let $K\geqslant 2$ be an integer and $N_K:=\floor{K/c}$, and define the \emph{$i$-th bin} as the interval
\[B_i:=\left[c\frac{i-1}{K},c\frac{i}{K}\right),\qquad i\in [N_K].\]
Note that this definition differs from the bins presented in Section \ref{sec:capacitatedpolicy}. The definition here is convenient as the union of any $K$ distinct bins has combined volume of precisely $c$. Let $\mathcal{D}_K$ denote the collection of all subsets of $[N_K]$ of size $K$, i.e.,\
\[\mathcal{D}_K:= \big\{I\subseteq[N_K]:|I|=K\big\}.\]
For each collection of bins $I \in \mathcal{D}_K$ we now define a preference function $v_I$ that, roughly speaking, consists of the baseline preference function with small, positive bumps added at all bins $B_i$, $i \in I$. In particular, define the bump function $b(x)$ as the normal probability density function with parameters $\mu=0$ and $\sigma = 0.3$: 
\[b(x) := \frac{1}{\sigma\sqrt{2\pi}}e^{-x^2/2\sigma^2}, \quad x \in \R.\]
This function is shifted and re-scaled such that the probability mass on $[-1,1)$ is mapped onto $B_i$, as follows. For $i \in [N_K]$ and $x \in \R$, let
 \[ \phi_i(x) := \frac{2Kx}{c} - 2i+1,\]
be a linear transformation that satisfies $\phi_i(B_i) = [-1, 1)$, 
 and define
\[\tau_i(x) := \frac{c}{K}\: b\big(\phi_i(x)\big).\]
Finally, define the constant
\[ \beta := \frac{c}{K} \frac{1}{\sigma \sqrt{2 \pi}} \sum_{n \in \Z } \exp\left(- \frac{(2n-1)^2}{2 \sigma^2} \right), \]
and, for each $I \in \mathcal{D}_K$, define the preference function 
\[v_I(x):= v_0(x)\left(1+\sum_{i\in I}\tau_i(x) - \beta\right), \quad x \in [0,1].\]

\newpage\null
\begin{figure}[H]
\begin{center}
\begin{tikzpicture}
\begin{axis}[
    title={},
    width = 0.4\linewidth,
    ylabel={$b(x)$},
    xlabel={$x$},
    xmin=-1, xmax=1,
    ymin=0,
]

\addplot[
	smooth,
	samples=500,
    solid,
    color=black,
    ]
    {1/(0.3*sqrt(2*pi))*exp(-((x)^2)/(2*0.3^2))};

\end{axis}
\end{tikzpicture}
\begin{tikzpicture}
\begin{axis}[
    title={},
    width = 0.4\linewidth,
    ylabel={$v_I(x)$},
    xlabel={$x$},
    xmin=0, xmax=1,
    ymin=0, ymax=2.4,
]

\addplot[
    thin,
    color=black,
    ]
    table {vI.dat};

\foreach \x in {1,...,7}
{
	\addplot[line width=0.05pt,color=gray,dashed] coordinates {(0.125*\x,0) (0.125*\x,2.4)};
}

\node[color=gray] at (axis cs:  0.0625, 0.25) {\small $B_1$};
\node[color=gray] at (axis cs:  0.1875, 0.25) {\small $B_2$};
\node[color=gray] at (axis cs:  0.3125, 0.25) {\small $B_3$};
\node[color=gray] at (axis cs:  0.4375, 0.25) {\small $B_4$};
\node[color=gray] at (axis cs:  0.5625, 0.25) {\small $B_5$};
\node[color=gray] at (axis cs:  0.6875, 0.25) {\small $B_6$};
\node[color=gray] at (axis cs:  0.8125, 0.25) {\small $B_7$};
\node[color=gray] at (axis cs:  0.9375, 0.25) {\small $B_8$};

\end{axis}
\end{tikzpicture}
\caption{Left: bump function $b(\cdot)$. Right: preference function $v_I(\cdot)$ for $c=0.25$, $K=2$ and $I=\{2,5\}$.}
\label{fig:bumpyroad}
\end{center}
\end{figure}

\noindent The subtraction of the (small) constant $\beta$ ensures that $v_I(x)\leqslant v_0(x)$ for all $x\notin\bigcup_{i\in I}B_i$, i.e., the preference function dips just below the baseline function $v_0(x)$ for $x$ outside the collection of bins in $I$. This ensures that the optimal assortment corresponding to $v_I$ is approximately equal to the collection of intervals $\bigcup_{i\in I}B_i$ at which small bumps have been added.

Having defined a collection of preference functions, we now proceed in proving a regret lower bound. 
First, for any policy and any $I \in \mathcal{D}_K$, we bound the regret corresponding to preference function $v_I$ from below by an expression that counts how often products from the approximately optimal assortment
$\bigcup_{i\in I} B_i$ were not offered. To state the result, let 
\[\epsilon_I(x) := \frac{v_I(x) - v_0(x)}{v_0(x)} = \sum_{i\in I} \tau_i(x) -\beta,
\qquad I \in \mathcal{D}_K, x \in [0,1],
\]
and let
\[k(x) := \sum_{t=1}^T 1_{S_t}(x),\qquad x\in[0,1],\]
count the number of times that $x \in [0,1]$ is offered to consumers. Throughout the remainder of this section we fix an arbitrary policy $\pi$, and let $\P_I$ and $\E_I$ denote the probability law and the expectation operator under policy $\pi$ and preference function $v_I$. 
\begin{proposition} \label{prop:lower-bound-step-1}
There are constants $C_1 > 0$, $C_2 > 0$, independent of $\pi$, such that, for any $T \in \N$ and $I \in \mathcal{D}_K$,
\[\Delta_\pi(T,v_I) \geqslant C_1\int_{\bigcup_{i\in I}B_i} \big(T-\E_I[k(x)]\big)\epsilon_I(x){\rm d}x - C_2 \frac{T}{K}.\] 
\end{proposition}
The proposition is proven by exploiting the structure of the optimal assortment as outlined in Section \ref{sec:capacitatedfullinfo} and the fact that the definition of $v_I$ implies that the corresponding optimal assortment is approximately equal to $\bigcup_{i\in I}B_i$. The constants $C_1$, $C_2$ are given explicitly in the proof of Proposition \ref{prop:lower-bound-step-1}. 

The second step in the proof of the regret lower bound is the following result, which provides an upper bound on how the expected number of times that a product $x \in [0,1]$ is offered changes when the preference function is changed from $v_I$ to $v_{I \backslash \{i\}}$, for some $i \in I$. 

\begin{proposition}\label{prop:steptwo}
Let $x\in[0,1]$, $I\in \mathcal{D}_K$, and $J = I\backslash \{i\}$ for some $i\in I$. Then there is a constant $C_c > 0$ independent of $\pi$, such that 
\begin{align} \label{eq:steptwo}
\Big|\E_I[k(x)]-\E_J[k(x)]\Big|\leqslant C_c\left(\frac{T}{K}\right)^{3/2}.
\end{align}
\end{proposition}
This bound is proven by relating the left-hand side of \eqref{eq:steptwo} to the Kullback-Leibler divergence of $\P_I$ and $\P_J$, using Pinsker's inequality, and subsequently bounding this expression from above by carefully analysing its dependence on $v_I$ and $v_J$. The constant $C_c$ is given explicitly in the proof of Proposition \ref{prop:steptwo}. 

With Propositions \ref{prop:lower-bound-step-1} and \ref{prop:steptwo} at hand, we finally arrive at our regret lower bound. \begin{theorem}\label{th:lowerboundc<1}
There is a $\Cmin > 0$, independent of $\pi$, such that, for $T \in \N$,
\[\Delta_\pi(T) \geqslant \Cmin\,T^{2/3}.\]
\end{theorem}
To prove the theorem, we first show that the preference functions $\{v_I : I \in \mathcal{D}_K, K \in \N\}$ satisfy Assumption \ref{A1}. This implies that the worst-case regret $\sup_{v \in \cV} \Delta_{\pi}(T, v)$ is bounded from below by the expected regret when the preference function is chosen uniformly at random from $\{ v_I : I \in \mathcal{D}_K\}$, for any fixed $K$. The regret corresponding to each $v_I$ is then bounded from below by an expression that involves the expected number of times that products from the approximate optimal assortment $\bigcup_{i \in I} B_i$ are not offered, using Proposition \ref{prop:lower-bound-step-1}. 
Proceeding in a similar fashion as in the proof of the regret lower bound obtained by \cite{Chen2017} for discrete assortments, while dealing with all the intricacies of having a continuum product space, we connect the expression in Proposition~\ref{prop:lower-bound-step-1} to the statement \eqref{eq:steptwo} of Proposition~\ref{prop:steptwo}. By carefully selecting $K$, we arrive at the stated lower bound.

\section{Numerical experiments}
\label{sec:numerical}

In this section we compare the numerical performance of the policies proposed in this study to alternative policies that are specifically designed for the discrete assortment problem.  We use the notations and concepts introduced in Sections \ref{sec:capacitatedpolicy} and \ref{sec:capacitatedupperbound}. 
In the uncapacitated case, we compare our algorithm SAP to  (i) the Thompson Sampling based algorithm by \cite{Agrawal2017}, and (ii) the Trisection-based algorithm by \cite{Chen2018}, both applied to discretized versions of the continuous problem. To have a fair comparison, we use in all our numerical experiments the same discretization of the product space as in our DUCB algorithm. We refer to these two policies from the literature, applied to discretized versions of the continuous assortment problem, as Discretized Thompson Sampling (DTS) and Discretized Trisection (DTR). 

In the capacitated case, we compare our algorithm DUCB to DTS but not to DTR, since the Trisection-based algorithm of \cite{Chen2018} is not designed to handle capacity constraints. In addition, in the capacitated case we also evaluate the performance of an adjusted version of DUCB (called \emph{ADUCB}) in which we replace the constant $48$ in \eqref{eq:updatevUCB} by $1$; our numerical results indicate that changing this constant significantly improves performance. Optimally tuning this constant is an interesting direction for future research but is outside the scope of this paper. In this section we report numerical results on the regret behavior for these different algorithms; Appendix E contains additional numerical experiments on the predictive performance of our continuous model. 

We set the preference function $v$ as the bi-modal function that is plotted in Figure \ref{fig:simulationv}. This function is defined as
\begin{equation*}
  v(x) = \frac{1}{10}+\frac{1}{5}(2+x)(1-x) + \frac{2}{7}\phi(x;0.33,0.1) + \frac{1}{5}\phi(x;0.8,0.1), \qquad x\in[0,1],
\end{equation*}
where $\phi(\,\cdot\,;\mu,\sigma)$ denotes the normal probability density function with mean $\mu$ and standard deviation $\sigma$. In addition, we set
$w(x) = x$, $x\in[0,1]$, as the marginal revenue function. We test our algorithms with $c = 1$ and $c = 0.5$, corresponding to capacity constraints $K = N$ and $K = \floor{N/2}$ in the discretized versions. In line with Theorem \ref{th:upperboundc<1}, we set the discretization parameter $N$ as $\floor{\gamma T^{1/3}}$ with $\gamma = \max\{\vmax,1/c+1\}$ and $\vmax = 2$. The parameters of SAP are set to $\alpha = 3$, $\beta = 2$, $\varrho_1 = 0$.
The algorithms' average regrets over 100 simulations after $T$ time periods, for $T \in \{1\,000, 2\,000, \ldots, 10\,000\}$, are recorded in Table \ref{tab:T2} and \ref{tab:T3}.

\newpage\null
\begin{table}[H]
  \centering
  \small
    \begin{tabular}{|l|c|c|c|c|c|c|c|c|c|c|}
      \hline\hline
       & \multicolumn{10}{c|}{Time horizon $T$}\\
      \hline
      Policy & 1\,000 & 2\,000 & 3\,000 & 4\,000 & 5\,000 & 6\,000 & 7\,000 & 8\,000 & 9\,000 & 10\,000 \vphantom{\Big|}\\
      \hline
       DTR & 8.67 & 18.2 & 25.5 & 32.1 & 35.9 & 40.9 & 48.3 & 55.8 & 63.7 & 70.0  \\
       DTS & 1.46 & 1.94 & 2.10 & 2.25 & 2.56 & 2.45 & 2.87 & 3.06 & 2.85 & 3.34  \\
       SAP & 0.380 & 0.417 & 0.439 & 0.452 & 0.463 & 0.474 & 0.483 & 0.49 & 0.500 & 0.507  \\
       \hline
       $N$ & 19 & 25 & 28 & 31 & 34 & 36 & 38 & 39 & 41 & 43  \\
       \hline\hline
    \end{tabular}\vspace{3mm}
  \caption{\label{tab:T2} Simulated average regret of the policies with $c = 1$ based on 100 simulations.}
\end{table}

\begin{table}[H]
  \centering
  \small
    \begin{tabular}{|l|c|c|c|c|c|c|c|c|c|c|}
      \hline\hline
       & \multicolumn{10}{c|}{Time horizon $T$}\\
      \hline
      Policy & 1\,000 & 2\,000 & 3\,000 & 4\,000 & 5\,000 & 6\,000 & 7\,000 & 8\,000 & 9\,000 & 10\,000  \vphantom{\Big|}\\
      \hline
       DTS & 12.8 & 19.8 & 26.8 & 32.0 & 38.8 & 33.2 & 46.8 & 52.6 & 43.8 & 48.0  \\
       DUCB & 89.6 & 153 & 206 & 252 & 295 & 334 & 371 & 403 & 440 & 470   \\
       ADUCB & 9.81 & 16.9 & 23.6 & 29.8 & 35.3 & 32.6 & 46.7 & 52.1 & 45.8 & 50.1  \\
       \hline
       $N$ & 29 & 37 & 43 & 47 & 51 & 54 & 57 & 59 & 62 & 64   \\
       \hline\hline
    \end{tabular}\vspace{3mm}
  \caption{\label{tab:T3} Simulated average regret of the policies with $c=0.5$ based on 100 simulations.}
\end{table}



Table \ref{tab:T2} shows that our algorithm SAP outperforms the alternatives DTR and DTS by a significant margin. The top row of Figure \ref{fig:regression} plots the regret of SAP as function of $T$, both on a linear (left-hand panel) and a logarithmic scale (right-hand panel). The linear growth rate of regret as function of $\log T$ in Figure \ref{fig:regression} confirms our theoretical result on the regret behavior of SAP. Fitting the curve $\mathcal{R}(T)= \gamma_1+\gamma_2\log T$ using linear regression, we find that $\gamma_1 = 0.00171$ and $\gamma_2 = 0.0545$.

Table \ref{tab:T3} records the regret of DTS, DUCB, and ADUCB; the results are visualized in the middle and bottom row of Figure \ref{fig:regression}. The figure illustrates that the regrets of both DUCB and ADUCB grow sublinearly. The adjusted policy ADUCB performs on par with DTS, while both ADUCB and DTS outperform DUCB. This suggests that fine-tuning the constants in the updating formula for the upper confidence bounds can lead to less regret.
Fitting the curve $\log \mathcal{R}(T) = \gamma_1 + \gamma_2 \log T$ using linear regression, we find that $\gamma_2 = 0.70$ for DUCB and $\gamma_2=0.67$ for ADUCB. This confirms, particularly for ADUCB, our theoretical regret bounds of $T^{2/3}$ (up to a logarithmic term). It is worth observing 
and illustrated by Figure \ref{fig:regression}
that the regret for our policies is not necessarily monotone in $T$; this is a result of the discretization to an integer number of products. 




\newpage\null
\vfill
\begin{figure}[H]
\begin{center}
\begin{tikzpicture}
\begin{axis}[
  width = 0.45\linewidth,
  xmin = 0, xmax = 10500,
  ymin = 0,
  xtick={2000,4000,6000,8000,10000},
  xticklabels={$2$,$4$,$6$,$8$,$10$},
  xtick scale label code/.code={$\cdot 10^4$},
  xlabel={$T$},
  ylabel={$\mathcal{R}(T)$},
  legend pos = south east,
]

\addplot [box plot median] table {CI1.dat};
\addplot [box plot top whisker] table {CI1.dat};
\addplot [box plot bottom whisker] table {CI1.dat};

\addplot[
	smooth,
  dashed,
  color=blue,
  domain=3:10500,
  samples=1000,
  ]
  {0.054529040582322*ln(x) + 0.001712639620561};

\end{axis}
\end{tikzpicture}
\begin{tikzpicture}
\begin{axis}[
  width = 0.45\linewidth,
  xmin = 750, xmax = 10500,
  xmode = log,
  xlabel={$T$},
  ylabel={$\mathcal{R}(T)$},
  legend pos = south east,
]

\addplot [box plot median] table {CI1.dat};
\addplot [box plot top whisker] table {CI1.dat};
\addplot [box plot bottom whisker] table {CI1.dat};

\addplot[
	smooth,
  dashed,
  color=blue,
  domain=3:10500,
  samples=1000,
  ]
  {0.054529040582322*ln(x) + 0.001712639620561};

\end{axis}
\end{tikzpicture}
\begin{tikzpicture}
\begin{axis}[
  width = 0.45\linewidth,
  xmin = 0, xmax = 10500,
  ymin = 0,
  xtick={2000,4000,6000,8000,10000},
  xticklabels={$2$,$4$,$6$,$8$,$10$},
  xtick scale label code/.code={$\cdot 10^4$},
  xlabel={$T$},
  ylabel={$\mathcal{R}(T)$},
  legend pos=south east,
]

\addplot [box plot median] table {CI2.dat};
\addplot [box plot top whisker] table {CI2.dat};
\addplot [box plot bottom whisker] table {CI2.dat};

\addplot[
	smooth,
  dashed,
  color=blue,
  samples=1000,
  domain=3:10500,
  ]
  {0.782896754202266*x^0.695286586477471};

\end{axis}
\end{tikzpicture}
\begin{tikzpicture}
\begin{axis}[
  width = 0.45\linewidth,
  xmin = 750, xmax = 10500,
  xmode = log,
  xlabel={$T$},
  ylabel={$\mathcal{R}(T)$},
  legend pos=south east,
]

\addplot [box plot median] table {CI2.dat};
\addplot [box plot top whisker] table {CI2.dat};
\addplot [box plot bottom whisker] table {CI2.dat};

\addplot[
	smooth,
  dashed,
  color=blue,
  samples=1000,
  domain=3:10500,
  ]
  {0.782896754202266*x^0.695286586477471};

\end{axis}
\end{tikzpicture}
\begin{tikzpicture}
\begin{axis}[
  width = 0.45\linewidth,
  xmin = 0, xmax = 10500,
  ymin = 0,
  xtick={2000,4000,6000,8000,10000},
  xticklabels={$2$,$4$,$6$,$8$,$10$},
  xtick scale label code/.code={$\cdot 10^4$},
  xlabel={$T$},
  ylabel={$\mathcal{R}(T)$},
  legend pos=south east,
]

\addplot [box plot median] table {CI3.dat};
\addplot [box plot top whisker] table {CI3.dat};
\addplot [box plot bottom whisker] table {CI3.dat};

\addplot[
	smooth,
  dashed,
  color=blue,
  samples=1000,
  domain=3:10500,
  ]
  {0.114967208041740*x^0.666086080343409};

\end{axis}
\end{tikzpicture}
\begin{tikzpicture}
\begin{axis}[
  width = 0.45\linewidth,
  xmin = 750, xmax = 10500,
  xmode = log,
  xlabel={$T$},
  ylabel={$\mathcal{R}(T)$},
  legend pos=south east,
]

\addplot [box plot median] table {CI3.dat};
\addplot [box plot top whisker] table {CI3.dat};
\addplot [box plot bottom whisker] table {CI3.dat};

\addplot[
	smooth,
  dashed,
  color=blue,
  samples=1000,
  domain=3:10500,
  ]
  {0.114967208041740*x^0.666086080343409};

\end{axis}
\end{tikzpicture}
\caption{The whiskers show the 95\% confidence interval of the mean cumulative regret for SAP (top row), DUCB (middle row) and ADUCB (bottom row) with regular axes (left panels) and a logarithmic axis for $T$ (right panels), based on $100$ simulations. The blue solid line shows the fitted curves $\gamma_1 + \gamma_2\log T$ with $\gamma_1 = 0.00171$ and $\gamma_2 = 0.0545$ (first row), $\gamma_1 T^{\gamma_2}$ with $\gamma_1 = 0.783$ and $\gamma_2 = 0.700$ (second row) and $\gamma_1 T^{\gamma_2}$ with $\gamma_1 = 0.115$ and $\gamma_2 = 0.666$ (third row).}
\label{fig:regression}
\end{center}
\end{figure}
\vfill

\section{Discussion}
In this paper we have introduced the  concept of continuous assortment optimization with demand learning. We distinguish between the capacitated and uncapacitated case, revealing intrinsically different regret behavior: we show that the asymptotically optimal regret rate in the absence of a capacity constraint grows logarithmically in the time horizon, whereas imposing a capacity constraint leads to $T^{2/3}$ regret. 
To our knowledge, this paper is the first to extend discrete assortment optimization problems to the continuous realm.

Our work points to various directions for future research. 
First, the customer-purchase model used in this paper is the natural continuous equivalent of the well-studied discrete multinomial logit choice model. It remains an open question how one constructs a random utility model that serves as a theoretical justification of the continuous choice model. 
Second, in line with the majority of the assortment optimization literature, our setup assumes that product prices are exogenous. A question of  practical interest is to consider price and assortment decisions simultaneously in our continuous model, potentially in a competitive setting. 
Third, we have constructed an example in which the optimal assortment is not an uninterrupted interval. It would be interesting to study under which conditions a single interval solution is optimal, and whether one can bound the maximum loss when the decision maker is restricted to offering a single interval. 


\bibliographystyle{abbrvnat} 
\bibliography{myrefs} 

\newpage

\section*{Appendix A: Mathematical proofs of Section \ref{sec:uncapacitated}}
\label{app:uncapacitated}

\subsection*{A.1. Proofs of the results in Section \ref{sec:upperbounduncapacitated}}

\subsubsection*{Proof of Theorem \ref{th1}.}
Define $g(y) := r([y,1],v)$ for $y \in [0,1]$ and $h(\varrho):= g(w^{-1}(\varrho))$ for $\varrho\in[0,1]$. Also, let $\varrho^*$ denote the optimal expected profit, i.e.,
\[\varrho^* := \max\{r(S,v):S\in\cS\}.\]
The following auxiliary results turn out to be useful; the proof of Lemma \ref{lem:properties-of-g} follows after the proof of Theorem \ref{th1}.
\begin{lemma} \label{lem:properties-of-g}
It holds that $h(\varrho^*) = \varrho^*$. Moreover, for $\varrho\in[0,1]$, the following properties hold:
\begin{align} \label{eq:k0}
 (\varrho - \varrho^*) ( h(\varrho) - \varrho) &\leqslant - \frac{1}{1+\vmax} (\varrho-\varrho^*)^2, \\
 h(\varrho^*) - h(\varrho) &\leqslant C (\varrho - \varrho^*)^2, \label{eq:k3}
\end{align}
for a universal constant $C>0$.
\end{lemma}
Note that by our choice of $\beta\geqslant\max\{0,\alpha-1\}$ it follows that $\varrho_t\in[0,1]$ for all $t\in\N$. With these properties at our disposal, we continue the proof of the worst-case bound for Case~1, which closely follows the analysis of \cite{Broadie2011} on stochastic approximation schemes. For the policy $\pi={\rm SAP}(\alpha,\beta)$, it holds for all $t \in \N$ that
\begin{align*}
 \E_\pi[ (\varrho_{t+1} &- \varrho^*)^2 \,|\, \varrho_t] \\
&= \E_\pi\left[ \left( \varrho_t + a_t (R_t - \varrho) - \varrho^*
\right)^2\,|\, \varrho_t\right] \\
&= \E_\pi\left[ (\varrho_t - \varrho^*)^2 + 2 (\varrho_t - \varrho^*) a_t (R_t - \varrho_t) + a_t^2 (R_t - \varrho_t)^2 \,|\, \varrho_t\right]\\
&\leqslant (\varrho_t - \varrho^*)^2 + 2 (\varrho_t - \varrho^*) a_t (h(\varrho_t) -\varrho_t) + a_t^2 \\
&\leqslant (\varrho_t - \varrho^*)^2 \left(1 - \frac{2 a_t}{1+\vmax}\right) + a_t^2
 \end{align*}
where the first inequality follows from $R_t-\varrho_t\in[-1,1]$ and the second inequality from Lemma \ref{lem:properties-of-g}, i.e.,\ \eqref{eq:k0}. 
Recalling the definition of $a_t$, an immediate consequence of the above bound is that we have, with $\delta_t := \E_\pi[ (\varrho_t - \varrho^*)^2]$, for any $t \in \N$, 
\begin{align} \label{eq:delta-recursion}
\delta_{t+1} \leqslant \delta_t \left(1 - \frac{2}{1+\vmax}\cdot\frac{\alpha}{t + \beta} \right) + \frac{\alpha^2}{(t + \beta)^2}.
\end{align}

From the inequality in \eqref{eq:delta-recursion} one can derive the following lemma in a relatively straightforward way. Its (inductive) proof follows after the proof of Theorem \ref{th1}. 

\begin{lemma}\label{L1}
There exists a $\CL>0$ such that for all $t\in\N$
\begin{align} \label{eq:delta-bound}
\delta_t \leqslant \frac{\CL}{t + \beta}. 
\end{align} 
\end{lemma}

We proceed by deriving an upper bound on the regret of the policy $\pi={\rm SAP}(\alpha,\beta)$, relying on the upper bound on $\delta_t$ stated in Lemma \ref{L1}. Let $C$ denote the constant as in Lemma \ref{lem:properties-of-g}. The regret can be majorized as follows:
\begin{align*}
\Delta_{\pi}(T, v) &= 
\sum_{t=1}^T \E_{\pi}[h(\varrho^*) - h(\varrho_t)] \leqslant 
C\sum_{t=1}^T \delta_t \\
&\leqslant C \sum_{t=1}^T \frac{\CL}{t + \beta} \leqslant 3 C \CL \log T,
\end{align*}
for all $T \geqslant 2$, where the first inequality follows by \eqref{eq:k3}, the second inequality by \eqref{eq:delta-bound}, 
and the third inequality by $\sum_{t=1}^T (t + \beta)^{-1} \leqslant 3 \log T$ for all $T \geqslant 2$. We have proven the stated with $\Cmax:=3 C \CL$.
\hfill$\Box$
\vspace{3mm}

\subsubsection*{Proof of Lemma \ref{lem:properties-of-g}.}
We prove the three claims separately.

\noindent $\rhd$ Following the reasoning at \eqref{eq:rewrittenSb}, we find that
\begin{align*}
  \varrho^* &= \max\left\{\varrho\in[0,1]: \max_{S \in \cS} \int_{S} v(x)\big(w(x)-\varrho\big){\rm d}x\geqslant \varrho\right\} \\
  & =\max\left\{\varrho\in[0,1]:\int_{w^{-1}(\varrho)}^1 v(x)\big(w(x)-\varrho\big){\rm d}x\geqslant \varrho\right\}.
\end{align*}
Since $w^{-1}(\cdot)$ is continuous, we know that, with $\varrho\in[0,1]$,
\[\mathcal{I}(\varrho) := \int_{w^{-1}(\varrho)}^1 v(x)\big(w(x)-\varrho\big){\rm d}x\] is continuous. Also, since $w^{-1}(\cdot)$ is non-decreasing and $\varrho\mapsto v(x)\big(w(x)-\varrho\big)$ is decreasing, we know that $\mathcal{I}(\cdot)$ is non-increasing. Moreover, note that $\mathcal{I}(0)>0$ and $\mathcal{I}(1) = 0$. As a result, there exists a unique solution to $\mathcal{I}(\varrho) = \varrho$, and that this equation is precisely solved by $\varrho^*$. The proof is completed by observing that the equation $\mathcal{I}(\varrho)=\varrho$ is equivalent to $h(\varrho) = \varrho$.

\noindent $\rhd$ For $\varrho =\varrho^*$, \eqref{eq:k0} immediately holds. Now, assume that $\varrho\in[0,\varrho^*)$, then
\begin{align*}
h(\varrho) -\varrho & = \frac{\int_{w^{-1}(\varrho)}^1 v(x)w(x)\d x}{1 + \int_{w^{-1}(\varrho)}^1v(x)\d x} -\varrho  = \frac{\mathcal{I}(\varrho) - \varrho}{1 + \int_{w^{-1}(\varrho)}^1v(x)\d x} \\
& \geqslant \frac{\mathcal{I}(\varrho^*) - \varrho}{1 + \int_{w^{-1}(\varrho)}^1v(x)\d x}  = \frac{\varrho^* - \varrho}{1 + \int_{w^{-1}(\varrho)}^1v(x)\d x} \\
& \geqslant -\frac{1}{1+\vmax}(\varrho - \varrho^*).
\end{align*}
where the first inequality holds by the non-increasingness of $\mathcal{I}(\cdot)$. As a result,
\[(\varrho-\varrho^*)(h(\varrho) - \varrho) \leqslant -\frac{1}{1+\vmax}(\varrho - \varrho^*)^2.\]
Next, assume that $\varrho\in(\varrho^*,1]$. It holds that
$h(\varrho) \leqslant h(\varrho^*) = \varrho^*$
which implies
$h(\varrho) - \varrho\leqslant -(\varrho - \varrho^*)$
and therefore
\[(\varrho-\varrho^*)(h(\varrho) - \varrho) \leqslant - (\varrho - \varrho^*)^2\leqslant -\frac{1}{1+\vmax}(\varrho - \varrho^*)^2.\]
Hence, for all $\varrho\in[0,1]$ it holds that
\[(\varrho-\varrho^*)(h(\varrho) - \varrho) \leqslant -\frac{1}{1+\vmax}(\varrho - \varrho^*)^2.\]

\noindent $\rhd$ Firstly, note that
\begin{align*} 
g'(y) &=
\frac{{\rm d} }{{\rm d} y}\frac{\int_y^1 v(x)w(x){\rm d}x}{1 + \int_y^1 v(x) {\rm d}x}
= \big(r([y,1],v) - w(y) \big) \cdot \frac{v(y)}{1 + \int_y^1 v(x) {\rm d}x} = \big(g(y)-w(y)\big)\cdot \xi(y),
\end{align*}
where, for $y \in [0,1]$,  
\[ \xi(y) := \frac{v(y)}{1 + \int_y^1 v(x) {\rm d}x}.\]
Secondly, we show that there exists a universal constant $C_0$ such that
\begin{equation}
\sup_{y\in (0,1)}\left\{ -g''(y)\right\}\leqslant C_0.
\label{eq:secondderibound}
\end{equation}
To prove \eqref{eq:secondderibound} observe that $g''(y) = \big(g(y) - w(y)\big)\big( \xi'(y) + \xi(y)^2\big) - w'(y) \xi(y)$, and
\begin{align*}
\xi'(y) = \frac{v'(y)}{1 + \int_y^1 v(x){\rm d}x} + \xi(y)^2.
\end{align*}
Since $g(y) - w(y) \in [-1,1]$ for all $y \in (0,1)$, we obtain 
\begin{align*}
-g''(y) &=
- \big(g(y) - w(y)\big) \big(\xi'(y) + \xi(y)^2\big) + w'(y) \xi(y) \\ 
&\leqslant \sup_{y \in (0,1)} \{ |\xi'(y)| + \xi(y)^2 \}
+ \sup_{y \in [0,1]} w'(y) \vmax \\ 
&\leqslant  \sup_{y \in (0,1), v \in \cV} \{ |v'(y)| + 2 \vmax^2 \}
+ \sup_{y \in [0,1]} w'(y) \vmax 
=: C_0.
\end{align*}
Now, let $\varrho\in[0,1]$ and denote $y=w^{-1}(\varrho)$ and $y^* = w^{-1}(\varrho^*)$. We distinguish two cases. Firstly, assume that $\varrho^*\geqslant w(0)$ or, equivalently, $g'(y^*)=0$. Then there is a $\tilde{y} \in (0,1)$ such that
$g(y) = g(y^*) + \tfrac{1}{2} g''(\tilde{y}) (y - y^*)^2.$
Therefore, we can apply \eqref{eq:secondderibound} to obtain, with
\[k_w:=\inf_{x\in(0,1)}w'(x)\]
that
\begin{align*}
h(\varrho^*) - h(\varrho) &= g(y^*) - g(y)  = -\tfrac{1}{2} g''(\tilde{y})(y-y^*)^2 \\
& \leqslant \tfrac{1}{2}C_0 (y-y^*)^2  \leqslant \frac{C_0}{2(k_w)^2} (\varrho-\varrho^*)^2,
\end{align*}
where at the final inequality we used that $w^{-1}(\cdot)$ is $(k_w)^{-1}$-Lipschitz continuous on $[0,1]$; note that $k_w$ is strictly positive due to the assumptions imposed on $w$. 
Now we consider the second case: assume that $\varrho^*<w(0)$ or, equivalently, $g'(y^*)<0$. In this case, $\varrho^* = g(0)$ and $w^{-1}(\varrho^*) = 0$. For $\varrho\in[0,w(0))$, $w^{-1}(\varrho) = w^{-1}(\varrho^*)$, and 
statement \eqref{eq:k3} holds for any constant $C\geqslant 0$. Now, let $\varrho\in [w(0),1]$. Then note that by \eqref{eq:secondderibound}
\[g(0) - g(y) \leqslant -g'(0) y + \tfrac{1}{2}C_0 y^2.\]
Next, note that since $w^{-1}(\cdot)$ is non-decreasing and $(k_w)^{-1}$-Lipschitz continuous
\[y = w^{-1}(\varrho) - w^{-1}(\varrho^*)\leqslant \frac{1}{k_w}(\varrho - \varrho^*)\]
and note that
\[0\leqslant -g'(0) = \big(w(0) - g(0)\big)\xi(0) \leqslant \xi(0)(\varrho-\varrho^*).\]
We conclude that
\begin{align*}
h(\varrho^*) - h(\varrho) &= g(0) - g(y) \\
&\leqslant \left( \frac{\xi(0)}{k_w} + \frac{C_0}{2(k_w)^2}\right)(\varrho-\varrho^*)^2 \leqslant \left( \frac{\vmax}{k_w} + \frac{C_0}{2(k_w)^2} \right)(\varrho-\varrho^*)^2.
\end{align*}
This proves \eqref{eq:k3} for all $\varrho\in[0,1]$ with
\[C = \frac{\vmax}{k_w} + \frac{C_0}{2(k_w)^2}.\]
\hfill$\Box$
\vspace{3mm}

\subsubsection*{Proof of Lemma \ref{L1}.}
We show, by induction, that the inequality \eqref{eq:delta-recursion} implies that, for some $\CL>0$, for all $t \in \N$ it holds that $\delta_t \leqslant {\CL}/({t + \beta})$.
To this end, let $K_0:=(1+\vmax)^{-1}$ and
\[\CL := \max\left\{ 1 + \beta, \alpha(1+\vmax) \right\}. \]
For $t =1$, we note that
\[\delta_1 \leqslant 1 \leqslant \frac{\CL}{1+\beta}.\] 
Now, suppose $\delta_t \leqslant \CL / (t + \beta)$ for $t \leqslant t_0$ for some $t_0$. Then, for $t > t_0$, it follows that
\begin{align*}
\frac{t+\beta}{t+ \beta + 1} - 2 \alpha K_0 &< 1 - 2 \alpha K_0
\leqslant - \alpha K_0,
\end{align*}
since $\alpha \geqslant K_0^{-1}$ and therefore
\begin{align*}
 \CL \left( \frac{t+\beta}{t+ \beta + 1} - 2 \alpha K_0\right) + \alpha^2
< -\CL \alpha K_0+ \alpha^2\leqslant 0,
\end{align*}
by definition of $\CL$. This implies that
\begin{align*}
\CL \left( (t+ \beta) - 2 \alpha K_0 - \frac{(t + \beta)^2}{t+\beta+1} \right) + \alpha^2 \leqslant 0,
\end{align*}
and thus
\begin{align*}
\frac{\CL}{t + \beta} \left(1 - 2 \frac{\alpha K_0}{t + \beta}\right) + \frac{\alpha^2}{(t + \beta)^2} \leqslant \frac{\CL}{(t + \beta + 1)}.
\end{align*}
This, by \eqref{eq:delta-recursion} in combination with the induction hypothesis, yields $\delta_{t+1} \leqslant {\CL}/{(t+1+\beta)}$, so that we have proven the lemma.
\hfill$\Box$

\subsection*{A.2. Proofs of the results in Section \ref{sec:uncapacitatedlowerbound}}

\subsubsection*{Proof of Theorem \ref{thm:lower-bound-c-is-1}.}
This proof relies on the Van Trees inequality, which can be seen as a Bayesian counterpart of the 
Cram\'er-Rao lower bound. 
Let $\Theta := [\thetamin, \thetamax]$, with
$\thetamax = \vmax$, $\thetamin = c_0 + (\vmax - c_0) / 2$, and 
\[ c_0 := \max\left\{ \underline{v}, \frac{w(0)}{\int_0^1( w(x)-w(0)) \dx} \right\}.\]
Observe that $\underline{v} < \thetamin < \thetamax = \vmax$, because of the assumption $\vmax > {w(0)}/{\int_0^1 (w(x)-w(0)) \dx}$.

For later reference, we introduce the probability density function $\lambda(\cdot)$ on $\Theta$ by
\[ \lambda(\theta) := 
\frac{2}{\thetamax - \thetamin} \cos^2\left( \pi \frac{ \theta - \thetamin}{\thetamax - \thetamin} - \pi/2 \right). \]
Observe that $\lambda(\cdot)$ is zero on the boundary of $\Theta$. Later, when applying the Van Trees inequality, we work with a random $\theta$, sampled from a distribution with density $\lambda(\cdot)$.

We start the proof with a number of definitions and preliminary observations. Let $v_{\theta}(x) := \theta$ for all $x \in [0,1]$ and all $\theta \in \Theta$.
Also, define $g(y, \theta) := r( [y,1], v_{\theta})$, for $y \in [0,1]$ and $\theta \in \Theta$. 
Let $g'(y, \theta)$ denote the partial derivative of $g(y, \theta)$ with respect to $y$, for $y \in (0,1)$. 
As in the proof of Theorem~\ref{th1}, 
\[g'(y, \theta) = (g(y,\theta) - w(y)) \cdot \xi(y, \theta), \:\:\:\:\:\:\:\mbox{where}\:\:\:\:
\xi(y, \theta) := \frac{v_{\theta}(y)}{1 + \int_y^1 v_{\theta}(x) {\rm d}x}.\] In addition, all $y \in (0,1)$ such that $g'(y, \theta) = 0$ satisfy $g''(y, \theta) < 0$, where $g''(y, \theta)$ is the second derivative of $g(y, \theta)$ to $y$. Observe that
$g(0, \theta) - w(0) > 0$ for all $\theta \in \Theta$, since $\thetamin > c_0$. 
It follows that for all $\theta \in \Theta$ there is a unique maximizer $y(\theta) \in (0,1)$ of $g(y, \theta)$ with respect to $y$; this maximizer is the unique solution $y \in [0,1]$ to the equation $g(y, \theta) = w(y)$. 
Moreover, observe that
$g(y, \theta)$ is strictly increasing in $\theta$, for all $y \in (0,1)$, and therefore
\[0 = g(y(\theta), \theta) - w(y(\theta)) < g(y(\theta), \theta') - w(y(\theta))\]
for all $\thetamin \leqslant \theta < \theta' \leqslant \thetamax$, which implies that $y(\theta') > y(\theta)$. Thus, $y(\theta)$ is increasing in $\theta$, for $\theta \in \Theta$. 

A complication in the proof is that in principle we can optimize over all sets $S\in \cS$, which we would like to somehow convert into an optimization over intervals. This explains the relevance of the following objects: for $\theta \in \Theta$ and $S \in \cS$, we define
\[ \psi(\theta) := {\rm vol}([y(\theta), 1]) = 1 - y(\theta), \:\:\:\: \psi^S := {\rm vol}(S). \]

\noindent $\rhd$ Step~1. 
We first show that 
$r( [y(\theta), 1], v_{\theta})$ and $r(S, v_{\theta})$ can only be close if $[y(\theta), 1]$ and $S$ are close (a necessary condition for which is that $\psi(\theta)$ and $\psi^S$ are close). More concretely,
for all $\theta \in \Theta$ and all $S \in \cS$, 
\[ r( [y(\theta), 1], v_{\theta}) - r(S, v_{\theta}) \geqslant \kappa_0 (\psi(\theta) - \psi^S)^2,\:\:\:\:\:\:\:\:
\mbox{where}\:\:\:\:\kappa_0 := \frac{\thetamin k_w / 2}{1 + \thetamax}. \]
To this end, for $v \in \mathcal{V}$ let $\varrho^*_v = \max_{S \in \cS} r(S,v)$, and let $S^*(v)$ be a corresponding maximizer.
From \[\varrho^*_v = \frac{\int_{S^*(v)} v(x) w(x) {\rm d}x } {1 + \int_{S^*(v)} v(x) {\rm d}x},\]
it follows $\varrho^*_v = \int_{S^*(v)} v(x) (w(x) - \varrho^*_v) {\rm d}x$, and thus, for all $S \in \cS$, 
\begin{align*}
 r(S^*(v),& v) - r(S,v) = 
\varrho^*_v \frac{1 + \int_S v(x) {\rm d}x }{1 + \int_S v(x) {\rm d}x} - \frac{\int_S v(x) w(x) {\rm d}x}{1 + \int_S v(x)} \\
&= \frac{1}{1 + \int_S v(x) {\rm d}x} \left( \varrho^*_v + \int_S v(x) (\varrho^*_v - w(x)) {\rm d}x \right) \\
&= \frac{1}{1 + \int_S v(x) {\rm d}x} \left( \int_{S^*(v)} v(x) (w(x) - \varrho^*_v) {\rm d}x - \int_S v(x) (w(x) - \varrho^*_v) {\rm d}x \right) \\
&= \frac{1}{1 + \int_S v(x) {\rm d}x} \left( \int_{S^*(v) \backslash S} v(x) (w(x) - \varrho^*_v) {\rm d}x 
+ \int_{S \backslash S^*(v)} v(x) (\varrho^*_v - w(x)) {\rm d}x \right).
\end{align*}
Let $\theta \in \Theta$ and $S \in \cS$. If $x \in S^*(v_{\theta}) \backslash S$, then $x \in S^*(v_{\theta}) = [y(\theta), 1]$, which implies that $w(x) - \varrho^*_{v_{\theta}} \geqslant w(y(\theta)) - \varrho^*_{v_{\theta}} = w(y(\theta)) - g(y(\theta), \theta) = 0$. Similarly, if $x \in S \backslash S^*(v_{\theta})$, then $x \in [0, y(\theta))$ and consequently $\varrho^*_{v_{\theta}} - w(x) \geqslant \varrho^*_{v_{\theta}} - w(y(\theta)) = g(y(\theta), \theta) - w(y(\theta)) = 0$. It follows that
\begin{align*}
 r(S^*(v_{\theta}), v_{\theta}) - r(S,v_{\theta}) 
\geqslant \frac{\thetamin}{1 + \thetamax} 
\left( \int_{[y(\theta), 1] \backslash S} ( w(x) - \varrho^*_{v_{\theta}} ) {\rm d} x 
   + \int_{S \backslash [y(\theta), 1]} ( \varrho^*_{v_{\theta}} - w(x) ) {\rm d}x
\right).
\end{align*}
Recall that $k_w = \inf_{y \in (0,1)} w'(y) > 0$. 
Since $\varrho^*_{v_{\theta}} = w(y(\theta))$, we have by the mean value theorem
\begin{align*}
w(x) - \varrho^*_{v_{\theta}} = w(x) - w(y(\theta)) \geqslant k_w (x - y(\theta)),
\end{align*}
for all $x \in [y(\theta), 1]$, 
and
\begin{align*}
\varrho^*_{v_{\theta}} - w(x) = w(y(\theta)) - w(x) \geqslant k_w (y(\theta) - x),
\end{align*}
for all $x \in [0, y(\theta))$. Upon combining the above, we arrive at the lower bound
\begin{align*}
 r(S^*(v_{\theta}), v_{\theta}) - r(S,v_{\theta}) 
\geqslant \frac{\thetamin k_w}{1 + \thetamax} 
\left( \int_{[y(\theta), 1] \backslash S} ( x - y(\theta) ) {\rm d} x 
   + \int_{S \backslash [y(\theta), 1]} ( y(\theta) - x ) {\rm d}x
\right). 
\end{align*}
Let $m_1 := | [y(\theta), 1] \cap S^c|$ and $m_2 := | [0, y(\theta)) \cap S |$. 
Observe that
\begin{align*}
    \int_{[y(\theta), 1] \backslash S} ( x - y(\theta) ) {\rm d} x 
&\geqslant  \int_{y(\theta)}^{y(\theta) + m_1}( x - y(\theta) ) {\rm d} x = \tfrac{1}{2} m_1^2,\\
			\int_{S \backslash [y(\theta), 1]} ( y(\theta) - x ) {\rm d}x
&\geqslant \int_{y(\theta) - m_2}^{y(\theta)} ( y(\theta) - x ) {\rm d}x = \tfrac{1}{2} m_2^2.
\end{align*}
In addition, 
\begin{align*}
\psi^S - \psi(\theta) &= 
 | S \cap [0, y(\theta))| + |S \cap [y(\theta), 1]| 
- | S \cap [y(\theta), 1]| - |S^c \cap [y(\theta),1]| 
= m_2 - m_1,\\
m_1^2 + m_2^2 &\geqslant m_1^2 + m_2^2 - 2 m_1 m_2 = (m_1 - m_2)^2 = 
(\psi^S - \psi(\theta))^2. \end{align*}
From the above we conclude that our claim applies: for all $\theta \in \Theta$ and $S \in \cS$,
\begin{align*}
 r(S^*(v_{\theta}), v_{\theta}) - r(S,v_{\theta}) 
\geqslant \frac{\thetamin k_w / 2}{1 + \thetamax} (\psi^S - \psi(\theta))^2.
\end{align*}

\noindent $\rhd$ {Step 2.} 
For $S \in \cS$ and $\theta \in \Theta$, let 
$Z^S_{\theta}$ be the random variable with support $[0, 2]$ and probability density function
\[ f_S(z \mid \theta) := 
\left\{
\begin{array}{ll}
\frac{\displaystyle v_{\theta}(z)}{\displaystyle 1 + \int_{S} v_{\theta}(\xi) {\rm d} \xi} & \text{ if } z \in S, \\
&\\
\frac{\displaystyle | [0, 2] \backslash S|^{-1} }{\displaystyle 1 + \int_{S} v_{\theta}(\xi) {\rm d} \xi} & \text{ if } z \in [0, 2] \backslash S. 
\end{array}
\right.
\]
Observe that, when $v = v_{\theta}$, $X^S$ is in distribution equal to the random variable that
equals $Z^S_{\theta}$ if $Z^S_{\theta} \in S$ and equals $\emptyset$ if $Z^S_{\theta} \in [0, 2] \backslash S$. Hence, for each $t \in \{1, \ldots, T\}$ there is a function $\pi_t: [0, 2]^{t-1} \ra \cS$ such that $S_t = \pi_t(Z_1, \ldots, Z_t)$ a.s., where $Z_t \stackrel{\rm d}{=} Z^{S_t}_{\theta}$ for all $t = 1, \ldots, T$, and where we write $\pi_{1}(\emptyset) := S_1$. In other words: to prove the regret lower bound we may assume that assortments are a function of the observations $Z_1,Z_2, \ldots$ instead of the purchase observations $X_1, X_2, \ldots$. 

Let $t \in \{1, \ldots, T\}$ and let $\cZ := [0, 2]^t$. The probability density function of $(Z_1, \ldots, Z_t)$ is equal to 
\begin{align*}
f({\boldsymbol z}_t \mid \theta) &= \prod_{i=1}^t f_{\pi_i({\boldsymbol z}_{i-1})}(z_i \mid \theta),
\end{align*}
for all ${\boldsymbol z}_t = (z_1, \ldots, z_t) \in \cZ$, where we write ${\boldsymbol z}_{i-1} = (z_1, \ldots, z_{i-1})$ for the first $i-1$ components of ${\boldsymbol z}_t$, for all $i=1, \ldots, t$, and ${\boldsymbol z}_0 := \emptyset$. 
We have 
\begin{align*} 
\frac{\rm d}{{\rm d} \theta} \log f({\boldsymbol z}_t \mid \theta) 
&= \sum_{i=1}^{t} \frac{\rm d}{{\rm d} \theta} \log f_{\pi_i({\boldsymbol z}_{i-1})}(z_i \mid \theta) \\
&= \sum_{i=1}^{t} \frac{\rm d}{{\rm d} \theta} \left\{ \log \theta \cdot {\boldsymbol 1}\{z_i \in \pi_i({\boldsymbol z}_{i-1}) \}
- \log\left( 1 + \theta\int_{\pi_i({\boldsymbol z}_{i-1}) } {\rm d} \xi \right) \right\} \\
&= \sum_{i=1}^{t} \theta^{-1} {\boldsymbol 1}\{z_i \in \pi_i({\boldsymbol z}_{i-1}) \}
- \frac{ | \pi_i({\boldsymbol z}_{i-1}) | }{1 + \theta | \pi_i({\boldsymbol z}_{i-1}) |},
\end{align*}
and
\begin{align*} 
-\frac{{\rm d}^2}{{\rm d} \theta^2} \log f({\boldsymbol z}_t \,|\, \theta) 
&= \sum_{i=1}^{t} \theta^{-2} {\boldsymbol 1}\{z_i \in \pi_i({\boldsymbol z}_{i-1}) \}
- \frac{ | \pi_i({\boldsymbol z}_{i-1}) |^2 }{(1 + \theta | \pi_i({\boldsymbol z}_{i-1}) | )^2} 
\leqslant \frac{t}{\underline{v}^2},
\end{align*}
since $\thetamin \geqslant \underline{v}$. 
By taking expectation, it follows that the Fisher information corresponding to ${\boldsymbol Z}_t$ satisfies
 \[\mathcal{I}_t(\theta) = \E\left[ -\frac{{\rm d}^2}{{\rm d} \theta^2} \log f({\boldsymbol Z}_t \mid \theta) \right] \leqslant \frac{t}{\underline{v}^2}.\]
The Fisher information $\mathcal{I}(\lambda)$ corresponding to the density $\lambda(\cdot)$ equals
\[ 
\int_{\thetamin}^{\thetamax} \left(\frac{\rm d}{{\rm d} \theta} \log \lambda(\theta) \right)^2 \lambda(\theta) {\rm d} \theta = 
\frac{4 \pi^2}{(\thetamax - \thetamin)^2} = \frac{\pi^2}{(\vmax - c_0)^2 }.
\] 
For each $\theta \in \Theta$, $y(\theta)$ is the unique solution to $g(y, \theta) - w(y) = 0$. 
By the Implicit Function theorem, the derivative $\psi'(\theta)$ of $\psi(\theta)$ exists and is equal to
\begin{align*}
\psi'(\theta) &= - \frac{\rm d}{{\rm d} \theta} y(\theta) 
= \frac{ \frac{{\rm d} g}{{\rm d} \theta} (y(\theta), \theta) }{ \frac{{\rm d} g}{{\rm d} y} (y(\theta),\theta) - \frac{{\rm d} w}{{\rm d} y}y(\theta)\} } \\
&= -\frac{ (1 + \theta (1 - y(\theta)))^{-2} }{ w'(y(\theta))} 
\leqslant -\frac{1}{ \max\{ w'(y) : y \in (0,1)\}} =: \kappa_1;
\end{align*}
for the last step, observe that $w$ being continuously differentiable implies that $\max\{w'(y) : y \in (0,1)\}$ is finite. 
Now, let $\theta$ be a random variable with probability density function $\lambda(\cdot)$; we denote by ${\mathbb E}_\lambda[\cdot]$ expectation with respect to this density. 
Let $\psi_t := \psi^{S_{t+1}}$. Now, we are in a position to apply the Van Trees inequality, in particular the form featuring in \cite{Gill1995}. Using the notation used there, their Equation (4) directly yields (realizing that $\psi'(\theta)\leqslant \kappa_1<0$ uniformly in $\theta$)
\begin{align*}
\E_\lambda[ (\psi_t - \psi(\theta))^2] 
&\geqslant \frac{ \E_\lambda[ \psi'(\theta)]^2 }{\E_\lambda[ \mathcal{I}_t(\theta)] + \mathcal{I}(\lambda)} 
\geqslant \frac{ \kappa_1^2 }{ t / \underline{v}^2 + \pi^2 / (\vmax - c_0)^2}.
\end{align*}
With this lower bound essentially behaving as $t^{-1}$, the corresponding partial sums (up to the $T$-th term) grow as $\log T$, as desired. More formally, summing over all $t=1,\ldots, T-1$, we obtain, applying the lower bound established in Step 1, 
\begin{align*}
\Delta_{\pi}(T) &= \sup_{v \in \cV} \Delta_{\pi}(T, v)
\geqslant \E_\lambda[ \Delta_{\pi}(T, v_{\theta}) ]\\
&\geqslant \kappa_0 \sum_{t=1}^{T-1} \E_\lambda[ (\psi_t - \psi(\theta))^2] 
\geqslant \kappa_0 \sum_{t=1}^{T-1} \frac{ \kappa_1^2 \underline{v}^2}{ t + \pi^2 \underline{v}^2 / (\vmax - c_0)^2} 
\geqslant \Cmin \log T,
\end{align*}
where $\Cmin := \kappa_0 \kappa_1^2 \underline{v}^2 / (1 + \pi^2 \underline{v}^2 / (\vmax - c_0)^2)>0$,
and where we used that \[\sum_{t=1}^{T-1} (t+a)^{-1} \geqslant 
(1+a)^{-1} \sum_{t=1}^{T-1} t^{-1} \geqslant (1+a)^{-1} 
\log T\] for all $T \geqslant 2$ and $a \geqslant 0$.
\hfill$\Box$

\section*{Appendix B: Mathematical proofs of Section \ref{sec:capacitated}}
\label{app:capacitated}


\subsection*{B.1. Proofs of the results in Section \ref{sec:capacitatedfullinfo}}

\subsubsection*{Proof of Lemma \ref{lem:perrho}.} 
We start the proof by the general remark that it is clear that the optimizing $S$ should only contain $x$ such that $h(x,\varrho)\geqslant 0$, i.e., $x\in W_\varrho$.

First consider case ($i$), i.e., $\text{vol}(W_\varrho)\leqslant c.$ Including in $S$ {\it all} $x\in W_\varrho$ thus leads to a set in $\cS$. Since $h(x,\varrho)< 0$ for $x\notin W_\varrho$, we conclude that the maximum of $\mathcal{I}(S,\varrho)$ over sets in $\cS$ is attained by $S= W_\varrho.$

Now, we consider case ($ii$), i.e., $\text{vol}(W_\varrho)> c$; this means that we should select the subset of $W_\varrho$ that maximizes $\mathcal{I}(S,\varrho)$. 
Our construction makes use of the following technical properties of $m_{\varrho}(l)$; their proofs will be given below. 
\begin{lemma} 
Let $\varrho\in[0,1]$. Then $m_\varrho(\ell)$ is non-increasing and left-continuous in $\ell$, as well as $m_\varrho(\ell)\to 0$ as $\ell\to\infty.$
\label{lem:mfunction}
\end{lemma}
We first concentrate on claim (1). To this end, observe that $m_\varrho(0)=\text{vol}(L_\varrho(0))=\text{vol}(W_\varrho) \geqslant c>0$. In addition, by virtue of Lemma \ref{lem:mfunction}, $m_\varrho(\ell)\to0$ as $\ell\to\infty.$ Hence, the set of $\ell\geqslant 0$ such that $m_\varrho(\ell)\geqslant c$ is nonempty and bounded, so that its supremum exists; because of the left-continuity that has been established in Lemma \ref{lem:mfunction} the supremum is actually attained (and hence is a maximum). This proves the first claim of ($ii$). 

We now consider the second claim of ($ii$). The intuitive idea is that we start with $S=\emptyset$, and that we keep adding $x$ from $W_\varrho$ to $S$ that have the highest value of $h(x,\varrho)$, until $\text{vol}(S)=c$; at that point $S$ consists of $x$ such that $h(x,\varrho)\geqslant \ell_\varrho$.
Bearing in mind, though, that the set of $x\in[0,1]$ such that $h(x,\varrho)$ equals some given value may have positive Lebesgue measure, there may be still a degree of freedom, which is reflected in the way the set $L_\varrho^\circlearrowleft$ has been defined. 

The formal argumentation is as follows. First we prove that $\text{vol}(L_\varrho^+)\leqslant c$:
as a consequence of the continuity of the Lebesgue measure and the fact that $m_\varrho(\ell)$ is non-increasing in $\ell$,
\begin{align*}
\text{vol}(L_\varrho^+)&=\text{vol}\left(\bigcup_{k=1}^\infty L_\varrho(\ell_\varrho+1/k)\right)
=\text{vol}\left(\lim_{n\to\infty}\bigcup_{k=1}^n L_\varrho(\ell_\varrho+1/k)\right)\\
&=\lim_{n\to\infty}
\text{vol}\left(\bigcup_{k=1}^n L_\varrho(\ell_\varrho+1/k)\right)=\lim_{n\to\infty}
\text{vol}\left(L_\varrho(\ell_\varrho+1/n)\right)=\lim_{n\to\infty}m_\varrho(\ell_\varrho+1/n)\leqslant c.
\end{align*}
Hence, there exists a set $L_\varrho^\circlearrowleft$ that is a (possibly empty) subset of $L_\varrho^=$ and that is such that $\text{vol}(S)=\text{vol}(L_\varrho^+)+ 
\text{vol}(L_\varrho^\circlearrowleft)=c.$

The next objective is to prove that $S=L_\varrho^+\cup L_\varrho^\circlearrowleft$ maximizes $\mathcal{I}(\,\cdot\,,\varrho)$ over sets in $\cS$. Take an arbitrary $R\in \cS$. Since $\text{vol}(S)=c$, we know that
\[c = \text{vol}(S) = \text{vol}(S\cap R) + \text{vol}(S\backslash R)= \text{vol}(R) - \text{vol}(R\backslash S) + \text{vol}(S\backslash R)\]
and since $\text{vol}(R)\leqslant c$, we obtain $\text{vol}(S\backslash R)\geqslant \text{vol}(R\backslash S)$. Now, since $x\in S$ implies $h(x,\varrho)\geqslant \ell_\varrho$ and $x\in R\backslash S$ implies $h(x,\varrho)\leqslant \ell_\varrho$ we conclude
\[\mathcal{I}(S,\varrho) - \mathcal{I}(R,\varrho) = \mathcal{I}(S\backslash R,\varrho)-\mathcal{I}(R\backslash S,\varrho) \geqslant \ell_\varrho\big(\text{vol}(S\backslash R) - \text{vol}(R\backslash S)\big) \geqslant 0.\]
This proves the second claim of ($ii$). 
\hfill$\Box$
\vspace{3mm}


\subsubsection*{Proof of Lemma \ref{lem:mfunction}.} 
The set $L_{\varrho}(\ell)$ is non-increasing in $\ell$, hence so is the function $m_\varrho(\ell)$. 
The next step is to prove that $m_\varrho(\ell)$ is left-continuous. To this end, let $\ell_n$ be a strictly increasing sequence converging to $\ell<\infty$ as $n\to\infty.$ As we have seen, $L_\varrho(\ell_n)\supseteq L_\varrho(\ell)$, and therefore
\begin{align*}
m_\varrho(\ell)- m_\varrho(\ell_n)=\text{vol}\big(\{x\in[0,1]: h(x,\varrho)\in[\ell_n,\ell)\}\big)
=\sum_{k=n}^\infty \text{vol}\big(\{x\in[0,1]: h(x,\varrho)\in[\ell_n,\ell_{n+1})\}\big).
\end{align*}
From the fact that the left-hand side is finite, it follows that the right-hand side is finite as well, implying left-continuity.

Along the same lines,
\begin{align*}
1=\text{vol}\big([0,1]\big) 
=\sum_{k=-\infty}^\infty \text{vol}\big(\{x\in[0,1]: h(x,\varrho)\in[k,k+1)\}\big).
\end{align*}
This entails that, with $n\to\infty$ along the integers,
\[\lim_{n\to\infty}m_\varrho(n) = \lim_{n\to\infty} \sum_{k=n}^\infty \text{vol}(\{x\in[0,1]: h(x,\varrho)\in[k,k+1)\})=0.\]
From the monotonicity of $m_\varrho(\ell)$, we also have that $m_\varrho(\ell)\to 0$ as $\ell\to\infty$ along the reals.
\hfill$\Box$
\vspace{3mm}

\subsubsection*{Proof of Proposition \ref{prop:fullinformationcapacitated}.}
Firstly, we show that there exists a unique solution to the fixed-point equation
\begin{equation}
g(\varrho)=\varrho,
\label{eq:fixedpoint}
\end{equation}
where $g(\varrho):=\mathcal{I}(S_\varrho,\varrho)$ for $\varrho\in[0,1]$. As the right-hand side of (\ref{eq:fixedpoint}) is strictly increasing in $\varrho$, it suffices to prove that $g(\cdot)$ is continuous and non-increasing in $\varrho$, and that $g(0)\geqslant 0$ and $g(1)=0.$ To this end, consider $0\leqslant\varrho_1\leqslant\varrho_2\leqslant 1$. Then, indeed, as $\mathcal{I}(S,\varrho)$ is non-increasing in $\varrho$ for any fixed $S\in \cS$, and recalling that $S_{\varrho_1}$ maximizes $\mathcal{I}(S,\varrho_1)$, 
\[g(\varrho_1)= \mathcal{I}(S_{\varrho_1},\varrho_1) \geqslant \mathcal{I}(S_{\varrho_2},\varrho_1)\geqslant \mathcal{I}(S_{\varrho_2},\varrho_2)= g(\varrho_2)
.\]
The next step is to prove that $g(\cdot)$ is continuous. Let $\varrho_1,\varrho_2\in[0,1].$ Then
\[\mathcal{I}(S_{\varrho_1},\varrho_1)-\mathcal{I}(S_{\varrho_2},\varrho_2)
\leqslant \mathcal{I}(S_{\varrho_1},\varrho_1) - \mathcal{I}(S_{\varrho_1},\varrho_2)
= (\varrho_2-\varrho_1)\int_{S_{\varrho_1}}v(x){\rm d}x
\leqslant |\varrho_1-\varrho_2|\int_{[0,1]}v(x){\rm d}x ,\]
where the first inequality is due to the fact that $S_{\varrho_2}$ maximizes $\mathcal{I}(\,\cdot\,,\varrho_2)$. With the same token, the same upper bound applies when the roles of the $\varrho_1$ and $\varrho_2$ in the left-hand side are interchanged. It thus follows that $g(\cdot)$ is continuous; it is actually even Lipschitz continuous. 

Obviously, $g(0)\geqslant 0$. Using that $\sup_{x\in[0,1]}w(x)\leqslant 1$, we also obtain
\[g(1)=\max_{S\in \cS}\int_Sv(x)(w(x)-1){\rm d}x=0.\]

Secondly, we show that $S_{\varrho^{*}}$ has the maximum expected revenue over all sets in $\cS$. Note that, since $g(\varrho^{*}) = \varrho^{*}$, it follows that $r(S_{\varrho^{*}}) = \varrho^{*}$. Hence, as we proceed from (\ref{eq:rewrittenSb}) by invoking Lemma \ref{lem:perrho}, we obtain
\[\max\left\{\varrho\in[0,1]:\max_{S\in \cS} \mathcal{I}(S_\varrho,\varrho)\geqslant \varrho\right\} = \max\left\{\varrho\in[0,1]:g(\varrho)\geqslant\varrho\right\}= \varrho^{*} = r(S_{\varrho^{*}}).\]
\hfill$\Box$
\vspace{3mm}

\subsection*{B.2. Proofs of the results in Section \ref{sec:capacitatedupperbound}}

\subsubsection*{Proof of Proposition \ref{prop:bridge}.}
In addition to optimal assortments $S^*$ and $\check{S}$ as in \eqref{eq:SstarandcheckS}, we define $S^p$ as the optimal assortment under $\check{v}$ and $w$, that is,
\[r(S^p,\check{v},w) = \max_{S\in\cS} r(S,\check{v},w).\]
This assortment $S^p$ plays a pivotal role as we break up the left-hand side of \eqref{eq:L1diff} as follows:
\begin{align}{}
r(S^*,v,w) - r(\check{S},\check{v},\check{w}) =&\: r(S^*,v,w) - r(S^p,\check{v},w) \:+ \label{eq:appsplit1}\\
&\: r(S^p,\check{v},w) - r(\check{S},\check{v},\check{w})\label{eq:appsplit2}.
\end{align}
We start by bounding the right-hand side of \eqref{eq:appsplit1} from above. Define
\[\mathcal{I}(S,\varrho)=\int_{S}v(x)(w(x)-\varrho){\rm d}x\qquad\text{and}\qquad \mathcal{I}^p(S,\varrho):=\int_{S}\check{v}(x)(w(x)-\varrho){\rm d}x\]
for $S\in\cS$ and $\varrho\in[-\vmax,1]$. Note that these definitions allow for negative values of $\varrho$ (as opposed to \eqref{eq:innermaximization}). 
Next, denote the $L_1$-distance between $v$ and $\hat{v}$ as $\delta:= |\!|v-\tilde{v}|\!|_1$. For $\varrho\in[-\vmax,1]$, let $S_\varrho$ be the maximizer of $\mathcal{I}(\,\cdot\,,\varrho)$ over $\cS$ and let $S^p_\varrho$ be the maximizer of $\mathcal{I}^p(\,\cdot\,,\varrho)$ over $\cS$, that is,
\[\mathcal{I}(S_\varrho,\varrho) = \max_{S\in\cS} \mathcal{I}(S,\varrho) \qquad\text{and}\qquad \mathcal{I}^p(S^p_\varrho,\varrho) = \max_{S\in\cS} \mathcal{I}^p(S,\varrho).\]
Then let $\varrho^{*}$ and $\varrho^p$ solve the fixed-point equations
\[\varrho =\mathcal{I}(S_\varrho,\varrho)\qquad\text{and}\qquad \varrho = \mathcal{I}^p(S^p_\varrho,\varrho),\]
respectively. Note that $S^p_{\varrho^p}$ is an optimal assortment under $\check{v}$ and $w$ by Proposition \ref{prop:fullinformationcapacitated}. Hence, we may assume that $S^p=S^p_{\varrho^p}$. Also, we have $0\leqslant w(x)-\varrho^*\leqslant 1$ for all $x\in S^*$ and therefore,
\begin{align*}
\mathcal{I}^p(S^*,\varrho^*) - \mathcal{I}(S^*,\varrho^*) &= \int_{S^*}\check{v}(x)(w(x)-\varrho^*){\rm d}x - \int_{S^*}v(x)(w(x)-\varrho^*){\rm d}x \\
&\leqslant \int_{S^*} |v(x) - \check{v}(x)|{\rm d}x\leqslant \delta.
\end{align*}
Now, we find that
\begin{equation*}
\mathcal{I}^p(S^*,\varrho^*-\delta) \geqslant \mathcal{I}^p(S^*,\varrho^*) \geqslant \mathcal{I}(S^*,\varrho^*)- \delta = \varrho^* - \delta. 
\end{equation*}
Hence, there exists an $S\in\cS$ such that $\mathcal{I}^p(S,\varrho^*-\delta) \geqslant \varrho^* - \delta$, which by \eqref{eq:rewrittenSb} entails $\varrho^p\geqslant \varrho^* -\delta$. Thus, \eqref{eq:appsplit1} is bounded from above as
\[ r(S^*,v,w) - r(S^p,\check{v},w)\leqslant |\!|v-\check{v}|\!|_1.\]
Bounding \eqref{eq:appsplit2} from above follows in almost an identical manner, but instead of $0\leqslant w(x)-\varrho^*\leqslant 1$ we now use $0\leqslant \check{v}(x) \leqslant \vmax$. As a result, we conclude that
\[r(S^p,\check{v},w) - r(\check{S},\check{v},\check{w})\leqslant \vmax\,|\!|w-\check{w}|\!|_1.\]
Combining the above concludes the proof.
\hfill$\Box$
\vspace{3mm}

\subsubsection*{Proof of Lemma \ref{lem:optimizingoversubset}.}
First, let $\varrho^d = r(S^d,\check{v},\check{w})$ and define the sets $\check{\mathcal{M}}$ and $\mathcal{M}^d$ as arguments of maxima as
\[\check{\mathcal{M}} := \argmax_{S\in\cS}\int_S \check{v}(x)(\check{w}(x)-\varrho^d)\d x \quad\text{ and }\quad \mathcal{M}^d := \argmax_{S\in\mathcal{A}_K}\int_S \check{v}(x)(\check{w}(x)-\varrho^d)\d x,\qquad \varrho \in [0,1].\]
Note that since $\mathcal{A}_K\subset \cS$, we know for any $S_1\in\check{\mathcal{M}}$ and $S_2\in\mathcal{M}^d$ that
\begin{equation}
\int_{S_1}\check{v}(x)(\check{w}(x) -\varrho^d)\d x \geqslant \int_{S_2} \check{v}(x)(\check{w}(x) -\varrho^d)\d x\geqslant 0.
\label{eq:orderinM}
\end{equation}
Since $\varrho^d\geqslant r(S,\check{v},\check{w})$ for any $S\in\mathcal{A}_K$, it also holds for $S\in\mathcal{M}^d$ that
\begin{equation}
\varrho^d \geqslant \int_{S} \check{v}(x)(\check{w}(x) -\varrho^d)\d x.
\label{eq:rhodfixedpoint}
\end{equation}
Then, for any $S_1\in\check{\mathcal{M}}$ and $S_2\in\mathcal{M}^d$, it follows that
\begin{align}
r(\check{S},\check{v},\check{w}) - r(S^d,\check{v},\check{w}) & = \frac{\int_{\check{S}} \check{v}(x)\check{w}(x) \d x}{1+\int_{\check{S}} \check{v}(x)\check{w}(x) \d x} - \varrho^d \notag\\
& = \frac{1}{1+\int_{\check{S}} \check{v}(x)\check{w}(x) \d x}\left(\int_{\check{S}} \check{v}(x)(\check{w}(x) -\varrho^d)\d x - \varrho^d\right)\notag\\
& \leqslant^{(*)} \frac{1}{1+\int_{\check{S}} \check{v}(x)\check{w}(x) \d x} \left(\int_{S_1} \check{v}(x)(\check{w}(x) -\varrho^d)\d x - \varrho^d\right)\notag\\
& \leqslant^{(**)} \frac{1}{1+\int_{\check{S}} \check{v}(x)\check{w}(x) \d x}\left(\int_{S_1} \check{v}(x)(\check{w}(x) -\varrho^d)\d x - \int_{S_2} \check{v}(x)(\check{w}(x) -\varrho^d)\d x\right)\notag\\
& \leqslant^{(***)} \int_{S_1} \check{v}(x)(\check{w}(x) -\varrho^d)\d x - \int_{S_2} \check{v}(x)(\check{w}(x) -\varrho^d)\d x. \label{eq:battleofgiants}
\end{align}
Here at $(*)$ we use that $S_1\in\check{\mathcal{M}}$, at $(**)$ we use \eqref{eq:rhodfixedpoint} and $(***)$ holds because of \eqref{eq:orderinM}.

Now, we claim  there exist assortments $S_1\in \check{\mathcal{M}}$ and $S_2\in\mathcal{M}^d$, such that $S_2\subseteq S_1$ and $\text{vol}(S_1\backslash S_2) \leqslant 1/N$. To this end, let $y_i\in B_i$ for $i\in[N]$ and define $h_i$ as
\[h_i:=\check{v}(y_{i}) \big(\check{w}(y_{i}) - \varrho^d\big),\quad i\in [N].\]
In addition, let $\sigma:[N]\to[N]$ be an ordering,
such that,
\[h_{\sigma(1)}\geqslant\ldots\geqslant h_{\sigma(N)},\]
where we break ties arbitrarily. As in Lemma \ref{lem:perrho}, we first consider the case that $\text{vol}(W_{\varrho^d})\leqslant c$. Then we know by Lemma \ref{lem:perrho} that $W_{\varrho^d} \in \check{\mathcal{M}}$. Since $\check{w}$ is constant on each bin, there exists an integer $n$ such that $\text{vol}(W_{\varrho^d}) = n/N$. If $n/N\leqslant c$, then $n\leqslant K$ and hence $W_\varrho \in \mathcal{M}^d$ as well. This concludes the claim for $\text{vol}(W_{\varrho^d})\leqslant c$. Next, we consider the case that $\text{vol}(W_{\varrho^d})> c\geqslant K/N$. Then $h_{\sigma(K)}\geqslant 0$ and
\[S_1:= \bigcup_{i=1}^K B_{\sigma(i)} \in \mathcal{M}^d.\]
In addition, note that $h_{\sigma(K+1)}\geqslant 0$ as well as $K<N$ since $c<1$ and define
\[R:=\left[\frac{\sigma(K+1)-1}{N},\frac{\sigma(K+1)-1}{N} + c - \frac{K}{N}\right)\subset B_{\sigma(K+1).}\]
Recall the definitions from Lemma \ref{lem:perrho} and note that, as $\check{v}$ and $\check{w}$ are constant on each bin,
\[m_{\varrho^d}(\ell) = \frac{i}{N},\quad \ell\in (h_{\sigma(i+1)},h_{\sigma(i)}]\cap[0,\infty),\:\:i=1,\ldots,N-1.\]
As a result, $c=K/N$ implies $\ell_{\varrho^d}=h_{\sigma(K)}$ and $R=\emptyset$, and $c>K/N$ implies $\ell_{\varrho^d}=h_{\sigma(K+1)}$. Either way, it follows that
\[L^+_{\varrho^d}\subseteq S_1\subseteq S_1\cup R \subseteq L^+_{\varrho^d}\cup L^=_{\varrho^d}.\]
Since $\text{vol}(S_1\cup R)=c$, it follows from Lemma \ref{lem:perrho} that $S_2:=S_1\cup R \in \check{\mathcal{M}}$. This concludes the claim for $\text{vol}(W_{\varrho^d})> c$.

From \eqref{eq:battleofgiants}, the shown claim and the fact that $\check{w}(x)-\varrho^d\leqslant 1$, it follows that
\[r(\check{S},\check{v},\check{w}) - r(S^d,\check{v},\check{w})\leqslant \frac{\vmax}{N}.\]
\hfill$\Box$
\vspace{3mm}

\subsubsection*{Proof of Lemma \ref{lem:mistranslatingSt}.}
Since $S\in\mathcal{A}_K$, we know that 
\[\int_S v(x)\d x = \int_S \check{v}(x) \d x.\]
Therefore,
\begin{align*}
r(S,\check{v},\check{w}) - r(S,v,w) &= \frac{1}{1+\int_S v(x)\d x}\int_S\big(\check{v}(x)\check{w}(x) - v(x)w(x)\big)\d x\\
& \leqslant |\!|vw - \check{v}\check{w}|\!|_1  = |\!|vw -\check{v}w + \check{v}w- \check{v}\check{w}|\!|_1 \\
& \leqslant |\!|v-\check{v}|\!|_1 + \vmax\,|\!|w-\check{w}|\!|_1,
\end{align*}
where we have used that $w(x)\leqslant 1$ and $\check{v}(x)\leqslant\vmax$ for all $x\in[0,1]$.
\hfill$\Box$
\vspace{3mm}

\subsubsection*{Proof of Theorem \ref{th:upperboundc<1}.}
We start by showing that $|\!|v-\check{v}|\!|_1$ and $|\!|w-\check{w}|\!|_1$ are of order $1/N$. For $i\in[N]$, denote the constant $b_i = \check{v}(x)$ for some $x\in B_i$. Note that $b_i=\check{v}(x)$ for all $x\in B_i$ and that
\[|\!|v-\check{v}|\!|_1 = \int_0^1 |v(x)-\check{v}(x)|\d x = \sum_{i=1}^N\int_{B_i} |v(x)-b_i| \d x.\]
By the Mean Value Theorem, for every $i\in[N]$, there exists a $c_i$ in the closure of $B_i$ such that $v(c_i) = b_i$. Hence,
\begin{equation}
|\!|v-\check{v}|\!|_1 =\sum_{i=1}^N\int_{B_i} |v(x)-b_i| \d x = \sum_{i=1}^N\int_{B_i} |v(x)-v(c_i)| \d x \leqslant L \sum_{i=1}^N\int_{B_i} |x-c_i| \d x \leqslant L \sum_{i=1}^N \frac{1}{2N^2}\leqslant \frac{L}{2N},
\label{eq:L1discretizationerror}
\end{equation}
where $L:= \sup_{x\in[0,1]}|v'(x)|$. Likewise,
\[|\!|w-\check{w}|\!|_1 \leqslant \frac{Q}{2N},\]
where $Q:= \sup_{x\in[0,1]} |w'(x)|$.

Now, let $\Delta_{\rm UCB}(T)$ denote the cumulative regret of UCB within the discrete MNL model. Recall that the preference parameters $v_1,\ldots,v_N$ satisfy 
\[v_i = \int_{B_i}v(x)\d x,\qquad i\in[0,1],\]
and the parameters $w_1,\ldots,w_N$ satisfy
\[w_i = N\int_{B_i}v(x)\d x,\qquad i\in[0,1].\]
Let $S=\bigcup_{i\in D} B_i\in\mathcal{A}_K$ for some $D\subset[N]$. Then the probability under $v$, as well as under $\check{v}$, that a purchase from assortment $S$ lies in $B_i$ is
\[\P(X^S\in B_i) = \frac{v_i}{1+\sum_{i\in D}v_i},\]
In addition, the expected profit of assortment $S\in\mathcal{A}_K$ under $\check{v}$ and $\check{w}$ is
\[r(S,\check{v},\check{w}) = \frac{\sum_{i\in D}v_iw_i}{1+\sum_{i\in D}v_i}.\]
As a result, if $S_1,\ldots,S_T$ denote the offered assortment under DUCB($N$) and $S^d$ as in \eqref{eq:checkSandSd}, then
\[\sum_{t=1}^T\E_\pi\left[r(S^d,\check{v},\check{w}) - r(S_t,\check{v},\check{w})\right] = \Delta_{\rm UCB}(T).\]
Following the steps of \eqref{eq:split1}--\eqref{eq:split4}, in combination with the above and Proposition \ref{prop:bridge}, Lemma \ref{lem:optimizingoversubset} and Lemma \ref{lem:mistranslatingSt}, we find that, with $C_1:= L+\vmax(Q+1)$,
\[\Delta_{\pi}(T)\leqslant C_1\frac{T}{N} + \Delta_{\rm UCB}(T).\]

By our choice of $\gamma$, we know that $\floor{\gamma}\geqslant 1/c$. Hence, $N\geqslant 1/c\geqslant 1$ and $K\geqslant 1$. Second, $\gamma$ is chosen such that $\vmax\leqslant N$ and therefore $v_i\leqslant 1$ for all $i\in[N]$. By Theorem 1 from \cite{Agrawal2019}, there exists constants $C_2$ and $C_3$ such that
\[\Delta_{\rm UCB}(T)\leqslant C_2\sqrt{NT\log NT} + C_3 N\log^2 NT.\]
Since $N\leqslant \gamma T^{1/3}$, it follows that
\[\log NT \leqslant \log \gamma T^{4/3} = \frac{4}{3}\log T + \log \gamma\leqslant C_4\log T,\]
where
$C_4:= \frac{4}{3} +{\log \gamma}/{\log 2}.$
Hence,
\[\Delta_{\rm UCB}(T)\leqslant C_2\sqrt{\gamma C_4}\sqrt{T^{4/3}\log T} + \gamma C_3 C_4^2\,T^{1/3}\log^2 T .\]
Now we note that 
\[\log T \leqslant \frac{9}{2e}T^{2/9}\]
and therefore
\[T^{1/3}\log^2 T\leqslant \left(\frac{9}{2e}\right)^{3/2} T^{2/3}(\log T)^{1/2}.\]
Thus we obtain that $\Delta_{\rm UCB}(T)\leqslant C_5 \,T^{2/3} (\log T)^{1/2}$,
where
\[C_5:=C_2\sqrt{\gamma C_4} + \left(\frac{9}{2e}\right)^{3/2}\gamma C_3 C_4^2.\]
Next, we point out that $N\geqslant (\gamma-1)T^{1/3}$ with $\gamma\geqslant 2$. Thus,
\[\frac{T}{N} \leqslant \frac{1}{\gamma-1} T^{2/3}\leqslant \frac{1}{(\gamma-1)(\log 2)^{1/2}} T^{2/3} (\log T)^{1/2}.\]
From this we conclude that
\[\Delta_{\pi}(T)\leqslant C_1\frac{T}{N} + C_5 \,T^{2/3} (\log T)^{1/2}\leqslant \Cmax\,T^{2/3}(\log T)^{1/2},\]
where
\[\Cmax := \frac{C_1}{(\gamma-1)(\log 2)^{1/2}} + C_5.\]
\hfill$\Box$
\vspace{3mm}

\subsection*{B.3. Proofs of the results in Section \ref{sec:capacitatedlowerbound}}

Before stating the proofs of the results in Section \ref{sec:capacitatedlowerbound}, we recollect the notations and concepts introduced in that section. 
Let $c\in(0,\tfrac{1}{4}]$, $s=0.8c$, $\delta=\tfrac{1}{2}$ and $\sigma=0.3$. Let $K \geqslant 2$ be an integer, chosen at the end of the proof of Theorem \ref{th:lowerboundc<1}. Furthermore, for all $x \in [0,1]$, $i \in \{1, \ldots, N_K\}$, and $I \subseteq \{1,\ldots,N_K\}$, let
\begin{align*}
  N_K&=\floor{K/c}, & [N_K]&= \{1,\ldots,N_K\}, \\
  \mathcal{D}_K&=\{I\subseteq[N_K]:|I|=K\}, & B_i&= \left[c\frac{i-1}{K},c\frac{i}{K}\right), \\
  w(x) &= (1-s)\frac{1-\delta}{1-\delta x}+s, & v_0(x) &= \frac{s}{c(w(x)-s)}, \\
  b(x)&=\frac{1}{\sigma\sqrt{2\pi}} e^{-x^2/2\sigma^2}, & \phi_i(x)&= \frac{2Kx}{c} -2i + 1, \\
  \tau_i(x)& = \frac{c}{K}b\big(\phi_i(x)\big), & \beta & = \frac{c}{K}\frac{1}{\sigma\sqrt{2\pi}}\sum_{n\in\Z}e^{-(2n-1)^2/2\sigma^2}, \\
  \epsilon_I(x) &= \sum_{i\in I}\tau_i(x) - \beta, & v_I(x) &= v_0(x)\big(1+\epsilon_I(x)\big).
\end{align*}
In addition, we use the following notation throughout this section. For $I \in \mathcal{D}_K$ we write 
\[ I^{\dagger} := \bigcup_{i \in I} B_i. \]
Furthermore, we define the following quantities.
\begin{align*}
&H := \frac{1}{\sigma\sqrt{2\pi}}\sum_{n\in\Z}e^{-2n^2/\sigma^2}, \\
&L := \frac{1}{\sigma\sqrt{2\pi}}\sum_{n\in\Z}e^{-(2n-1)^2/2\sigma^2}\qquad\text{and} \\
&P := \P(-1/\sigma\leqslant Z\leqslant 1/\sigma),
\end{align*}
where $Z\sim N(0,1)$. Observe that 
$\beta = L {c}/{K}.$

We proceed by stating two preliminary lemmas that will be used throughout the proofs.
Lemma~\ref{lem:preliminaries} contains a number of inequalities related to the quantities defined above, and Lemma \ref{lem:anicecoincidence} shows that the optimal expected profit under $v_0$ is precisely equal to $s$. The proof of these lemmas is given below. 
\begin{lemma} \label{lem:preliminaries}
Let $I\subseteq [N_K]$. Then
\begin{enumerate}
\item[\rm ($i$)] for any $x\in[0,1]$, it holds that $\ds \sum_{i\in I}\tau_i(x)\leqslant H\frac{c}{K}$.
\item[\rm ($ii$)] for any $S\in \cS$ and $\beta >0$, it holds that 
\begin{itemize}
\item[\rm 1.] $\ds\int_{S}v_I(x){\rm d}x\leqslant\frac{s}{(1-s)(1-\delta)}\left(1+H\right)$ and 
\item[\rm 2.] $\ds\int_{S}(v_I(x))^2{\rm d}x\leqslant \frac{s^2}{c(1-s)^2(1-\delta)^2}\left(1+H\right)^2$,
\end{itemize}
\item[\rm ($iii$)] for 
$x\notin I^\dag$, it holds that $\ds \sum_{i\in I}\tau_i(x) \leqslant \beta$,
\item[\rm ($iv$)] if $|I|=K$ and $S\in \cS$, it holds that ${\rm vol}(I^\dag\backslash S)\geqslant {\rm vol}(S\backslash I^\dag)$,
\item[\rm ($v$)] for all $i\in[N]$;
\begin{itemize}
\item[\rm 1.] $\displaystyle\frac{c^2}{2K^2}P = \int_{B_i} \tau_i(x){\rm d}x \leqslant \int_0^1 \tau_i(x){\rm d}x \leqslant \frac{c^2}{2K^2}$ and 
\item[\rm 2.] $\displaystyle \int_0^1 (\tau_i(x))^2{\rm d}x \leqslant \frac{c^3}{4\sigma \sqrt{\pi}K^3}$,
\end{itemize}
\item[\rm ($vi$)] for any $i\in I$, $x\in B_i$ and $\beta'\geqslant \beta$, it holds that $|\epsilon_I(x;\beta')|\leqslant \tau_i(x) + \beta'$.
\end{enumerate}
\end{lemma}

\begin{lemma}\label{lem:anicecoincidence}
The optimal expected revenue under the preference function $v_0(\cdot)$ equals $s$: 
\begin{equation*}
  \max_{S\in\cS} r(S,v_0) = s.
\end{equation*}
\end{lemma}

\subsubsection*{Proof of Proposition \ref{prop:lower-bound-step-1}. }
Let 
\[C_1 := \frac{s(1-s)(1-\delta)}{c(1-s)(1-\delta)+cs}\qquad\text{and}\qquad C_2 : =\frac{s^2(c+2L)}{\big((1-s)(1-\delta)+s\big)(1-s)(1-\delta)}.\]
Let $\pi$ be a policy, $T \in \N$, and let $I \in \mathcal{D}_K$. Write $v(x) := v_I(x)$, and let $S^{*}$ denote an optimal assortment under $v$. Recall that $S^{*}$ also maximizes the inner maximization problem \eqref{eq:innermaximization} for $\varrho = \varrho^* = \max_{S\in\cS} r(S,v)$. Therefore,  
\begin{equation}
\int_{S^*}v(x)\big(w(x)-\varrho^*\big){\rm d}x \geqslant \int_{I^\dagger}v(x)\big(w(x)-\varrho^*\big){\rm d}x.
\label{eq:differentmax}
\end{equation}
Observe in addition that
\begin{equation}
  \int_S v_0(x){\rm d}x\leqslant \frac{s}{(1-s)(1-\delta)}.
  \label{eq:loosebound}
\end{equation}
It now follows that, for all $S \in \cS$, 
\begin{align}
r(S^*,v) - r(S,v) & = \varrho^* - \frac{\int_{S} v(x)w(x){\rm d}x}{1+\int_{S} v(x){\rm d}x} & \nonumber \\
& = \frac{1}{1+\int_S v(x){\rm d}x}\left(\varrho^*- \int_{S} v(x)\big(w(x)-\varrho^*\big){\rm d}x\right) \nonumber \\
& =^{(*)} \frac{1}{1+\int_S v(x){\rm d}x}\left(\int_{S^*} v(x)\big(w(x)-\varrho^*\big){\rm d}x- \int_{S} v(x)\big(w(x)-\varrho^*\big){\rm d}x\right) \nonumber \\
&\geqslant^{(**)}\frac{(1-s)(1-\delta)}{(1-s)(1-\delta)+s}\left(\int_{S^*} v(x)\big(w(x)-\varrho^*\big){\rm d}x- \int_{S} v(x)\big(w(x)-\varrho^*\big){\rm d}x\right) \nonumber \\
&\geqslant^{(***)}\frac{(1-s)(1-\delta)}{(1-s)(1-\delta)+s}\left(\int_{I^\dagger} v(x)\big(w(x)-\varrho^*\big){\rm d}x- \int_{S} v(x)\big(w(x)-\varrho^*\big){\rm d}x\right). \label{eq:inbrackets}
\end{align}
Here $(*)$ follows from $\varrho^* = \mathcal{I}(S^*,\varrho^*)$ by Proposition \ref{prop:fullinformationcapacitated}, $(**)$ follows by \eqref{eq:loosebound}, and $(***)$ follows by \eqref{eq:differentmax}. The terms within the large parentheses in \eqref{eq:inbrackets} can be bounded from below as
\begin{align*}
  \int_{I^\dagger} v(x)\big(w(x)-\varrho^*\big){\rm d}x- &\int_{S} v(x)\big(w(x)-\varrho^*\big){\rm d}x \\
  &=\int_{I^\dagger} v(x)\big(w(x)-s\big){\rm d}x- \int_{S} v(x)\big(w(x)-s\big){\rm d}x \\
  &\quad + (s-\varrho^*)\left(\int_{I^\dagger} v(x){\rm d}x- \int_{S} v(x){\rm d}x \right)\\
  &\geqslant^{(*)} \frac{s}{c}\int_{I^\dagger\backslash S}\big(1 + \epsilon_I(x)\big){\rm d}x- \frac{s}{c}\int_{S\backslash I^\dagger} \big(1 + \epsilon_I(x)\big){\rm d}x \\
  &\quad - |s - \varrho^*| \frac{2s}{(1-s)(1-\delta)},
\end{align*}
where at $(*)$ we use that by design $v(x)\big(w(x)-s\big) = \frac{s}{c}(1+\epsilon_I(x))$, together with inequality \eqref{eq:loosebound}. The absolute difference between $\varrho^*$ and $s$ can be bounded from above by the $L_1$-difference between $v$ and $v_0$, as follows. For $S\in\cS$ and $\varrho\in[0,1]$, let 
\[\mathcal{I}_0(S,\varrho) = \int_{S} v_0\big(w(x)-\varrho\big) \qquad\text{and}\qquad \mathcal{I}(S,\varrho) = \int_{S} v\big(w(x)-\varrho\big),\]
and let $S_0:=[0,c]$. As a consequence of Proposition \ref{prop:fullinformationcapacitated} and Lemma \ref{lem:anicecoincidence}, we obtain that $s = \mathcal{I}_0(S_0,s)$. Since $w(x)-\varrho^*\in[0,1]$ for all $x\in S^*$, we therefore know that 
\begin{align*}
  \mathcal{I}(S^*,\varrho^*) - \mathcal{I}_0(S^*,s)& \leqslant \int_{S^*}\big|v(x)-v_0(x)\big|{\rm d}x \\
  &\leqslant \int_{0}^1 \big|v(x)-v_0(x)\big|{\rm d}x = |\!|v-v_0|\!|_1.
\end{align*}
Furthermore,
\[\mathcal{I}_0(S^*,\varrho^*-|\!|v-v_0|\!|_1)\geqslant \mathcal{I}_0(S^*,\varrho^*) \geqslant \mathcal{I}(S^*,\varrho^*) - |\!|v-v_0|\!|_1 = \varrho^* - |\!|v-v_0|\!|_1.\]
Hence, there exists an $S\in\cS$ such that $\mathcal{I}(S,\varrho^*-|\!|v-v_0|\!|_1)\geqslant \varrho^*-|\!|v-v_0|\!|_1$ and by \eqref{eq:rewrittenSb} this entails $s\geqslant\varrho^*-|\!|v-v_0|\!|_1$. Likewise, we derive $\varrho^*\geqslant s-\delta$ and so $|\varrho^*-s|\leqslant |\!|v-v_0|\!|_1$.

We proceed by developing an upper bound on the $L_1$-difference between $v
$ and $v_0$:
\begin{align*}
  \int_0^1\big|v(x) - v_0(x)\big|{\rm d}x &= \int_0^1v_0(x)\big|\epsilon_I(x)\big|{\rm d}x \\
  &\leqslant \frac{s}{c(1-s)(1-\delta)} \int_0^1\big|\epsilon_I(x)\big|{\rm d}x \\
  &\leqslant \frac{s}{c(1-s)(1-\delta)} \left(\sum_{i\in I}\int_0^1\tau_i(x){\rm d}x + \beta\right) \\
  &\leqslant^{(*)} \frac{s}{c(1-s)(1-\delta)} \left(\frac{c^2}{2} + L c\right)\frac{1}{K}.
\end{align*}
Here $(*)$ is justified by Lemma \ref{lem:preliminaries}.($v$).1. In addition, since $\epsilon_I(x)\leqslant 0$ for $x\notin I^\dagger$ by Lemma \ref{lem:preliminaries}.($iii$) and $\text{vol}(S\backslash I^\dagger)\leqslant \text{vol}(I^\dagger\backslash S)$ by Lemma \ref{lem:preliminaries}.($iv$), we conclude that
\begin{align*}
  \int_{I^\dagger\backslash S}\left(1 + \epsilon_I(x) \right){\rm d}x -& \int_{S\backslash I^\dagger} \left(1 + \epsilon_I(x)\right){\rm d}x \\
  & \geqslant \int_{I^\dagger\backslash S}\left(1 + \epsilon_I(x)\right){\rm d}x- \text{vol}(S\backslash I^\dagger) \\
  & \geqslant \int_{I^\dagger\backslash S}\left(1 + \epsilon_I(x)\right){\rm d}x- \text{vol}(I^\dagger\backslash S)  = \int_{I^\dagger\backslash S}\epsilon_I(x){\rm d}x.
\end{align*}
Hence,
\begin{align*}
  r(S^*,v) - r(S,v) \geqslant& \frac{s(1-s)(1-\delta)}{c(1-s)(1-\delta)+cs}\int_{I^\dagger\backslash S}\epsilon_I(x){\rm d}x \\
  &- \frac{s^2(c+2 L)}{\big((1-s)(1-\delta)+s\big)(1-s)(1-\delta)}\frac{1}{K}.  
\end{align*}
Applying the latter inequality to $S = S_t$, for $t=1, \ldots, T$, and taking the expectation of the sum of these terms yields the desired result, since 
\begin{align*}
\E_I \left[\sum_{t=1}^T\int_{I^\dag\backslash {S}_t}\epsilon_I(x){\rm d}x \right] & =\int_{I^\dag}\E_I \left[\sum_{t=1}^T(1-1_{{S}_t}(x))\epsilon_I(x) {\rm d}x \right]\\
& =\int_{I^\dag} (T- \E_I[k(x)])\epsilon_I(x) {\rm d}x.
\end{align*}
\hfill$\Box$
\vspace{3mm}

\subsubsection*{Proof of Proposition \ref{prop:steptwo}.}
Let $x \in [0,1]$, $I \in \mathcal{D}_K$, $i \in I$, and $J = I \backslash \{i\}$. It suffices to show that there is a $C_c > 0$ such that 
\begin{equation}
\Big|\mathbb{E}_I[k(x)]-\mathbb{E}_J[k(x)]\Big| \leqslant T\sqrt{2\KL(\mathbb{P}_I|\!|\mathbb{P}_J)},
\label{eq:firstpart}
\end{equation}
and
\begin{equation}
\KL(\mathbb{P}_I|\!|\mathbb{P}_J) \leqslant \tfrac{1}{2}C_c^2\frac{T}{K^3}.
\label{eq:secondpart}
\end{equation}
We first prove \eqref{eq:firstpart}, using Pinsker's inequality, that states that for any probability measures $\mathbb{P}$ and $\mathbb{Q}$ defined on the same probability space $(\Omega,\F)$,
\[2\sup_{A\in\F}\Big(\mathbb{P}(A)-\mathbb{Q}(A)\Big)^2 \leqslant \KL(\mathbb{P}|\!|\mathbb{Q}),\]
or, equivalently,
\begin{equation}
\sup_{A\in\F}\Big|\mathbb{P}(A)-\mathbb{Q}(A)\Big| \leqslant \sqrt{\tfrac{1}{2}\KL(\mathbb{P}|\!|\mathbb{Q})}.
\label{eq:pinsk}
\end{equation}
Consider the probability measures $p$ and $q$ on $\{0,\ldots,T\}$, defined by 
\[p(n):=\mathbb{P}_I(k(x)=n)\qquad\text{and}\qquad q(n):=\mathbb{P}_J(k(x)=n), \quad (n \in \{0, \ldots, T\}).\]
From the equality
\begin{equation}
\sup_{n=0,\ldots,T} |p(n)-q(n)| = \frac{1}{2}\sum_{n=0}^T|p(n)-q(n)|.
\label{eq:totalvar}
\end{equation}
we obtain
\begin{align*}
\Big|\mathbb{E}_I[k(x)]-\mathbb{E}_J[k(x)]\Big| &= \left|\sum_{n=0}^T n(p(n)-q(n))\right| \\
&\leqslant \sum_{n=0}^T n \left| p(n)-q(n)\right| \leqslant T\sum_{n=0}^T \left| p(n)-q(n)\right| \\
&=^{(*)} 2T\sup_{n=0,\ldots,T} |p(n)-q(n)|  \leqslant^{(**)} T\sqrt{2\,\KL(\mathbb{P}_I|\!|\mathbb{P}_J)},
\end{align*}
where $(*)$ follows by \eqref{eq:totalvar}, and $(**)$ follows by \eqref{eq:pinsk}. 
This proves \eqref{eq:firstpart}.

We now prove \eqref{eq:secondpart}. Write $v(x) = v_I(x)$ and $u(x) = v_J(x)$, for $x \in [0,1]$. We denote the no-purchase probabilities at time $t$ as
\[p_t :=\frac{1}{1+\int_{S_t}v(x){\rm d}x} \qquad\text{and}\qquad q_t:=\frac{1}{1+\int_{S_t}u(x){\rm d}x}.\]
Note by Lemma \ref{lem:preliminaries}.($ii$).1 that $p_t,q_t\in\left[p_0,1\right]$, where
\[p_0:= \frac{(1-s)(1-\delta)}{(1-s)(1-\delta) + s(1+H)}.\]
The Kullback-Leibler (KL) divergence 
$\KL(\mathbb{P}_I|\!|\mathbb{P}_J)$ 
can be written as
\begin{align*}
\KL(\mathbb{P}_{\it I}|\!|\mathbb{P}_{\it J})&=\mathbb{E}_I\sum_{t=1}^T\left(p_t\log\frac{p_t}{q_t} +\int_{S_t}\log\left(\frac{p_tv(x)}{q_tu(x)}\right)p_tv(x){\rm d}x\right)\\
&=\mathbb{E}_I\sum_{t=1}^T\left(p_t\log\left(1+\frac{p_t-q_t}{q_t}\right)\right. + \left.\int_{S_t}\log\left(1+\frac{p_tv(x)-q_tu(x)}{q_tu(x)}\right)p_tv(x){\rm d}x\right).
\end{align*}
Since $\log(1+x)\leqslant x$ for all $x>-1$, we find the following upper bound:
\begin{align*}
\KL(\mathbb{P}_{\it I}|\!|\mathbb{P}_{\it J})&=\mathbb{E}_I\sum_{t=1}^T\left(p_t\log\left(1+\frac{p_t-q_t}{q_t}\right)\right. + \left.\int_{S_t}\log\left(1+\frac{p_tv(x)-q_tu(x)}{q_tu(x)}\right)p_tv(x){\rm d}x\right)\\
&\leqslant \mathbb{E}_I\sum_{t=1}^T\left(p_t\frac{p_t-q_t}{q_t}+\int_{S_t}\frac{p_tv(x)-q_tu(x)}{q_tu(x)}p_tv(x){\rm d}x\right) \\
&=\mathbb{E}_I\sum_{t=1}^T\left(\frac{(p_t-q_t)^2}{q_t}+\int_{S_t}\frac{\big(p_tv(x)-q_tu(x)\big)^2}{q_tu(x)}{\rm d}x\right)\\
&\hphantom{=}+\mathbb{E}_I\sum_{t=1}^T\left(p_t-q_t +\int_{S_t}\big(p_tv(x)-q_tu(x)\big){\rm d}x\right)\\
&= \mathbb{E}_I\sum_{t=1}^T\left(\frac{(p_t-q_t)^2}{q_t}+\int_{S_t}\frac{\big(p_tv(x)-q_tu(x)\big)^2}{q_tu(x)}{\rm d}x\right)\\
&\phantom{=}+ \mathbb{E}_I\sum_{t=1}^T\big(p_t-q_t + (1-p_t) - (1-q_t)\big)\\
&= \mathbb{E}_I\sum_{t=1}^T\left(\frac{(p_t-q_t)^2}{q_t}+\int_{S_t}\frac{\big(p_tv(x)-q_tu(x)\big)^2}{q_tu(x)}{\rm d}x\right).
\end{align*}
Note that $q_t\geqslant p_0$ and $u(x) \geqslant 1/C_1$ for all $x\in[0,1]$, where
\[C_1:=\frac{c(1-s)}{s(1-\beta)} > 0.\]
Hence, we can bound the KL divergence further as
\begin{align}
\KL(\mathbb{P}_{\it I}|\!|\mathbb{P}_{\it J})&\leqslant \mathbb{E}_I\sum_{t=1}^T\left(\frac{(p_t-q_t)^2}{q_t}+\int_{S_t}\frac{\big(p_tv(x)-q_t u(x)\big)^2}{q_tu(x)}{\rm d}x\right) \notag \\
&\leqslant \frac{1}{p_0}\mathbb{E}_I\sum_{t=1}^T\left(
\underbrace{
(p_t-q_t)^2}_{(a)} +C_1\underbrace{ \int_{S_t}\big(p_tv(x)-q_tu(x)\big)^2 {\rm d}x }_{(b)} \right). \label{eq:interbellum}
\end{align}
We bound both ($a$) and ($b$) in \eqref{eq:interbellum} from above. Let $t \in \{1, \ldots, T\}$. For ($a$), observe that 
\begin{align}
(p_t-q_t)^2 &=^{(*)} \frac{\left(\int_{S_t}(v(x)-u(x)){\rm d}x\right)^2}{\left(1+\int_{S_t}v(x){\rm d}x\right)^2\left(1+\int_{S_t}u(x){\rm d}x\right)^2}\notag\\
&\leqslant \left(\int_{S_t}(v(x)-u(x)){\rm d}x\right)^2\notag\\
&\leqslant \left(\frac{s}{c(1-s)(1-\delta)}\int_{S_t}\tau_{i	}(x){\rm d}x\right)^2\leqslant^{(**)} C_2\frac{c^2}{4K^4}\label{eq:ptqt},
\end{align}
where 
\[C_2 := \frac{s^2}{c^2(1-s)^2(1-\delta)^2}, \]
and where $(*)$ holds since the cross terms cancel out and $(**)$ follows by Lemma \ref{lem:preliminaries}.($v$).1.

We now bound ($b$) in \eqref{eq:interbellum} from above. Observe that 
\begin{align}
\int_{S_t}\big(p_tv(x)-q_tu(x)\big)^2{\rm d}x & = \int_{S_t}\big(p_tv(x)- q_tv(x) + q_tv(x)-q_tu(x)\big)^2{\rm d}x\notag\\
& = (p_t-q_t)^2\int_{S_t}v(x)^2{\rm d}x \label{eq:inte1} \\
&\phantom{=}+2q_t(p_t-q_t)\int_{S_t}v(x)\tau_i(x){\rm d}x \label{eq:inte2} \\
&\phantom{=} +q_t^2\int_{S_t}\left(\tau_i(x)\right)^2{\rm d}x. \label{eq:inte3}
\end{align}
The integral in (\ref{eq:inte1}) can be bounded by applying Lemma \ref{lem:preliminaries}.($v$).2. Combining that with the bound for $(p_t-q_t)^2$ from (\ref{eq:ptqt}), gives
\begin{align*}
(p_t-q_t)^2\int_{S_t}v(x)^2{\rm d}x &\leqslant C_2^2(1+H)\frac{c^3}{4K^4}.
\end{align*}
For the term (\ref{eq:inte2}), Lemma \ref{lem:preliminaries}.($i$) shows that $\tau_i(x)\leqslant Hc/K$. Together with (\ref{eq:ptqt}) and Lemma \ref{lem:preliminaries}.($ii$).1 we find
\begin{align*}
\ds 2q_t(p_t-q_t)\int_{S_t}v(x)\tau_i(x){\rm d}x & \leqslant 2|p_t-q_t|\int_{S_t}v(x)\tau_i(x){\rm d}x \\
& \leqslant 2|p_t-q_t|\left(\max_{x\in[0,1]}\tau_i(x)\right)\int_S v(x){\rm d}x \\
& \leqslant 2\sqrt{C_2}\frac{c}{2K^2} \left(H\frac{c}{K}\cdot c\sqrt{C_2}(1+H)\right) \\
& = C_2 H(1+H)\frac{c^3}{K^3}
\end{align*}
Finally, we bound the term (\ref{eq:inte3}). As a consequence of Lemma \ref{lem:preliminaries}.($v$).2, we have 
\begin{alignat*}{3}
q_t^2\int_{S_t}\left(\tau_i(y)\right)^2\d y &\leqslant &&\int_{S_t}\left(\tau_i(y)\right)^2\d y \leqslant &&\frac{c^3}{4\sigma\sqrt{\pi}K^3}.
\end{alignat*}
Inserting the derived upper bounds on \eqref{eq:inte1}, \eqref{eq:inte2}, \eqref{eq:inte3} in \eqref{eq:interbellum}, we obtain
\begin{align*}
\KL(\mathbb{P}_I|\!|\mathbb{P}_J) &\leqslant \frac{1}{p_0}\mathbb{E}_I\sum_{t=1}^T\left((p_t-q_t)^2+C_1\int_{S_t}\big(p_tv(y)-q_tu(y)\big)^2\d y\right) \\
&\leqslant \frac{1}{p_0}\mathbb{E}_I\sum_{t=1}^T\left(C_2\frac{c^2}{4K^4}+C_1\left(C_2^2(1+H)\frac{c^3}{4K^4} + C_2 H(1+H)\frac{c^3}{K^3} +\frac{c^3}{4\sigma\sqrt{\pi}K^3}\right)\right) \\
&\leqslant \frac{1}{4p_0}\left(c^2C_2+C_1\left(c^3C_2^2(1+H) + 4c^3C_2 H(1+H) +\frac{c^3}{\sigma\sqrt{\pi}}\right)\right)\frac{T}{K^3}.
\end{align*}
This implies \eqref{eq:secondpart}.
\hfill$\Box$
\vspace{3mm}

\subsubsection*{Proof of Theorem \ref{th:lowerboundc<1}.}
We first show that the preference functions $v_0$ and $\{ v_I : I \in \mathcal{D}_K, K \geqslant 2\}$ satisfy Assumption \ref{A1}. To see this observe that the choice $c\in(0,\tfrac{1}{4}]$, $s=0.8c$ and $\delta=\tfrac{1}{2}$ implies 
\[v_0(x)\in\left[\frac{0.8}{1-s},\frac{1.6}{1-s}\right]\subseteq\left[ 0.8,2\right].\]
Moreover, for all $K\geqslant 2$ and $I \in \mathcal{D}_K$ we have $\beta \leqslant L/8 \leqslant 0.0013$, and therefore 
\[v_I(x) \geqslant v_0(x) (1 - \beta) \geqslant 0.79,\]
and Lemma \ref{lem:preliminaries}.$(i)$ implies 
\[v_I(x) \leqslant v_0(x) \left(1 + \frac{H}{8}\right) \leqslant 2.01\leqslant 2.56 \leqslant \frac{w(0)}{\int_0^1 (w(x)-w(0)){\rm d}x} \leqslant 4,\]
for all choices of $c \in (0, \tfrac{1}{4}]$. This shows that Assumption \ref{A1}(i) is satisfied with $\vmax = 4$ and $\underline{v} = 0.79$. 

We now show that $v_I'(\cdot)$ is uniformly bounded and hence Assumption \ref{A1}(ii) is satisfied as well. To this end, observe that 
\begin{align*}
  |v'_I(x)| = |v_0'(x)|\left|1+\sum_{i\in I}\tau_i(x)\right| + |v_0(x)|\left|\sum_{i\in I}\tau_i'(x)\right|,
\end{align*}
for all $x \in [0,1]$. 
Therefore, by Lemma \ref{lem:preliminaries}.($i$) it suffices to show that $\sum_{i\in I}\tau_i'(x)$ is uniformly bounded. Note that
\[\tau_i'(x) = -\frac{2}{\sigma^2}\phi_i(x)b\big(\phi_i(x)\big).\]
For all $x \in [0,1]$, let $i_x:=\floor{Kx/c}$. Then $x\in B_{i_x}$ for all $x \in [0,1]$, where $B_{N_K+1}:=[0,1]\backslash \bigcup_{i\in[N_K]}B_i$, and $\phi_i(B_{i_x})= \big[2(i_x-i) - 1, 2(i_x-i) + 1\big)$. Since $|yb(y)|$ is decreasing for $y\geqslant 1$ and increasing for $y\leqslant-1$, we obtain that, for all $i<i_x$, 
\[0<\phi_i(x)b\big(\phi_i(x)\big)\leqslant \big(2(i_x-i) - 1\big) b\big(2(i_x-i) - 1\big),\]
and for all $i>i_x$,
\[0<-\phi_i(x)b\big(\phi_i(x)\big)\leqslant \big(2(i_x-i) + 1\big) b\big(2(i_x-i) + 1\big).\]
From this we conclude that
\begin{align*}
  \left|\sum_{i\in I}\tau_i'(x) \right|& = \left|\sum_{i\in I } \frac{2}{\sigma^2}\phi_i(x) b\big(\phi_i(x)\big)\right| \\
  & \leqslant \frac{2}{\sigma^2}\left(\left|\phi_{i_x}(x)b\big(\phi_{i_x}(x)\big)\right| + \sum_{i=1}^{i_x-1} \phi_i(x) b\big(\phi_i(x)\big) - \sum_{i=i_x+1}^{N_K} \phi_i(x) b\big(\phi_i(x)\big) \right) \\
  & \leqslant \frac{2}{\sigma^2}\left(\left|\phi_{i_x}(x)b\big(\phi_{i_x}(x)\big)\right| + \sum_{i=1}^{i_x-1} \big(2(i_x-i) - 1\big) b\big(2(i_x-i) - 1\big) \right.\\
  &\qquad \left.- \sum_{i=i_x+1}^{N_K} \big(2(i_x-i) + 1\big) b\big(2(i_x-i) + 1\big) \right) \\
  & \leqslant \frac{2}{\sigma^2}\left(\frac{\sigma}{\sqrt{e}} + 2\sum_{n=1}^{\infty} (2n-1) b\big(2n-1\big) \right) <\infty.
\end{align*}
As a result, $v_0$ and $\{ v_I : I \in \mathcal{D}_K, K \geqslant 2\}$ satisfy Assumption \ref{A1}. This implies 
\begin{align}
\Delta_\pi(T) &= \sup_{v \in \cV} \Delta_{\pi}(T,v) \notag \\
&\geqslant \frac{1}{|\mathcal{D}_K|}\sum_{I\in \mathcal{D}_K} \Delta_\pi(T,v_I) \notag\\
&\geqslant \frac{1}{|\mathcal{D}_K|}\sum_{I\in \mathcal{D}_K} \left( C_1 \int_{I^\dag} (T-\mathbb{E}_I [k(x)])\epsilon_I(x){\rm d}x - C_2\frac{T}{K} \right), \label{eq:710split}
\end{align}
where $C_1$ and $C_2$ are as in Proposition \ref{prop:lower-bound-step-1}.

The integral $\int_{I^\dag}\epsilon_I(x) {\rm d}x$ can be bounded from below as
\begin{align*}
\int_{I^\dag}\epsilon_I(x){\rm d}x & =\sum_{i\in I}\int_{I^\dagger}\tau_i(x){\rm d}x - \beta c = \sum_{i\in I}\int_{B_i}\tau_i(x){\rm d}x - \beta c \\
&\geqslant^{(*)} P\frac{c^2}{2K} - L \frac{c^2}{K} = \frac{c^2(P - 2L)}{2K},
\end{align*}
where at $(*)$ we used Lemma \ref{lem:preliminaries}.($v$).1. We use this lower bound to analyze \eqref{eq:710split}. To this end, let $C_3 := c^2C_1(P-2L)/2$. Then
\begin{equation}
\Delta_\pi(T) \geqslant (C_3 - C_2)\frac{T}{K} - \frac{C_1}{|\mathcal{D}_K|} \underbrace{ \sum_{I\in \mathcal{D}_K}\int_{I^\dag} \mathbb{E}_I [k(x)]\epsilon_I(x){\rm d}x }_{(a)}.
\label{eq:inbetween}
\end{equation}
We now bound the term ($a$) in \eqref{eq:inbetween} from above, using Proposition \ref{prop:steptwo}. 
Let $C_c$ denote the constant from Proposition \ref{prop:steptwo}, and let $I\in \mathcal{D}_K$ and $J = I\backslash\{i\}$ for some $i\in I$. Then, for $x \in B_i$,
\begin{equation}
\mathbb{E}_I [k(x)]\epsilon_I(x) \leqslant \left(\mathbb{E}_{J}[k(x)] + C_c\left(\frac{T}{K}\right)^{3/2}\right)|\epsilon_{I}(x)|.
\label{eq:twoface}
\end{equation}
To apply \eqref{eq:twoface} in order to bound ($a$) in \eqref{eq:inbetween}, we change the order of summation and integration and rewrite the summation itself. Let $U = \bigcup_{i=1}^{N_K} B_i$ denote the union of all bins, and for all $x\in U$, let $i_x=\floor{Kx/c}$ again denote the index of the bin $B_{i_x}$ such that $x\in B_{i_x}$, for all $x \in [0,1]$. Note that for each $x\in U$ that the mapping $I\mapsto I\backslash\{i_x\}$ between 
\[E_K^x :=\{I\in \mathcal{D}_K:x\in I^\dag\}\qquad\text{and}\qquad F_{K-1}^x :=\{J\in \mathcal{D}_{K-1}:x\notin J^\dag\}\]
is a bijection. Hence,
\begin{align}
\sum_{I\in \mathcal{D}_K}\int_{I^\dag} \mathbb{E}_I [k(x)]\epsilon_I(x){\rm d}x &= \int_{x\in U}\sum_{I\in E^x_K} \mathbb{E}_I [k(x)]\epsilon_I(x){\rm d}x \notag\\
&= \int_{x\in U}\sum_{J\in F_{K-1}^x} \mathbb{E}_{J\cup \{i_x\}}[k(x)]\epsilon_{J\cup \{i_x\}}(x){{\rm d}x} \notag\\
&\leqslant^{(*)} \int_{x\in U}\sum_{J\in F_{K-1}^x} \mathbb{E}_{J}[k(x)]\big|\epsilon_{J\cup\{i_x\}}(x)\big|{{\rm d}x} \label{eq:head1}\\
&\phantom{=}+ C_c\left(\frac{T}{K}\right)^{3/2}\int_{x\in U}\sum_{J\in F_{K-1}^x} \big|\epsilon_{J\cup\{i_x\}}(x)\big|{{\rm d}x}, \label{eq:head2}
\end{align}
where at $(*)$ we apply \eqref{eq:twoface}. We now bound \eqref{eq:head1} and \eqref{eq:head2} from above. For \eqref{eq:head1}, $|\epsilon_{I}(x)|$ is bounded uniformly in $x$ by Lemma \ref{lem:preliminaries}.($i$):
\begin{align*}
& \int_{x\in U}\sum_{J\in F_{K-1}^x}\mathbb{E}_J [k(x)]\big|\epsilon_{J\cup\{i_x\}}(x)\big|{\rm d}x \\
\leqslant & \left(H+L\right)\frac{c}{K}\int_{x\in U}\sum_{J\in F_{K-1}^x}\mathbb{E}_J [k(x)]{\rm d}x  \\
=& \left(H+L\right)\frac{c}{K}\sum_{J\in \mathcal{D}_{K-1}}\int_{x\in U\backslash{J^\dagger}}\mathbb{E}_J[k(x)]{\rm d}x \\
\leqslant& \left(H+L\right)\frac{c}{K}\sum_{J\in \mathcal{D}_{K-1}}\int_0^1\mathbb{E}_J[k(x)]{\rm d}x \\
\leqslant& \left(H+L\right)\frac{c}{K}\sum_{J\in \mathcal{D}_{K-1}}\sum_{t=1}^T\mathbb{E}_J[\text{vol}(S_t)]  \\ 
\leqslant& \left(H+L\right)\frac{c^2}{K}|\mathcal{D}_{K-1}|T.
\end{align*}
We now consider \eqref{eq:head2}. Observe that $|\epsilon_{I}(x)|$ is bounded locally on $B_i$:
\begin{align*}
\lefteqn{\int_{x\in U}\sum_{J\in F^x_{K-1}} \big|\epsilon_{J\cup\{i_x\}}(x)\big|{{\rm d}x}= \sum_{J\in \mathcal{D}_{K-1}}\int_{x\in U\backslash {J^\dagger}}\big|\epsilon_{J\cup\{i_x\}}(x)\big|{\rm d}x}\\
&= \sum_{J\in \mathcal{D}_{K-1}}\sum_{i\notin J}\int_{B_i}|\epsilon_{J\cup \{i\}}(x)|{\rm d}x 
\leqslant^{(*)} \sum_{J\in \mathcal{D}_{K-1}}\sum_{i\notin J}\int_{B_i}\big(\tau_i(x) + \beta\big){\rm d}x \\
&\leqslant^{(**)} \sum_{J\in \mathcal{D}_{K-1}}\sum_{i\notin J} \frac{c^2(1+2L)}{2K^2} = \frac{c^2(1+2L)}{2K^2}|\mathcal{D}_{K-1}|(N_K-K+1),
\end{align*}
where we apply Lemma \ref{lem:preliminaries}.($vi$) at $(*)$ and Lemma \ref{lem:preliminaries}.($v$).1 at $(**)$. After inserting these upper bounds for \eqref{eq:head1} and \eqref{eq:head2} into \eqref{eq:inbetween}, we conclude 
\begin{align*}
\Delta_\pi(T)&\geqslant (C_3-C_2)\frac{T}{K} - \frac{C_1}{|\mathcal{D}_K|}\left(\left(H+L\right)\frac{c^2}{K}|\mathcal{D}_{K-1}|T + \frac{c^2C_c(1+2L)}{2}|\mathcal{D}_{K-1}|(N-K+1) \frac{T^{3/2}}{K^{7/2}}\right) \\
&=\left(C_3-C_2 - \frac{c^2C_1\left(H+L\right)|\mathcal{D}_{K-1}|}{|\mathcal{D}_K|}\right)\frac{T}{K} - \frac{c^2C_1C_c(1+2L)|\mathcal{D}_{K-1}|}{2|\mathcal{D}_K|}(N-K+1) \frac{T^{3/2}}{K^{7/2}}.
\end{align*}
Next, note that
\[\frac{|\mathcal{D}_{K-1}|}{|\mathcal{D}_K|} = \frac{K}{N_K-K+1},\]
and therefore
\begin{equation}
\Delta_\pi(T) \geqslant
\underbrace{ \left(C_3-C_2-\frac{\left(H+L\right)c^2C_1K}{N_K-K+1}\right) }_{(b)} \frac{T}{K} - \frac{c^2C_1C_c(1+2L)}{2} \frac{T^{3/2}}{K^{5/2}}.\label{eq:monstrosity}
\end{equation}
We abbreviate the constant $C_4:= c^2C_1C_c(1+2L)/2$. The factor ($b$) in front of the $T/K$ term above can be bounded further from below. To this end, note that
\[N_K-K+1 \geqslant \left(\frac{1}{c}-1\right)K,\]
and therefore \eqref{eq:monstrosity} implies 
\[\Delta_\pi(T)\geqslant \left(C_3-C_2 -\frac{\left(H+L\right)c^3C_1}{1-c}\right)\frac{T}{K} - C_4 \frac{T^{3/2}}{K^{5/2}}.\]
Let
\[ C_6 := \frac{P-2L}{2} - \frac{(H-L)c}{1-c}, \]
and 
\[C_5:=C_3-C_2-\frac{(H+L)c^3C_1}{1-c} = c^2C_1 C_6 - C_2.\]
By computation and the assumption $c \in (0,\tfrac{1}{4}]$ we obtain $C_6\geqslant (P-2L)/2 - (H-L)/3 = 0.043 > 0$. In addition, our choice of $s = 0.8 c$ implies 
\[\frac{s}{1-s} < \frac{\sqrt{c C_6}(1-\delta)}{\sqrt{c+L}},\]
and therefore $C_5 = c^2 C_1 C_6 - C_2>0$. Now, choose
\[\gamma= \left(\frac{5C_4}{C_5}\right)^{2/3}\qquad\text{and}\qquad K=\max\{2,\ceil{\gamma T^{1/3}}\}.\]
For $T> 1/\gamma^3$, we know that $K = \ceil{\gamma T^{1/3}}$ as well as $K<\gamma T^{1/3}+1< 2\gamma T^{1/3}$ and $K\geqslant \gamma T^{1/3}$. Therefore, for $T> 1/\gamma^3$
\begin{align*}
  \Delta_\pi(T) & \geqslant \frac{C_5}{2\gamma} \: T^{1/3} - \frac{C_4}{\gamma^{5/2}} \:T^{1/3}\\
  &=\left(\tfrac{1}{2}\left(\tfrac{1}{5}\right)^{2/3} - \left(\tfrac{1}{5}\right)^{5/3}\right)\frac{C_5^{5/3}}{C_4^{2/3}}\:T^{2/3}.
\end{align*}
For $T$ such that $1\leqslant T\leqslant 1/\gamma^3$, we know that $K=2$ as well as $\sqrt{T}\leqslant C_5/5C_4$ and thus
\begin{align*}
  \Delta_\pi(T) & \geqslant \frac{C_5}{2} \: T - \frac{\sqrt{2}C_4}{8} \:T^{3/2}\\
  &= \left(\frac{C_5}{2} - \frac{\sqrt{2}C_4}{8} \:\sqrt{T} \right) T \\
  &\geqslant \left(\tfrac{1}{2} - \tfrac{\sqrt{2}}{40} \right) C_5 \:T\geqslant \left(\tfrac{1}{2} - \tfrac{\sqrt{2}}{40} \right) C_5 \:T^{2/3}.
\end{align*}
Therefore, we have shown the desired result for
\[\Cmin = \min\left\{\left(\tfrac{1}{2} - \tfrac{\sqrt{2}}{40} \right) C_5, \left(\tfrac{1}{2}\left(\tfrac{1}{5}\right)^{2/3} - \left(\tfrac{1}{5}\right)^{5/3}\right) \frac{C_5^{5/3}}{C_4^{2/3}}\right\}>0.\]
\vspace*{-1cm}
\hfill$\Box$
\vspace{3mm}

\subsubsection*{Proof of Lemma \ref{lem:preliminaries}.}
For $x\in[0,1]$, let $i_0\in [N_K]$ $y=2Kx/c - 2i_0+1$. Then we find that ($i$) holds due to
\begin{align*}
\sum_{i\in I} \tau_i(x) &= \frac{c}{K}\frac{1}{\sigma\sqrt{2\pi}}\sum_{i\in I}\exp\big(-\tfrac{1}{2\sigma^2}(y + 2i_0 - 2i)^2\big) 
\leqslant \frac{c}{K}\frac{1}{\sigma\sqrt{2\pi}}\sum_{n\in \Z}\exp\big(-\tfrac{1}{2\sigma^2}(y - 2n)^2\big) \\ 
&\leqslant \frac{c}{K}\frac{1}{\sigma\sqrt{2\pi}}\sum_{n\in \Z}\exp\left(-\frac{2n^2}{\sigma^2}\right) = H \frac{c}{K}.
\end{align*}
Observe that ($ii$) is a corollary of ($i$), since $v_0(x)\leqslant \frac{s}{c(1-s)(1-\delta)}$ for all $x \in [0,1]$, and therefore 
\begin{equation*}
v_I(x)\leqslant \frac{s}{c(1-s)(1-\delta)}\left(1+\sum_{i\in I}\tau_i(x)\right).
\end{equation*}
For ($iii$), let $x\notin I^\dag$ and $i_x:=\floor{Kx/c}$ such that $x\in B_{i_x}$, where $B_{N_K+1}:=[0,1]\backslash \bigcup_{i\in[N_K]}B_i$. Note that $\tau_i$ is either increasing or decreasing on $B_{i_x}$ for $i\neq i_x$. Then 
\[\tau_i(x)\leqslant \max\left\{\tau_i\left(c\frac{i_x-1}{K}\right),\tau_i\left(c\frac{i_x}{K}\right)\right\}=\frac{c}{K}\max\Big\{b\big(2(i_x-i)+1\big), b\big(2(i_x-i)-1\big)\Big\},\]
for $i\neq i_x$. From this, we derive for any $x\notin I^\dag$,
\begin{align*}
\sum_{i\in I}\tau_i(x) & \leqslant \frac{c}{K}\frac{1}{\sigma\sqrt{2\pi}} \sum_{i\in I}\max\Big\{\exp\big(-\tfrac{1}{2\sigma^2}(2(i_x-i) + 1)^2\big),\exp\big(-\tfrac{1}{2\sigma^2}(2(i_x-i) - 1)^2\big)\Big\}\\
& \ds \leqslant \frac{c}{K}\frac{1}{\sigma\sqrt{2\pi}} \sum_{n\in \Z}\exp\big(-\frac{1}{2\sigma^2}(2n - 1)^2\big),
\end{align*}
which implies ($iii$). For ($iv$), we observe that by $\text{vol}(I^\dag)=c$, 
\[c = \text{vol}(I^\dag) = \text{vol}(I^\dag\cap S_t) + \text{vol}(I^\dag\backslash S_t)= \text{vol}(S_t) - \text{vol}(S_t\backslash I^\dag) + \text{vol}(I^\dag\backslash S_t),\]
Since $\text{vol}(S_t)\leqslant c$, ($iv$) follows. Item ($v$) is derived by straightforward computation: for both results ($v$).1 and ($v$).2 we apply the variable substitution $y = 2Kx/c-2i+1$ to obtain 
\begin{align*}
\int_0^1\tau_i(x){\rm d}x &= \frac{c}{K} \int_0^1 b\left(\frac{2Kx}{c} - 2i + 1\right) {\rm d}x\\
&\leqslant \frac{c}{K}\int_{\R}b\left(\frac{2Kx}{c} - 2i + 1\right) {\rm d}x
=\frac{c^2}{2K^2} \int_{\R} b(y) \d y =\frac{c^2}{2K^2}.
\end{align*}
For the equality in $(v)$.1, we find that
\begin{align*}
\int_{B_i}\tau_i(x){\rm d}x &= \frac{c}{K} \int_{B_i} b\left(\frac{2Kx}{c} - 2i + 1\right) {\rm d}x\\
&=\frac{c^2}{2K^2} \int_{[-1,1]} b(y) \d y
 =\frac{c^2}{2K^2}P.
\end{align*}
For the integral in ($v$).2, we derive
\begin{align*}
\int_{[0,1]}(\tau_i(x))^2{\rm d}x &= \frac{c^2}{K^2} \int_{B_i} \left( b\left(\frac{2Kx}{c} - 2i + 1\right)\right)^2 {\rm d}x\\
&\leqslant \frac{c^2}{K^2} \int_{\R} \left( b\left(\frac{2Kx}{c} - 2i + 1\right)\right)^2{\rm d}x \\
&= \frac{c^3}{2K^3} \int_{\R} \left( b\left(y\right)\right)^2{\rm d}x = \frac{c^3}{4\sigma \sqrt{\pi}K^3}.
\end{align*}
Finally, for ($vi$) we point out that as a corollary of ($iii$), for $i\in I$, $\beta' \geqslant \beta$, and $x\in B_i$,
\[-\beta'\leqslant \epsilon_{I\backslash\{i\}}(x;\beta')\leqslant 0,\]
since $x\notin (I\backslash\{i\})^\dag$. Hence,
\begin{equation}\nonumber|\epsilon_I(x;\beta')| = \left|\tau_i(x) +\epsilon_{I\backslash\{i\}}(x;\beta')\right| \leqslant \tau_i(x) +\left|\epsilon_{I\backslash\{i\}}(x;\beta')\right|\leqslant \tau_i(x) + \beta'.\end{equation}
\hspace*{\fill}$\Box$

\vspace{3mm}
\subsubsection*{Proof of Lemma \ref{lem:anicecoincidence}.}
For any $\varrho\in[0,1-\delta]$ and any $x\in[0,1]$ it holds that $w(x)\geqslant 1-\delta\geqslant \varrho$ and therefore 
$\text{vol}(W_{\varrho}) = \text{vol}(\{x\in[0,1]:w(x)\geqslant \varrho\}) = 1$. In particular this implies that $\text{vol}(S_{\varrho}) = c$, for all $\varrho \in [0, 1 - \delta]$, where $S_{\varrho}$ is a maximizer of \eqref{eq:innermaximization}. Now, let $\varrho = s$. Since $s \in [0, 1 - \delta]$, it follows that
\[\mathcal{I}(S_\varrho,\varrho) = \int_{S_\varrho} v_0(x) \big(w(x)-\varrho\big){\rm d}x = \frac{s}{c}\int_{S_\varrho} \frac{w(x) - \varrho}{w(x)-s}{\rm d}x = s \frac{\text{vol}(S_\varrho)}{c} = \varrho,\]
and therefore $\varrho^* = s$ by Proposition \ref{prop:fullinformationcapacitated}.
\hfill$\Box$
\vspace{3mm}

\section*{Appendix C: Relation to discrete multinomial logit choice probabilities} 
\label{app:limitmodel}

The choice probabilities in our continuous assortment optimization model are closely connected to the discrete multinomial logit (MNL) model, in two regards.

First, our choice probabilities naturally arise as a limit of discrete models where the number of products grow large. To see this, consider a sequence of discrete MNL assortment optimization problems indexed by $n \in \N$, where the $n$-th problem corresponds to a setting with $n$ products labeled $i=1,\ldots, n$, each with associated location $i / (n+1)$ and valuation $v_i^{(n)} = v(i / (n+1)) / (n+1)$, for all $i=1,\ldots, n$ and some continuous function $v: [0,1] \ra (0,\infty)$.
Under the discrete MNL model, the probability that a customer selects a product in a (measurable) set $A \in [0,1]$ when being offered assortment $S$ is equal to 
\[ \frac{ \sum_{i: \frac{i}{n+1} \in A} v^{(n)}_i}{1 + \sum_{i: \frac{i}{n+1} \in S} v^{(n)}_i}. \]
It follows from classical results in integration theory \citep[see, e.g.,][]{Stroock1994} that this expression converges to \eqref{eq:prob-X-in-A} as $n \ra \infty$. 

Second, when the product space is discretized into finitely many products, each corresponding to a subinterval in $[0,1]$, then our model translates into choice probabilities that are described by a discrete MNL model. To see this, suppose that $I_1, \ldots, I_n$ are mutually disjoint subsets of $[0,1]$, each corresponding to a `discrete product', such that $\bigcup_{i=1}^n I_i = [0,1]$. Let $v_i := \int_{I_i} v(x) {\rm d}x$, for all $i$. Then, for each `discrete assortment' $\tilde{S} \subseteq \{1, \ldots, n\}$ and for each $i \in \tilde{S}$, the probability $P(i \mid \tilde{S})$ that a customer selects from $I_i$ when being offered assortment $\bigcup_{j \in \tilde{S}} I_j$, is equal to
\[ P(i \mid \tilde{S}) = \P\left(X^S \in I_i\right) = 
\frac{ \int_{I_i} v(x) {\rm d}x}{1 + \int_{\bigcup_{j \in \tilde{S}} I_j} v(x) {\rm d}x} = \frac{v_i}{1 + \sum_{j \in \tilde{S}} v_j}.\]
This is precisely the structure of a discrete MNL choice model.


\section*{Appendix D: Bisection algorithm for Section \ref{sec:capacitated}}
\label{app:bisection}

According to Proposition \ref{prop:fullinformationcapacitated}, the optimal assortment can be computed up to any desired accuracy $\epsilon > 0$. The algorithm $\text{COA}(n)$ below shows how this is done, where $n:=-\log\epsilon$. Recall that
\begin{equation}
  \mathcal{I}(S,\varrho) := \int_S v(x)\big(w(x)-\varrho\big){\rm d}x.
  \label{eq:defineIforappendix}
\end{equation}
The algorithm $\text{COA}(n)$ uses bisection to find the fixed-point solution $\varrho^*$ to the equation
\[ \mathcal{I}(S_\varrho,\varrho) = \varrho. \]
The value of $\mathcal{I}(S_\varrho,\varrho)$ is computed by relying on the level $\ell_\varrho$. This level value is calculated by an additional inner bisection using the algorithm $\text{IB}(n,\varrho)$. This algorithm is also presented below.

\begin{remark}
As mentioned, the calculation of the level $\ell_\varrho$ for a single $\varrho$ requires a bisection on its own. This means that the run time of $\text{IB}(n,\varrho)$ is $\mathcal{O}(-\log \epsilon)$, and hence the run time of $\text{COA}(n)$ is $\mathcal{O}((\log \epsilon)^2)$.
\end{remark}

\noindent\HRule
\begin{center}
{\tt Capacitated Optimal Assortment $\text{COA}(n)$}
\end{center}
{\tt 1. Initialization.} {\tt Let $n\geqslant 1$.} {\tt Put $a:=0$, $b:=1$, $\text{piv}:=(b-a)/2$ and $i:=1$. Go to 2.} \\
{\tt 2. Capacity check.} {\tt Put }
\[W_\texttt{piv} := \{x\in[0,1]:w(x)\geqslant \texttt{piv}\}.\]
\begin{itemize}
  \item[\tt (i)] {\tt If $\textrm{vol}(W_\text{piv})> c$, then go to 3.}
  \item[\tt (ii)] {\tt If $\textrm{vol}(W_\text{piv})\leqslant c$, then put $S_\text{piv}:=W_\text{piv}$ and $I_\text{piv}:=\mathcal{I}(S_\text{piv},\text{piv})$ as in \eqref{eq:defineIforappendix} and go to 5.}
\end{itemize}
{\tt 3. Inner bisection.} {\tt Compute $\ell_\text{piv}$ according to $\text{IB}(n,\text{piv})$. Go to 4.}\\
{\tt 4. Level set.} {\tt Put }
\[L_\texttt{piv}^+ := \{x\in[0,1]:v(x)(w(x)-\texttt{piv})>\ell_\texttt{piv}\},\]
\[L_\texttt{piv}^= := \{x\in[0,1]:v(x)(w(x)-\texttt{piv})=\ell_\texttt{piv}\}\]
{\tt and}
\[x_\texttt{piv} := \min\{x\in[0,1]:\textrm{vol}(L_\texttt{piv}^+) + \textrm{vol}\big(L_\texttt{piv}^=\cap[0,x]\big)=c\}\]
{\tt Put $S_\text{piv} = L_\text{piv}^+\cup\big(L_\text{piv}^=\cup[0,x_\text{piv}]\big)$ and $I_\text{piv}:=\mathcal{I}(S_\text{piv},\text{piv})$ as in \eqref{eq:defineIforappendix}. Go to 5.}\\
{\tt 5. Pivot.} 
\begin{itemize}
  \item[\tt (i)] {\tt If $I_\text{piv}> \text{piv}$, then put $a := \text{piv}$.} 
  \item[\tt (ii)] {\tt If $I_\text{piv}\leqslant \text{piv}$, then put $b := \text{piv}$.} 
\end{itemize}
{\tt Put $i:=i+1$. If $i\leqslant n$, then put $\text{piv}:=(b-a)/2$ and go to 2, else go to 6.}\\
{\tt 6. Optimization.} {\tt Put $S^* := S_\text{piv}$. Go to 7.}\\
{\tt 7. Terminate.}

\noindent\HRule

Recall that there is a possible degree of freedom for picking $S_\varrho$ if $\text{vol}(W_\varrho)>c$. By the definition of $x_\texttt{piv}$ above, we explicitly choose the left-most version. The algorithm $\text{IB}(n,\varrho)$ computes the level $\ell_\varrho$ for given $\varrho$. Recall by Lemma \ref{lem:perrho} that this level is defined as
\[\ell_\varrho := \max\{\ell \geqslant 0:\text{vol}\big(L(\varrho,\ell)\big)\geqslant c\}.\]
$\text{IB}(n,\varrho)$ also uses the bisection method, which is facilitated by the fact that, as a function of $\ell\geqslant 0$, $\text{vol}\big(L(\varrho,\ell)\big)$ is left-continuous and non-increasing by Lemma \ref{lem:mfunction}.

\noindent\HRule
\begin{center}
{\tt Inner Bisection $\text{IB}(n,\varrho)$}
\end{center}{\tt 1. Initialization.} {\tt Let $n\geqslant 1$ and $\varrho\in[0,1]$.} {\tt Put $a:=0$, $b:=v_{\max}(w_{\max}-\varrho)+1$, $\text{piv}:=(b-a)/2$ and $i:=1$. Go to 2.} \\
{\tt 2. Level set.} {\tt Put }
\[L^\texttt{piv} := \{x\in[0,1]:v(x)(w(x)-\varrho)\geqslant \texttt{piv}\}.\]
{\tt Go to 3.}\\
{\tt 3. Pivot.} 
\begin{itemize}
  \item[\tt (i)] {\tt If $\textrm{vol}(L^\text{piv})> c$, then put $a := \text{piv}$.} 
  \item[\tt (ii)] {\tt If $\textrm{vol}(L^\text{piv})\leqslant c$, then put $b := \text{piv}$.} 
\end{itemize}
{\tt Put $i:=i+1$. If $i\leqslant n$, then put $\text{piv}:=(b-a)/2$ and go to 2, else go to 4.}\\
{\tt 4. Optimization.} {\tt Put $\ell_\varrho := \text{piv}$. Go to 5.}\\
{\tt 5. Terminate.}\\
\noindent\HRule

\section*{Appendix E: Additional numerical experiments}
In this section we report the results of additional numerical experiments in which we compare the predictive performance of the continuous logit model with that of the discrete multinomial logit (MNL) model. Section E.1  describes the experimental set-up, and in Section E.2 we report our results.  Section E.3 contains additional details on the derivation of the maximum-likelihood estimator, and in Section E.4 we specify the kernel density estimator used in these numerical experiments.

\subsection*{E.1. Experimental set-up}

The goal of these additional numerical experiments is to compare the predictive performance of the continuous and the discrete logit choice model. To make such a comparison, we need to define an estimator of the model parameters, for both the continuous and the discrete choice model. For the discrete choice model we use the well-known maximum-likelihood estimator (MLE) to estimate the model parameters. To estimate the preference function of the continuous model, we develop a kernel density estimator (KDE).
Throughout this section we use the same  notations and concepts as in Sections \ref{sec:capacitatedpolicy} and \ref{sec:capacitatedupperbound}. 

We compare the predictive performance of the two models in different scenarios. For each scenario we randomly generate transaction data according to a true `ground truth model', which is either the discrete or the continuous model. Based on this data we estimate the preference values $v_1, \ldots, v_N$ of the discrete model and the preference function $v$ of the continuous model, using the MLE and KDE, respectively. We then evaluate the predictive performance of both models using three performance measures: (1) the relative revenue loss of the estimated optimal assortment compared to the true optimal revenue, (2) the $L_1$-difference between the estimated and true model parameters, and (3), following \cite{Berbeglia2018}, the absolute error of the estimated no-purchase probability. 


In what follows, we describe in detail the  different scenarios, the MLE and KDE, and the three performance measures  that we consider.

{\bf Scenarios. } We consider three different scenarios. In the first scenario the discrete model is the ground truth, with parameters $v^{(1)}_1,\ldots,v^{(1)}_N$ drawn uniformly at random from $[\frac{1}{10N},\frac{1}{2N}]$, for $N \in \{10, 30, 50\}$. This grossly violates our assumption imposed in the continuous model that the preference values are Lipschitz continuous. In the second scenario the discrete model is again the ground truth; however, the preference values 
$v^{(2)}_1,\ldots,v^{(2)}_N$ 
are set to $v^{(2)}_i := f(i/(N+1))/N$, for $i=1,\ldots, N$, where $N \in \{10, 30, 50\}$, 
\[f(x) = \frac{1}{10} + \phi(x;\mu,\sigma),\qquad x\in[0,1],\]
and where $\phi(\,\cdot\,;\mu,\sigma)$ is the normal probability density function with $\mu$ drawn uniformly at random from $[0,1]$, and $\sigma$ drawn uniformly at random from $[0.1,0.2]$.
Thus, in this second scenario, the continuous model might provide a relatively accurate description of the choice probabilities, despite being a misspecified model. Finally, in the third and last scenario we assume that the continuous model is the ground truth, and we test up to what extent the discrete model is able to produce accurate predictions of consumer's choice behavior. The preference function is set to \[ v^{(3)}(x) = \frac{1}{10} + \frac{1}{5} (2+x)(1-x) + \frac{2}{7}\phi(x;0.33,0.1) + \frac{1}{5}\phi(x;0.8,0.1),\qquad x \in [0,1].\]
The discrete model is estimated for $N \in \{10, 30, 50\}$ products. In all scenarios we set $w(x) := x$ for all $x \in [0,1]$.
For each scenario, for each $c \in \{ \frac{1}{2}, 1 \}$, and for each $N \in \{10,30,50\}$, we randomly generate 1\,000 transaction data sets of size 
$T\in\{50,100,200,500,1\,000,2\,000,5\,000\}$. In these transaction data sets, the assortments are set to the unit interval for $c = 1$. For $c = 1/2$ we let the assortments be $[0, 0.5]$ in the first $T/2$ time periods, and $[0.5, 1]$ in the second $T/2$ time periods. 
In the third scenario, in which the continuous model is the ground truth, the observed purchases for the discrete model are of the form  $Y_t=\sum_{i=1}^N i\mathbf{1}_{B_i}(X_t)$. 

We refer to a specific vector of preference parameters as an \emph{instance of the discrete model}, and to a specific preference function as an \emph{instance of the continuous model}. 
Each instance ${\boldsymbol v}=(v_1,\ldots,v_N)$ of the discrete model corresponds to an instance of the continuous model, by letting the discrete purchase $Y_t$ coincide with the continuous purchase $X_t\in B_{i_t}$ (and $X_t=\emptyset$ if $Y_t=0$) and by setting the preference function $v(x)$ equal to
\[v(x) := N\sum_{i=1}^N v_i \mathbf{1}_{B_i}(x).\]
Conversely, each instance of the continuous model with preference function $v$ that is constants on bins $B_1, \ldots, B_N$ corresponds to an instance of the discrete model by setting $v_i = \int_{B_i} v(x)\d x$, for all $i=1,\ldots, N$. 
Concretely, we let 
$v^{(1)}(\cdot)$ and $v^{(2)}(\cdot)$ be the preference functions of the continuous model that correspond to the (discrete) instance in scenario 1 and 2, and we let $(v^{(3)}_1,\ldots, v^{(3)}_N)$ be the vector of preference values that correspond to the (continuous) instance in scenario 3.


{\bf Estimators.}
For $j=1,2,3$, let $\hat{v}^{(j),{\rm KDE}}(x)$ denote the kernel density estimator of $v^{(j)}(x)$ (defined in more detail in Appendix E.4) and let $\hat{v}^{(j),{\rm MLE}}(x)$ denote the stepwise constant function
\[\hat{v}^{(j),{\rm MLE}}(x) := \sum_{i=1}^N \hat{v}^{(j),{\rm MLE}}_i \mathbf{1}_{B_i}(x),\]
where $\hat{v}^{(j),{\rm MLE}}_i$ denotes the MLE of $v^{(j)}_i$ for $i\in[N]$. That is,
\[\hat{v}^{(j),{\rm MLE}}_i :=\frac{\sum_{t=1}^T\mathbf{1}\{Y_t = i\}}{\sum_{t=1}^T\mathbf{1}\{Y_t = 0\}},\]
for $c=1$ and 
\[\hat{v}^{(j),{\rm MLE}}_i: = \frac{\sum_{t=(k-1)T/2+1}^{kT/2} \mathbf{1}\{Y_t=i\}}{\sum_{t=(k-1)T/2+1}^{kT/2} \mathbf{1}\{Y_t=0\}},\qquad i\in \{(k-1)N/2+1,\ldots,kN/2\},\:k=1,2,\]
for $c=0.5$, where $Y_t$ are simulated from scenario $j$. We set the assumed upper bound of $v(x)$ in all scenarios to $\vmax = 5$. For $c=1$, we let $\hat{v}^{(j),{\rm MLE}}_i$ be the fixed constant $\vmax/N$ if $\sum_{t=1}^T\mathbf{1}\{Y_t = 0\} = 0$ and for $c=0.5$, we let $\hat{v}^{(j),{\rm MLE}}_{i,k} = \vmax/N$ if $\sum_{t=(k-1)T/2 + 1}^{kT/2}\mathbf{1}\{Y_t = 0\}=0$ with $k=1,2$. For the derivation of the MLE we refer to Appendix E.3.

{\bf Performance measures.} 
Given a simulated data sample of size $T$, the predictive performance is measured in three ways: (1) the instantaneous relative regret of the estimated optimal assortment, (2) the $L_1$ error of the estimated  preference vector/function, and (3), in the same spirit as  \cite{Berbeglia2018}, the relative absolute difference between the estimated no-purchase probability and the actual no-purchase probability.

To ensure a fair comparison for the first performance measure, the optimal assortment in the first two scenarios is computed over $\mathcal{A}_K$, the collection of all unions of at most $K=cN$ bins. This is because, if the discrete model is the ground truth, then partial products can not be offered. 
In addition, in these first two scenarios, the estimated optimal assortment under the continuous model is computed with the function $w$ replaced by $\check{w}$, in line with Equation \eqref{eq:checkw}.
%
%
The instantaneous relative regret (IRR) is thus computed as
\[{\rm IRR}^{(j),{\rm E}} := \frac{r(S^{(j)},v^{(j)},w) - r(\hat{S}^{(j),{\rm E}},v^{(j)},w)}{r(S^{(j)},v^{(j)},w)},\qquad j=1,2,3,\:\:\:{\rm E} \in \{{\rm KDE},{\rm MLE}\},\]
where $S^{(j)}$ is the optimal assortment in scenario $j$ and $\hat{S}^{(j),{\rm E}}$ the estimated optimal assortment, for both estimators ${\rm E} \in \{{\rm KDE},{\rm MLE}\}$.
The second performance measure is defined as 
\[{\rm L}_1^{(j),{\rm E}} := \int_0^1 \Big| v^{(j)}(x)- \hat v^{(j),{\rm E}}(x)\Big| \d x,\qquad j=1,2,3,\:\:\:{\rm E} \in \{{\rm KDE},{\rm MLE}\},\]
where $\hat{v}^{(j),{\rm MLE}}$ and
$\hat{v}^{(j),{\rm KDE}}$ are the MLE and KDE estimator for scenario $j$, respectively. 
Finally, our third performance measure is the relative absolute difference of the actual no-purchase probability and the estimated no-purchase probability, where for $c = 1/2$ we average the relative absolute difference of the no-purchase probabilities for assortment $[0, 0.5]$ and $[0.5, 1]$. Thus, defining
\[Q^{(j)} := \frac{1}{1 + \int_0^1 v^{(j)}(x) \d x}\quad\text{and}\quad Q^{(j)}_k := \frac{1}{1 + \int_{S^k} v^{(j)}(x) \d x},\qquad j=1,2,3,\:k=1,2,\]
and
\[\hat{Q}^{(j),{\rm E}} := \frac{1}{1 + \int_0^1 \hat{v}^{(j),{\rm E}}(x) \d x}\quad\text{and}\quad \hat{Q}^{(j),{\rm E}}_k := \frac{1}{1 + \int_{S^k} \hat{v}^{(j),{\rm E}}(x) \d x},\:\:\: \begin{array}{l}
j=1,2,3,\:\:\:{\rm E} \in \{{\rm KDE},{\rm MLE}\},\\[-0.5em]
k=1,2,
\end{array}\]
then our third performance measure is equal to 
\[{\rm RAD}^{(j),{\rm E}} := \frac{\big|Q^{(j)} - \hat Q^{(j),{\rm E}}\big|}{Q^{(j)}},\qquad j=1,2,3,\:\:\:{\rm E}  \in \{{\rm KDE},{\rm MLE}\}.\]
for $c = 1$, and 
\[{\rm MRAD}^{(j),{\rm E}} := \frac{\big|Q^{(j)}_1 - \hat Q^{(j),{\rm E}}_1\big|}{2Q^{(j)}_1} + \frac{\big|Q^{(j)}_2 - \hat Q^{(j),{\rm E}}_2\big|}{2Q^{(j)}_2}\qquad j=1,2,3,\:\:\:{\rm E}  \in \{{\rm KDE},{\rm MLE}\},\]
for $c = 0.5$.

\subsection*{E.2. Results}

A priori one would expect that, in scenario 1, the predictive performance of the discrete model outperforms that of the continuous model, and that in scenario 3 it is the other way around. What happens in scenario 2 might be less predictable. 
The performance metrics in the three different scenarios are displayed in Figures \ref{fig:scenario1forc=0.5} through \ref{fig:scenario3forc=1}.

Regarding the third performance measure, there is hardly any difference between the continuous and discrete model. For the other two performance measures, however, we observe marked differences. In scenario 1 the continuous model outperforms the discrete model in several instances, especially for small values of $T$, both when $c=1$ and when $c = 0.5$. Similar behavior is seen in scenario 2: the continuous model outperforms the discrete model under the first two performance measures, except for $N = 10$ and sufficiently large $T$. In scenario 3, the continuous model outperforms the discrete model when measured by the first or second performance measure when $c=0.5$; when $c=1$, the first and second performance measure are either approximately equal, or the continuous model outperforms the discrete model.

These observations demonstrate that there is value in using the continuous model for predictive purposes, also in situations where this model is misspecified.

\input{tikzplots}

\subsection*{E.3. Maximum likelihood estimator}

Here we derive the maximum likelihood estimator for the preference parameters in the discrete MNL model. We denote as the estimators as $\hat{v}_1,\ldots,\hat{v}_N$. Following Appendix E.1 and E.2, we consider (i)~$K=N$ and offer the entire set of products $[N]$ at all time instances, as well as (ii)~$K=N/2$ and offer the assortments $\{1,\ldots,N/2\}$ and $\{N/2+1,\ldots,N\}$ (each in half of all time instances, that is).

First we consider that $K=N$ and $D_t=[N]$ for all $t\in[T]$. Let $i_t$ denote the discrete purchase observed at time $t$ when offering $D_t\subseteq[N]$. Then the log likelihood is
\[L(v_1,\ldots,v_N) = \sum_{t=1}^T \log\left(\frac{v_{i_t}}{1+\sum_{i=1}^N v_i}\right) = \sum_{t=1}^T \log v_{i_t} - \sum_{t=1}^T  \log\left(1+\sum_{i=1}^N v_i\right).\]
Taking the derivative of the log likelihood with respect to $v_j$ for $j\in [N]$ yields
\[\frac{\partial}{\partial v_j}L(v_1,\ldots,v_N) = \frac{1}{v_j}\sum_{t=1}^T\mathbf{1}\{i_t = j\}  - \sum_{t=1}^T \frac{1}{1+\sum_{i=1}^N v_i}.\]
These partial derivatives are equal to zero, so as to obtain $\hat{v}_j$ for $j\in [N]$; we obtain
\begin{equation}
\sum_{t=1}^T\mathbf{1}\{i_t = j\}  = \sum_{t=1}^T \frac{\hat{v}_j}{1+\sum_{i=1}^N \hat{v}_i}.
\label{eq:indipartial}
\end{equation}
Summing all these equations for $j\in [N]$ yields
\[\sum_{t=1}^T\mathbf{1}\{i_t \neq 0\} = \sum_{t=1}^T \frac{\sum_{j=1}^N\hat{v}_j}{1+\sum_{i=1}^N \hat{v}_i},\]
or, equivalently,
\begin{equation}
\sum_{t=1}^T\mathbf{1}\{i_t = 0\}  = \sum_{t=1}^T\frac{1}{1+ \sum_{i=1}^N\hat{v}_i}.
\label{eq:sumpartial}
\end{equation}
Combining \eqref{eq:indipartial} and \eqref{eq:sumpartial}, we obtain
\[\hat{v}_j := \frac{\sum_{t=1}^T \mathbf{1}\{i_t=j\}}{\sum_{t=1}^T \mathbf{1}\{i_t=0\}},\qquad j\in D,\]
where we set $\hat{v}_j:=\vmax/N$ if $\sum_{t=1}^T \mathbf{1}\{i_t=0\}=0$.
 
Next, we consider that $K=N/2$. Denote $D^1 = \{1,\ldots,N/2\}$ and $D^2 = \{N/2+1,\ldots,N\}$, as well as $\mathcal{T}^1 = \{1,\ldots,T/2\}$ and $\mathcal{T}^2 = \{T/2+1,\ldots,T\}$. Then $D_t=D^1$ for $t\in\mathcal{T}^1$ and $D_t=D^2$ for $t\in\mathcal{T}^2$. Let $i_1,\ldots,i_t$ denote the discrete purchases observed at time $t$ when offering $D_t\subseteq[N]$. Then the log likelihood is
\[L(v_1,\ldots,v_N) = \sum_{t=1}^T \log\left(\frac{v_{i_t}}{1+\sum_{i\in D_t} v_i}\right) = \sum_{t=1}^T \log v_{i_t} - \sum_{t=1}^T  \log\left(1+\sum_{i\in D_t} v_i\right).\]
Taking the derivative of the log likelihood with respect to $v_j$ for $j\in [N]$ yields
\[\frac{\partial}{\partial v_j}L(v_1,\ldots,v_N) =
\left\{\begin{array}{lll}
\ds \frac{1}{v_j}\sum_{t\in\mathcal{T}^1}\mathbf{1}\{i_t = j\}  - \sum_{t\in\mathcal{T}^1} \frac{1}{1+\sum_{i\in D^1} v_i}, && \text{for }j\in D^1, \\[1em]
\ds \frac{1}{v_j}\sum_{t\in\mathcal{T}^2}\mathbf{1}\{i_t = j\}  - \sum_{t\in\mathcal{T}^2} \frac{1}{1+\sum_{i\in D^2} v_i}, && \text{for }j\in D^2.
\end{array}\right.\]
These partial derivatives are set equal to zero, to obtain $\hat{v}_j$ for $j\in D^k$ and $k=1,2$. We thus obtain
\begin{equation}
\sum_{t\in\mathcal{T}^k}\mathbf{1}\{i_t = j\}  = \sum_{t\in\mathcal{T}^k} \frac{\hat{v}_j}{1+\sum_{i=D^k} \hat{v}_i}.
\label{eq:indipartial2}
\end{equation}
Summing all these equations over $j\in D^k$ yields
\[\sum_{t\in\mathcal{T}^k}\mathbf{1}\{i_t \neq 0\} = \sum_{t\in\mathcal{T}^k} \frac{\sum_{j\in D^k}\hat{v}_j}{1+\sum_{i\in D^k} \hat{v}_i},\]
or, equivalently,
\begin{equation}
\sum_{t\in\mathcal{T}^k}\mathbf{1}\{i_t = 0\}  = \sum_{t\in\mathcal{T}^k}\frac{1}{1+ \sum_{i\in D^k}\hat{v}_i}.
\label{eq:sumpartial2}
\end{equation}
Combining \eqref{eq:indipartial2} and \eqref{eq:sumpartial2}, we obtain
\[\hat{v}_j := \frac{\sum_{t\in\mathcal{T}^k} \mathbf{1}\{i_t=j\}}{\sum_{t\in\mathcal{T}^k} \mathbf{1}\{i_t=0\}},\qquad j\in D^k,\:k=1,2,\]
where we set $\hat{v}_j:=\vmax/N$ if $\sum_{t\in\mathcal{T}^k} \mathbf{1}\{i_t=0\}=0$.


\subsection*{E.4. Kernel density estimator}
In this section we define the kernel density estimator used to estimate the preference function $v$ in the continuous assortment model.

Because traditional kernel density estimation does not perform well near endpoints of the support, we construct the KDE based on the so-called boundary kernel method, that locally adjusts the kernels near the edges of the support \citep[see][for other demonstrations of this method]{Muller1991,Zhang1999}. Also contrary to traditional kernel density estimation, we allow the order of the kernel to depend on the number of observations. To construct such a kernel of arbitrarily high order, it is natural to work with an orthonormal basis of polynomials. We specifically choose Legendre polynomials since this choice allows us to bound the convergence rate explicitly for kernels of flexible order.

We define our estimator $\hat{v}$ of $v$ based on continuous purchases $X_1,\ldots, X_T\in[0,1]\cup\{\emptyset\}$. If our estimator is applied to scenario 1 or 2, in which case the observed purchases  $Y_1,\ldots,Y_T\in[N]\cup\{0\}$ are discrete, we draw $X_t$ uniformly at random from $B_{Y_t}$ if $Y_t\neq 0$ and set $X_t:=\emptyset$ if $Y_t=0$, for all $t\in[T]$.
We define the estimator $\hat{v}$ for the situation that there are $L \in \N$ so-called test assortments $S^1,\ldots, S_L$ each of which is offered during exactly $M \in \N$ time periods, and each of which has volume $c$. More precisely, the offered assortment at time $t\in[T]$ is $S_t=S^k$ if $t\in\{(k-1)M+1,\ldots,kM\}$. For all $x \in [0,1]$, let $e(x)$ denote the number of times that product $x$ is contained in the test assortments $S^1,\ldots,S^L$:
\[e(x):=\sum_{k=1}^L {\boldsymbol 1}_{S^k}(x).\]
We assume that the test assortments $S^1,\ldots,S^L$ cover the entire set of products $[0,1]$, that is, $e(x) > 0$ for all $x \in [0,1]$. For each test assortment $S^k$ we construct a corresponding estimate $\hat{v}_k(x)$ of $v(x){\boldsymbol 1}_{S^k}(x)$, and then combine these into our estimate
$\hat{v}$, as follows:
\begin{equation}
\hat{v}(x) := \frac{1}{e(x)} \sum_{k=1}^L \hat{v}_k(x),\qquad x\in[0,1].
\label{eq:hatv}
\end{equation}
To define $\hat{v}_k$, define the Legendre polynomials
\begin{equation*}
  \varphi_0(x) := \frac{1}{\sqrt{2}},\quad \varphi_j(x) := \sqrt{\frac{2j+1}{2}} \frac{1}{2^j j!}\frac{\text{d}^j}{{\rm d}x^j}\left[(x^2-1)^j\right],
\end{equation*}
for $j\in\N$, which form an orthonormal basis in $L_2([-1,1])$. Let $a_k$ and $b_k$ be such that $S^k= [a_k,b_k]$, for all $k \in [L]$, let $h\in(0,{c}/{2}]$ be a bandwidth parameter and for all $k\in [L]$ and $x \in \R$ define the shifted support $I^k_x$ as 
\[I^k_x = \left[ -\min\left\{1,\frac{x-a_k}{h}\right\}, \min\left\{1,\frac{b_k-x}{h}\right\}\right].\]
In addition, we define two shift coefficients $\gamma^k_x$ and $\zeta^k_x$ as
\[(\:\gamma^k_x\:,\:\zeta^k_x\:) = \left\{\begin{array}{rlll}
  \ds \left(\frac{2h}{h+x-a_k}\right. ,&\ds\left.-\frac{h-(x-a_k)}{h+x-a_k}\right)&& \text{ for } x\in [a_k,a_k+h), \\
  (1,&0) && \text{ for } x\in [a_k+h,b_k-h], \\
  \ds \left(\frac{2h}{h+b_k-x}\right.,&\ds\left.\frac{h-(b_k-x)}{h+b_k-x} \right)&& \text{ for } x\in (b_k-h,b_k], \\
\end{array}\right.\]
and define the Legendre kernel of order $\ell$ for $S^k$ by
\begin{equation*}
  K^k_x(u):=\gamma_x^k\sum_{j=0}^\ell \varphi_j\big(\zeta_x^k\big)\varphi_j\big(\gamma_x^k u + \zeta_x^k\big),\qquad x\in S^k,\: u\in I^k_x,
\end{equation*}
and $K^k_x(u):= 0$ for $x\in S^k$ and $u\notin I^k_x$.

Since $v(x)\mathbf{1}_{S^k}(x)$ is not a proper density, we re-scale the kernel estimator based on the number of (no)-purchases corresponding to test assortment $S^k$, for all $k \in [L]$. To this end, let $E_k$ denote the no-purchases observed when assortment $S^k$ is offered:
\[E_k:=\{X_t: X_t=\emptyset \:\text{ and }\:(k-1)M+1\leqslant t\leqslant kM\},\]
and let 
\[A_k:=\{X_t: X_t\neq\emptyset\:\text{ and }\:(k-1)M+1\leqslant t\leqslant kM\}\]
denote the actual purchases observed when $S^k$ is offered. Then we estimate $v(x)\mathbf{1}_{S^k}(x)$ by 
\begin{equation*}
\hat{v}_k(x):= \frac{1}{(|E_k|+1) h}\sum_{X\in A_k}K^k_x\left(\frac{X-x}{h}\right), \qquad x\in S^k,
\end{equation*}
and set $\hat{v}_k(x):=0$ for $x\notin S^k$. These estimates are combined into one estimate $\hat{v}(x)$ of $v(x)$, as given by \eqref{eq:hatv}. Analysis of the convergence rates reveals that an appropriate choice for the bandwidth parameter $h$ and order parameters $\ell$ is
\[h^* :=\min\left\{\frac{c}{2},\frac{1}{e}\right\}\qquad\text{and}\qquad\ell^* := \left[\frac{1}{2}\log\big(-2M\log h^*\big) -\frac{1}{2}\right],\]
respectively, where $[x]$ denotes the rounded value of $x\in\R$.

\end{document}